\documentclass[preprint]{elsarticle}

\advance\textwidth by 80pt

\advance\oddsidemargin by -40pt

\advance\evensidemargin by -40pt

\advance\textheight by 40pt

\advance\topmargin by -60pt

\usepackage[dvips]{epsfig}
\usepackage{epic}
\usepackage{array}
\usepackage{epstopdf}
\usepackage{booktabs}
\usepackage{a4wide}
\usepackage{verbatim}
\usepackage{amsmath}
\usepackage{amssymb}
\usepackage{amsfonts}
\usepackage{amstext}
\usepackage{amsbsy}
\usepackage{graphicx}
\usepackage[symbol]{footmisc}
\usepackage{subfigure}
\usepackage{longtable}
\usepackage{fancybox}
\usepackage{fancyvrb}
\usepackage{amsthm}
\usepackage{color}
\usepackage{url}
\usepackage{lineno}
\usepackage{natbib}

\newcommand{\beq}{\begin{equation}}
\newcommand{\eeq}{\end{equation}}

\newcommand{\bM}{\mathbf{M}}

\newcommand{\bS}{\mathbf{S}}

\newlength{\uvafigsize}
\DeclareMathOperator*{\argmax}{arg\,max}

\DeclareMathOperator*{\argmin}{arg\,min}

\DeclareMathOperator{\sign}{sign}

\begin{document}
	\begin{frontmatter}

\title{Cross-product Penalized Component Analysis (XCAN)}

\author{Jose Camacho \fnref{fn1}}
\author{Evrim Acar \fnref{fn2}}
\author{Morten A. Rasmussen \fnref{fn3}}
\author{Rasmus Bro \fnref{fn3}}
\fntext[fn1]{Department of Signal Theory, Telematics and Communications, School of Computer Science and Telecommunications - CITIC,	University of Granada, Spain}
\fntext[fn2]{Simula Metropolitan Center for Digital Engineering, Oslo, Norway}
\fntext[fn3]{Chemometrics and Analytical Technology, University of Copenhagen, Denmark}
\cortext[cor1]{Corresponding author: J Camacho josecamacho@ugr.es} 

\date{}

\begin{abstract}
Matrix factorization methods are extensively employed to understand complex data. In this paper, we introduce the cross-product penalized component analysis (XCAN), a sparse matrix factorization based on the optimization of a loss function that allows a trade-off between variance maximization and structural preservation. The approach is based on previous developments, notably (i) the Sparse Principal Component Analysis (SPCA) framework based on the LASSO, (ii) extensions of SPCA to constrain both modes of the factorization, like co-clustering or the Penalized Matrix Decomposition (PMD), and (iii) the Group-wise Principal Component Analysis (GPCA) method. The result is a flexible modeling approach that can be used for data exploration in a large variety of problems. We demonstrate its use with applications from different disciplines.
\end{abstract}

\end{frontmatter}

\section{Introduction}

Matrix factorization methods, which decompose a matrix into a product of factors, are extensively employed to understand complex data. Factors are often useful for highlighting and interpreting special observations (outliers), clusters of similar observations, groups of related variables, and crossed
relationships between observations and variables. 

Principal component analysis (PCA) \cite{Jolliffe02, Jackson03} is a key tool
for that purpose. PCA factorizes a matrix using the criterion of maximizing variance. The PCA model follows the expression:
\begin{equation} \label{eq:PCAm}
\mathbf{X} = \mathbf{\hat{T}}\mathbf{\hat{P}}^T +
\mathbf{E},
\end{equation}
where $\mathbf{X}$ is a $N \times M$ data matrix,
$\mathbf{\hat{T}} (N \times K)$ is the score matrix containing the
projection of the objects onto the principal components (PCs)
sub-space, $\mathbf{\hat{P}} (M \times K)$ is the loading matrix
containing the linear combination of the variables represented in
each of the PCs, and $\mathbf{E} (N \times M)$ is the matrix of
residuals. PCA satisfies: 

\begin{equation} \label{pca}
\{\mathbf{\hat{P}}, \mathbf{\hat{T}}\} = \argmin_{\mathbf{P},\mathbf{T}} \| \mathbf{X} - \mathbf{T}\mathbf{P}^T \|_F^2 
\end{equation}

\noindent where $\| \cdot \|_F$ stands for the Frobenius norm. Depending on the area of application, loading vectors are  constrained to unit length, in order to leave the data variance in the scores to ease interpretation. One interesting property of PCA is that loading vectors can be computed simultaneously or sequentially with exactly the same parameter estimates. This property is a consequence of the components being orthogonal, and while that leads to nice mathematical properties, it seldom reflects the underlying biological or chemical reality.

{When used for interpretation by exploring the PCs, PCA has a major shortcoming: The PCs are linear combinations of all the variables, and often combine unrelated sources of variance \cite{Camacho}. On the other hand, for easier interpretation, it is desirable to find factorizations that correspond to a limited number of original variables. 
This can be achieved by means of rotation \cite{Jolliffe02} or sparse methods like sparse principal component analysis (SPCA)  \cite{Jolliffe2003, Zou2006}. Another approach is to constrain loadings to agree with the structure of the correlation matrix, like in the group-wise principal component analysis (GPCA) \cite{GPCA}.}

While previous methods focus on the mode of the variables (columns) of the data, PCA shows exactly the same limitation for interpreting the mode of the observations: i) score vectors are typically non-sparse, and ii) unrelated observations can provide very similar scores. Extensions that apply the sparsity idea to both modes already exist, like some variants of co-clustering \cite{coclustering} or the penalized matrix decomposition (PMD) \cite{PMD}. 

In this paper, we introduce the cross-product penalized component analysis (XCAN). XCAN is a matrix factorization based on a loss function that allows a trade-off between variance maximization and structural preservation, aimed at solving the aforementioned problems. The result is a flexible modeling approach that can be used for data exploration in a large variety of problems. We will demonstrate its use with applications from different disciplines.

The rest of the paper is organized as follows. Section 2 introduces the methods on which XCAN is based. Section 3 presents the XCAN algorithm. Section 4 illustrates XCAN through four case studies, including simulated as well as real data from different application fields. Conclusions and future work are discussed in Section 5.

\section{Related methods}

The proposed XCAN method is inspired by previous developments, notably (i) the SPCA framework based on the lasso (least absolute shrinkage and selection operator), (ii) extensions of SPCA to constrain both modes of the factorization, like co-clustering or the PMD, and (iii) GPCA. From SPCA, we inherit the approach of defining a set of meta-parameters to define the loss function as a trade-off. This trade-off is between captured variance and structural penalties, following the GPCA strategy. The new loss function is intended to reflect both the structure among observations and among variables, in a similar way as in co-clustering or PMD.

\subsection{Sparse PCA}

The SPCA idea is grounded on the work of several authors more than two decades ago, as described in \cite{Mackey2008}. There are various versions of SPCA, most based on the modification of the PCA loss in eq. (\ref{pca}) by including sparsity-inducing constraints or penalties with the $L_0$ or $L_1$ norms \cite{Richt2012}. The $L_0$-norm of a vector refers to the number of non-zero elements in the vector, and the $L_1$-norm of a vector computes the sum of the absolute values of the vector entries. The application of the $L_1$-norm in a regression setting was originally called the least absolute shrinkage and selection operator (lasso) \citep{Tibshirani1994}. In this section, we focus on the lasso versions of SPCA for its widespread use in model interpretability \cite{Rasmussen2012}.

The SCoTLASS algorithm \citep{Jolliffe2003} incorporates the lasso in the PCA calibration as follows:

\begin{equation} \label{lasso}
\mathbf{\hat{p}}^{SL} = \argmax_{\mathbf{p}} \| \mathbf{X} \mathbf{p} \|_2^2 \ s.t. \  \|\mathbf{p}\|_1 \le c, \ \|\mathbf{p}\|_2^2 = 1
\end{equation}

\noindent where $\mathbf{\hat{p}}^{SL}$ is the resulting sparse loading, and $\| \cdot \|_1$ and $\| \cdot \|_2$ refer to the $L_1$ and $L_2$ norms, respectively. To obtain successive components, the SCoTLASS optimization constrains the second and further {sparse loadings} to be orthogonal to the rest. 

The SCoTLASS criterion is very computational demanding \cite{Hastie:2015:SLS:2834535} and has numerical limitations. For this reason, Zou \textit{et al.} \cite{Zou2006} introduced an alternative formulation to generate sparse components based on regularized regression, using a criterion close to the (naive\footnote{The regular elastic net makes use of a scaling factor between the lasso and the ridge penalty.}) elastic net \citep{Zou2005}, which is a combination of both the lasso and the ridge ($L_2$-norm) penalties:


\begin{equation} \label{SPCA}
\{\mathbf{\hat{P}}^{SP}, \mathbf{\hat{Q}}^{SP}\} = \argmin_{\mathbf{P},\mathbf{Q}} \| \textbf{X} - \mathbf{X} \mathbf{P} \mathbf{Q}^T\|_F^2 + \lambda_2 \sum_{h=1}^H \|\mathbf{p}_h\|^2_2 + \lambda_1 \sum_{h=1}^H \|\mathbf{p}_h\|_1 \ s.t.\  \mathbf{Q}^T\mathbf{Q} = \mathbf{I}
\end{equation}

\noindent where we distinguish between sparse loadings $\mathbf{\hat{P}}^{SP}$ (sometimes also referred to as weights in similar modeling frameworks) and orthonormal loadings $\mathbf{\hat{Q}}^{SP}$, $\mathbf{p}_h$ represents the $h$th column vector in $\mathbf{P}$, with $H$ the number of components. The solution proposed for eq. (\ref{SPCA}) is a biconvex optimization where sparse weights and orthogonal loadings are obtained using an alternating approach.
The algorithm is simultaneous, in the sense that all components are computed in the same alternating iteration. An alternative sequential variant is defined in \cite{Sjostrand2012}. A particular solution of SPCA in eq. (\ref{SPCA}) is when $\lambda_2 = \infty$, which is the most popular choice when the number of columns in the data is much higher than the number of rows. Then, the sparse loadings can be computed by soft-thresholding, simplifying and improving the efficiency of the computation: $g_{\lambda}(p) = \sign(p)\left(|p|-\lambda \right)_{+}$.

 \subsection{Extensions of sparsity to both modes}
 
Witten \textit{et al.} \cite{PMD} propose a new sparse algorithm referred to as the Penalized Matrix Decomposition (PMD). It can be used to constrain both the number of observations and variables contributing to each factor using soft-thresholding. The PMD follows:
\begin{equation} \label{PMD}
\{\mathbf{\hat{p}}^{P}, \mathbf{\hat{u}}^{P}\} = \argmax_{\mathbf{p},\mathbf{u}}\  \mathbf{u}^T \mathbf{X} \mathbf{p}   \ s.t. \ \|\mathbf{u}\|_1 \le c_1\ \|\mathbf{p}\|_1 \le c_2, \ \|\mathbf{u}\|_2^2 \le 1, \ \|\mathbf{p}\|_2^2 \le 1
\end{equation}
\noindent The corresponding pseudo-singular value is then obtained as:
\begin{equation} \label{PMD2}
\hat{d}^{P} = (\mathbf{\hat{u}}^{P})^T \mathbf{X} \mathbf{\hat{p}}^{P}
\end{equation}

\noindent After each component is obtained, projection deflation is performed as:
\begin{equation}
\mathbf{X} = \mathbf{X} (\mathbf{I} - \hat{d}^{P} \mathbf{\hat{u}}^{P}(\mathbf{\hat{p}}^{P})^T)
\end{equation}

\noindent The authors show that this solution, when only applying sparsity to the loadings, is connected with SCoTLASS and SPCA \cite{PMD}. 

A similar approach was introduced in \cite{coclustering} with the goal of performing coclustering:  
%
\begin{equation} \label{coclustering}
\{\mathbf{\hat{P}}^{C}, \mathbf{\hat{T}}^{C}\} = \argmin_{\mathbf{P},\mathbf{T}} \| \textbf{X} - \textbf{T} \mathbf{P}^T \|_F^2 + \lambda \sum_{h=1}^H \|\mathbf{p}_h\|_1 + \lambda  \sum_{h=1}^H \|\mathbf{t}_h\|_1
\end{equation}

\noindent where $\mathbf{t}_h$ represents the $h$th column vector in $\mathbf{T}$, and the same penalty variable is used in both modes. This loss is used within an alternating optimization, which, unlike PDM, produces all components in a single run.

\subsection{Group-wise PCA}

{GPCA limits each component to represent the variability of a single group of variables, which in turn allows to interpret each component independently (provided the variables are not overlapping)}. GPCA starts with the identification of a set of $K$ (possibly overlapping) groups of correlated variables. This is achieved by applying thresholding in a pseudo-correlation matrix $\bM (M \times M)$ of the data. In the original formulation of GPCA, the MEDA approach (Mis\-sing-data for Exploratory Data analysis) \cite{Camacho2011missing} was implemented to obtain $\bM$. 

Besides, GPCA {may be simpler to use in practice than sparse methods based on the lasso, because by inspecting $\bM$ we can often identify suitable values for the threshold or even when the GPCA model is not appropriate at all.} In comparison, the main challenge when using sparse methods is to find suitable values for meta-parameters like $c$ (\ref{lasso}), $\lambda_1$ and $\lambda_2$ (\ref{SPCA}), $c_1$ and $c_2$ (\ref{PMD}) or $\lambda$ (\ref{coclustering}). This advantage, however, comes at a price. While the capability to reflect the structure in the map  $\bM$ is an appealing property, GPCA is an ``all or nothing'' approach, meaning that a variable is either in a group or not, and different GPCA models can be obtained for {very similar} values of the threshold.

\section{Cross-product Penalized Component Analysis (XCAN)}

With XCAN, we would like to make the most of the advantages of sparse methods in one or the two modes and {combine that with the idea behind} GPCA. Like PMD, XCAN factorizes a matrix { into three matrices}:
\begin{equation} \label{eq:XPCAm}
\mathbf{X} = \mathbf{\hat{U}}^{X}\mathbf{\hat{S}}^{X}(\mathbf{\hat{P}}^{X})^T +
\mathbf{E}
\end{equation}

The loss function for the XCAN factorization is as follows: 
\begin{equation} \label{loss}
\{\mathbf{\hat{P}}^{X}, \mathbf{\hat{S}}^{X}, \mathbf{\hat{U}}^{X}\} = \argmin_{\mathbf{P},\mathbf{S},\mathbf{U}} \| \textbf{X} - \textbf{U} \mathbf{S} \mathbf{P}^T \|_F^2 + \lambda_r F_r + \lambda_c F_c \ s.t. \ \|\mathbf{u}_h\|_2^2 \le 1, \ \|\mathbf{p}_h\|_2^2 \le 1
\end{equation}

\noindent where the first part is {the actual model}, with $\bS$ constrained to be diagonal, and $F_c$ and $F_r$ are defined to constrain the structure of the model. The meta-parameters $\lambda_c$ and $\lambda_r$ control the level of these penalties. We define the penalties as follows:

\begin{equation} \label{Fr}
F_r = \sum_{h=1}^H \| (\mathbf{u}_h\mathbf{u}_h^T) \oslash \mathbf{XXt} \|_F^2,
\end{equation}

\begin{equation} \label{Fc}
F_c = \sum_{h=1}^H \| (\mathbf{p}_h\mathbf{p}_h^T) \oslash \mathbf{XtX}\|_F^2,
\end{equation}

\noindent where $\mathbf{u}_h$ and $\mathbf{p}_h$ are the $h$th column vectors in $\mathbf{U}$ and $\mathbf{P}$, respectively, $H$ is the total number of XCAN components (XCs), $\oslash$ is the Hadamard (element-wise) division. $\mathbf{XtX} (M \times M)$ and $\mathbf{XXt} (N\times N)$ denote maps of relationship between variables and observations, respectively, and are given as inputs. To avoid numerical problems in the divisions, values below a threshold in $\mathbf{XtX}$ and $\mathbf{XXt}$ are set to that threshold. In this paper, we fixed the value of this threshold to 0.01 in all experiments.

\subsection{Cross-product matrices and rationale}

{We will generally refer to $\mathbf{XtX}$ and $\mathbf{XXt}$ as cross-product matrices, because several of their possible definitions can be computed as cross-products, and we expect them to be symmetric. This is also the reason for the name XCAN, where ``X'' stands for ``cross". For instance, $\mathbf{XtX}$ may be set to the correlation matrix of $\mathbf{X}$ and $\mathbf{XXt}$ to the correlation matrix of $\mathbf{X}^T$, respectively. 
$\mathbf{XtX}$ can also be set to any map $\bM$ used in GPCA.
	
	 The rationale behind the definition of the loss in (\ref{loss})-(\ref{Fr}) is as follows. $\mathbf{XtX}$ and $\mathbf{XXt}$ contain the {correlation} structure in the variables and observations, respectively. Values close to 0 in those matrices identify unrelated variables or observations. Using an element-wise division, we prevent unrelated elements to be part of the same component. That way, we obtain sparse components that agree with the structure enforced by the input cross-product matrices. 
	 
	 The choice of matrices $\mathbf{XtX}$ and $\mathbf{XXt}$ is principal in XCAN, and should be carefully done taking into account the goal of the analysis, in a similar way to when selecting the data pre-processing. 
	 For instance, it is customary to mean-center data for some types of analyses, but not for others (e.g., for spectra). Although not compulsory, using matrices $\mathbf{XtX}$ and $\mathbf{XXt}$ that are consistent with the data pre-processing is expected to provide more coherent results. 
	 By taking that into account, we use the following definition of cross-product in our experiments in this paper, inspired by Pearson's correlation:
	 \begin{equation} \label{XtX}
	 \mathbf{XtX} = \mathbf{X}^T\mathbf{X} \oslash \|diag(\mathbf{X}^T\mathbf{X})\|_2
	 \end{equation}    
	  \begin{equation} \label{XXt}
	 \mathbf{XXt} = \mathbf{X}\mathbf{X}^T \oslash \|diag(\mathbf{X}\mathbf{X}^T)\|_2
	 \end{equation}     
	
	The advantage of these definitions is that they do not require data to be column-wise or row-wise mean-centered, like correlation matrices do.
	
	Cross-product matrices can be conveniently post-processed to modify the behavior of XCAN, which adds flexibility to the modeling approach. For instance, a possible post-processing operation is thresholding. As discussed before, GPCA was defined with the suitable property that its meta-parameter (i.e., the threshold) can be set upon visual inspection of the map $\bM$. Similarly, we can inherit this idea in XCAN by thresholding $\mathbf{XtX}$ and $\mathbf{XXt}$, in order to discard {minor} correlations from the analysis. This results in useful means to further impose sparsity, as in GPCA. We can also use thresholding in such a way that only positive correlations are kept in $\mathbf{XtX}$ and $\mathbf{XXt}$. This can be useful to derive loadings and scores vectors in XCAN where all non-zero elements have the same sign, something in line with non-negativity constraints. In the examples, we will use this thresholding idea or non-negativity constraints, when suitable. 
	
	Since cross-product matrices can be considered as a way to include structural penalties in both modes of the data in XCAN, we can also use them in different ways, e.g., as in chemometrics literature, to impose smoothness, to connect different data sets of same individuals or variables (data fusion) or to include apriori information into a model. Studying those applications in detail is out of the scope of this paper. We will, however, show an application of XCAN by incorporating the class labels of the samples into the model.}

\subsection{Algorithmic Approach}
The XCAN model is fitted to the data by solving for all components simultaneously using gradient-based all-at-once optimization. To constrain the vectors in $\mathbf{P}^{X}$ and $\mathbf{U}^{X}$ to unit length, like in the SVD, a suitable way is to redefine the loss as:
\begin{equation} \label{F}
\{\mathbf{\hat{P}}^{X}, \mathbf{\hat{S}}^{X}, \mathbf{\hat{U}}^{X}\} = \argmin_{\mathbf{P},\mathbf{S},\mathbf{U}} \| \textbf{X} - \textbf{U} \mathbf{S} \mathbf{P}^T \|_F^2 +  \lambda_0 F_0 + \lambda_c F_c + \lambda_r F_r, 
\end{equation}
\noindent where
\begin{equation} \label{F0}
F_0 = \sum_{h=1}^H (\|\mathbf{p}_h\|_2^2-1)^2 + (\|\mathbf{u}_h\|_2^2-1)^2
\end{equation}
If $\lambda_0$ is set to a sufficiently large value, this additional term in the loss will serve the purpose of normalizing factors $\mathbf{a}_h$ and $\mathbf{b}_h$. In our experiments in Section 4, we set $\lambda_0 = 1$. 

We solve eq. (\ref{F}) by computing the partial derivatives of the loss function with respect to $\mathbf{U}, \mathbf{S}$ and $\mathbf{P}$ (as given in the Appendix), constructing the gradient, and then using a gradient-based optimization algorithm.
In our experiments, we use the Poblano Toolbox \cite{DuKoAc10}, that has several unconstrained gradient-based optimization algorithms such as the nonlinear conjugate gradient (NCG), and limited-memory BFGS (L-BFGS) \cite{NoWr06}. When non-negativity constraints are desired, we can also use the limited-memory BFGS with bound constraints (LBFGS-B)\footnote{We use the implementation at \url{https://github.com/stephenbeckr/L-BFGS-B-C}.}.

It should be noted that the XCAN optimization can have many local minima. In the experiments, we initialized the algorithm using the PCA solution, but ideally, multiple random starts would help with the local minima problem.

\section{Experiments}

\subsection{Simulation}

We start with a simulated experiment to demonstrate the properties and flexibility of XCAN. We simulate three data sets $\mathbf{X}_1$, $\mathbf{X}_2$ and $\mathbf{X}_3$, each with 5 observations and 5 variables. Each data set is generated with high correlation between the variables, using the simuleMV tool \cite{CAMACHO201740}. We, then, construct $\mathbf{X}$ as follows:
 \begin{equation} \label{sim}
 \mathbf{X} = \left[ \begin{array}{cc}
 \mathbf{X}_1 & \mathbf{0} \\
 \mathbf{0} & \mathbf{X}_2 \\
 \mathbf{0} & 2\mathbf{X}_3 \\
 \end{array}\right] + 0.15 \cdot \mathbb{N}(\mathbf{0},\mathbf{I}),
 \end{equation}
	
\noindent where $\mathbb{N}(\mathbf{0},\mathbf{I})$ denotes noise randomly drawn from the normal distribution with 0 mean and unit variance. {Finally, $\mathbf{XtX}$ follows eq. (\ref{XtX}) and $\mathbf{XXt}$ eq. (\ref{XXt})}.

\begin{figure*}
	\centering \subfigure[]{\includegraphics[width=0.4\textwidth]{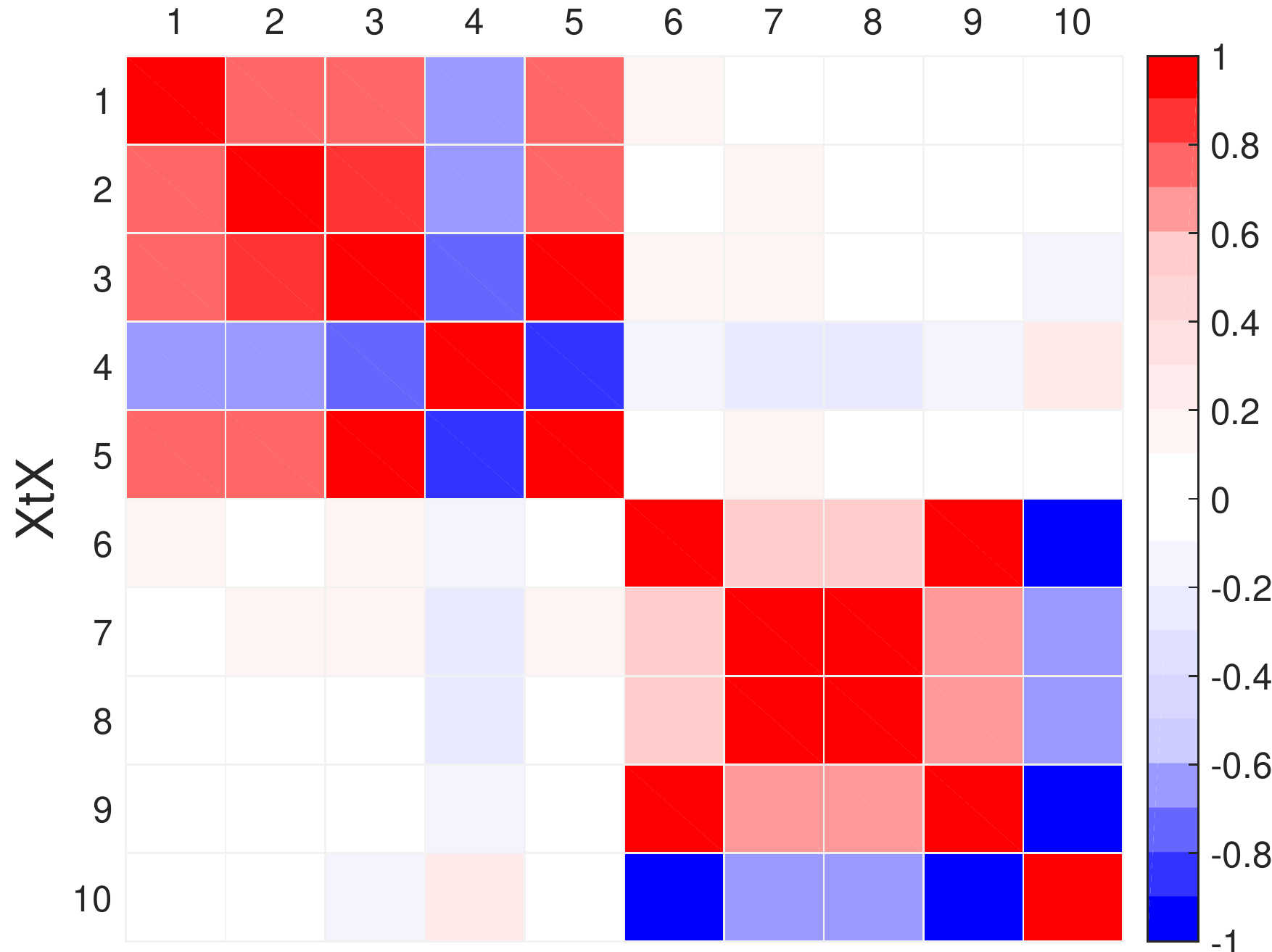}} 
	\hspace{1cm} \subfigure[]{\includegraphics[width=0.4\textwidth]{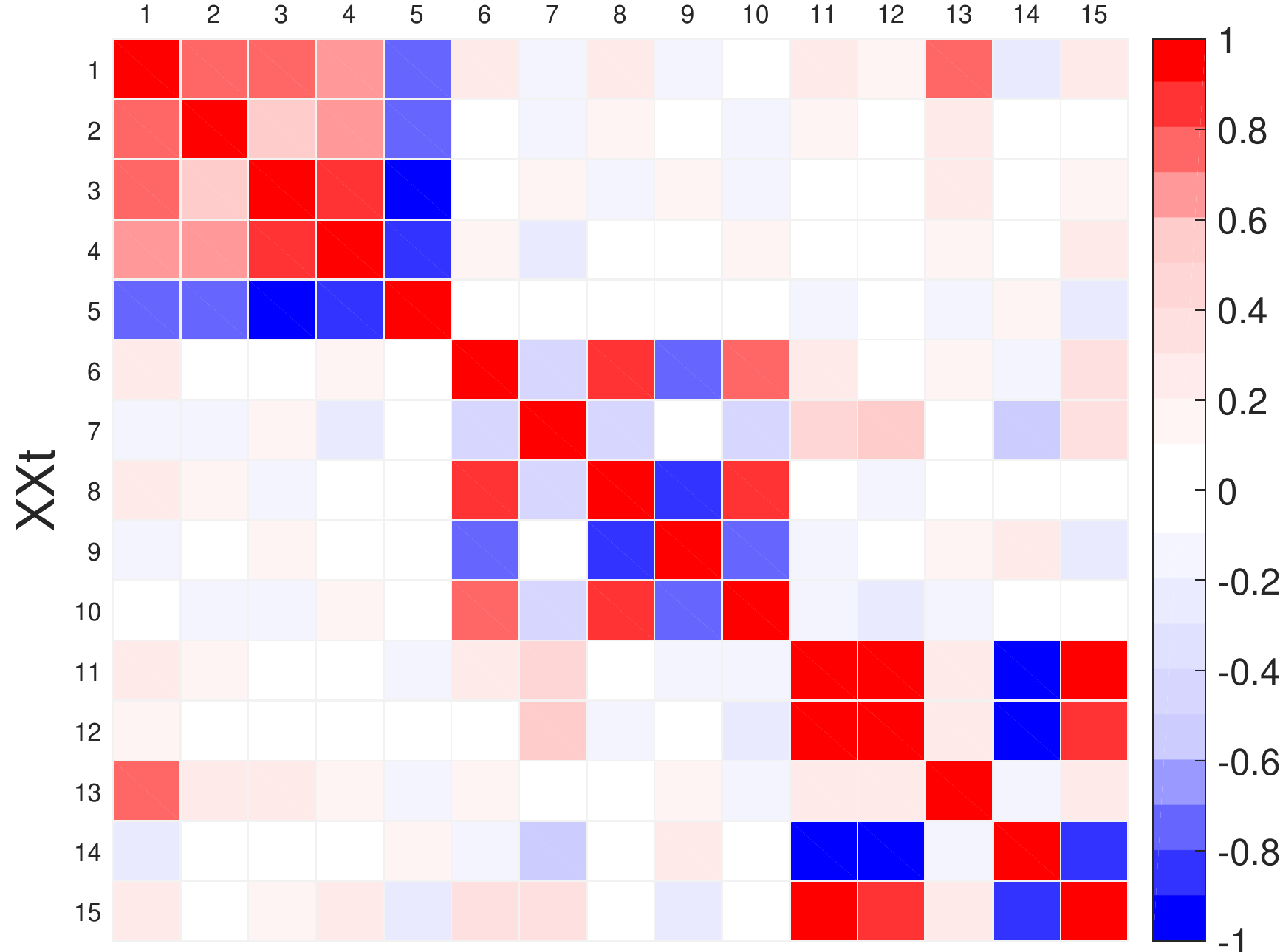}}
	\caption{Cross-product matrices to impose structural penalties in XCAN in the simulated example: (a) $\mathbf{XtX}$ and (b) $\mathbf{XXt}$.}
	\label{fig:Sim1}
\end{figure*}

Matrices $\mathbf{XtX}$ and $\mathbf{XXt}$ for one of the simulations are shown in Figure \ref{fig:Sim1}. We can see that variables (Figure \ref{fig:Sim1}(a))  are approximately grouped in two groups of 5 and observations (Figure \ref{fig:Sim1}(b)) describe three major groups.

\begin{figure*}
	\centering
	\subfigure[PCA]{\includegraphics[width=0.4\textwidth]{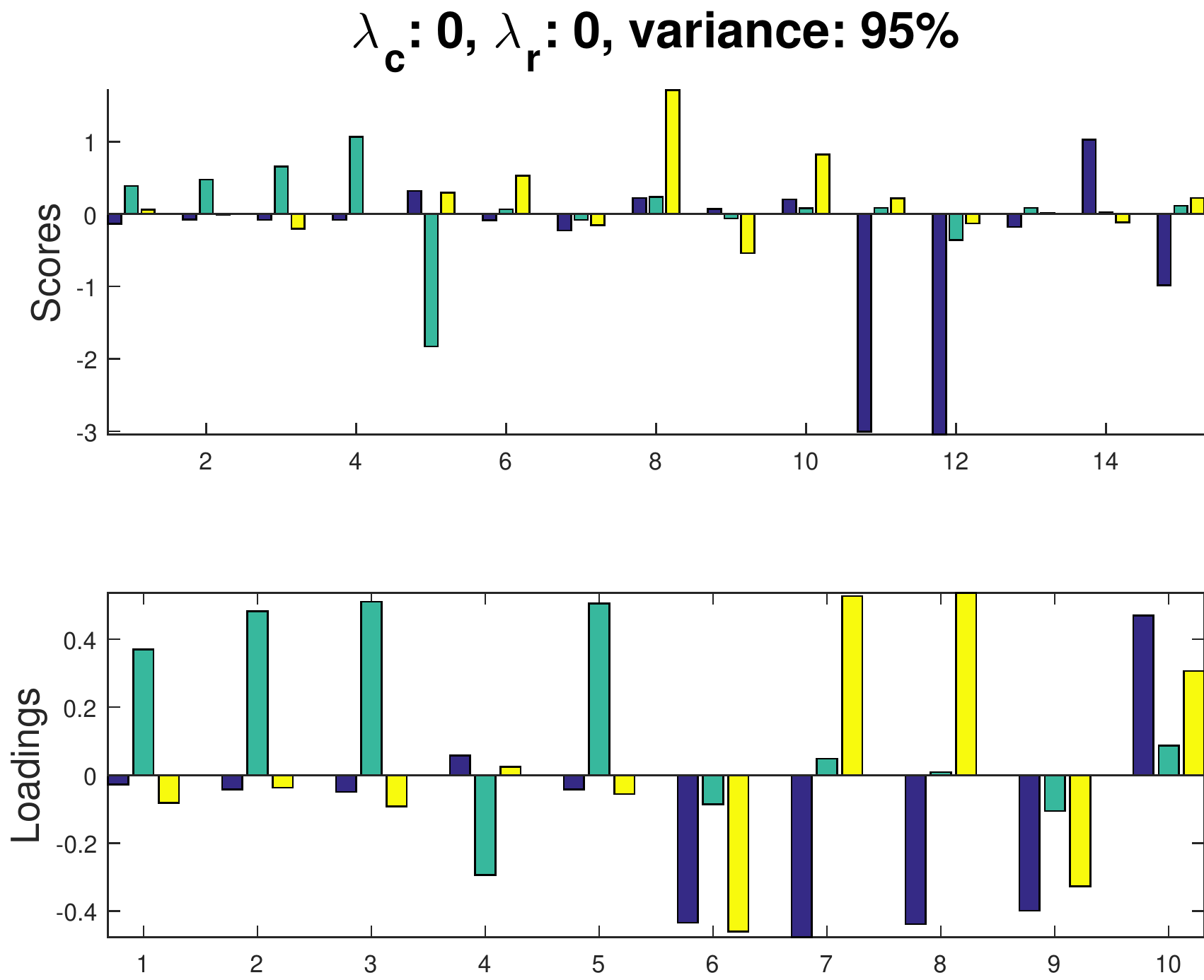}}  
	\hspace{1cm}
	\subfigure[sparse loadings]{\includegraphics[width=0.4\textwidth]{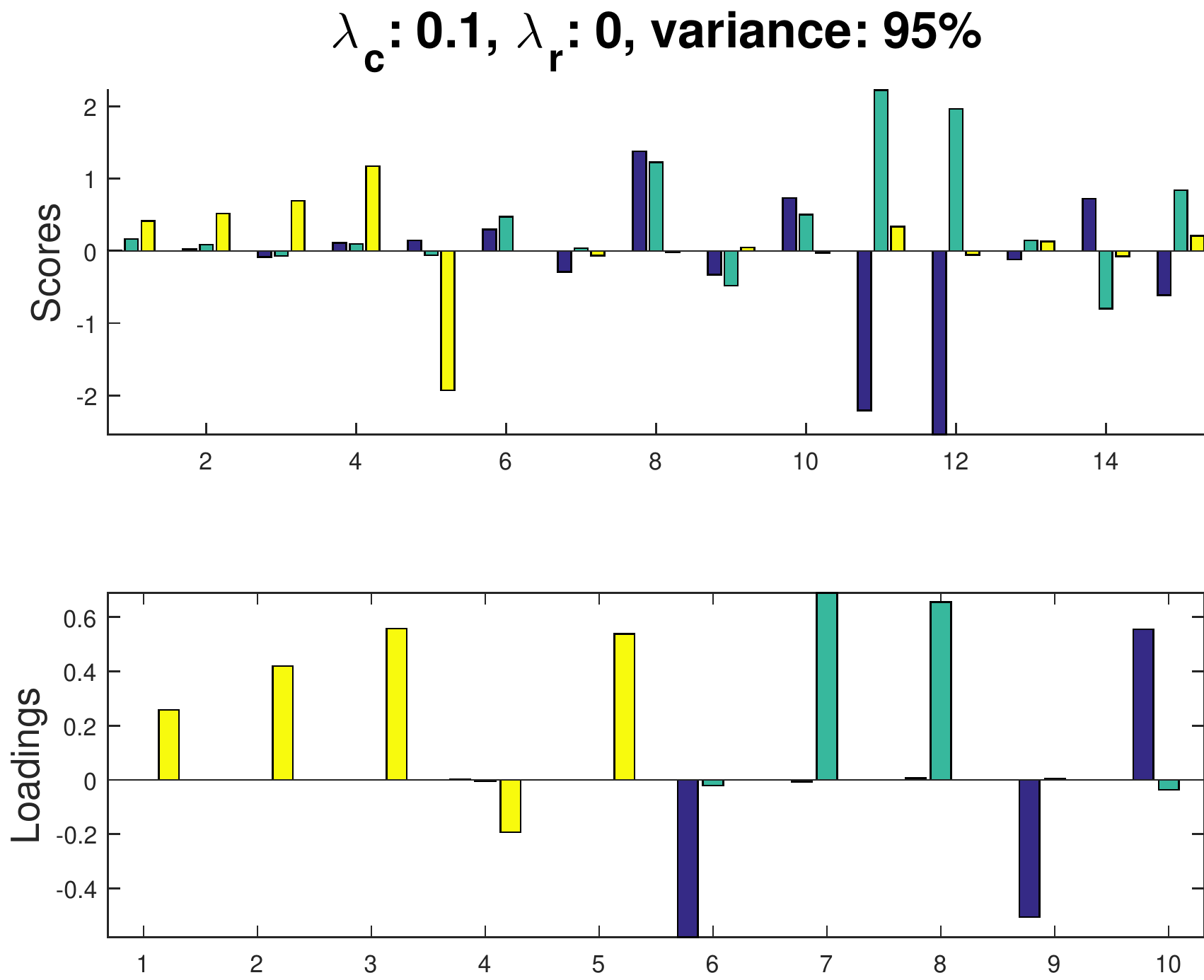}} \\
	\subfigure[sparse scores]{\includegraphics[width=0.4\textwidth]{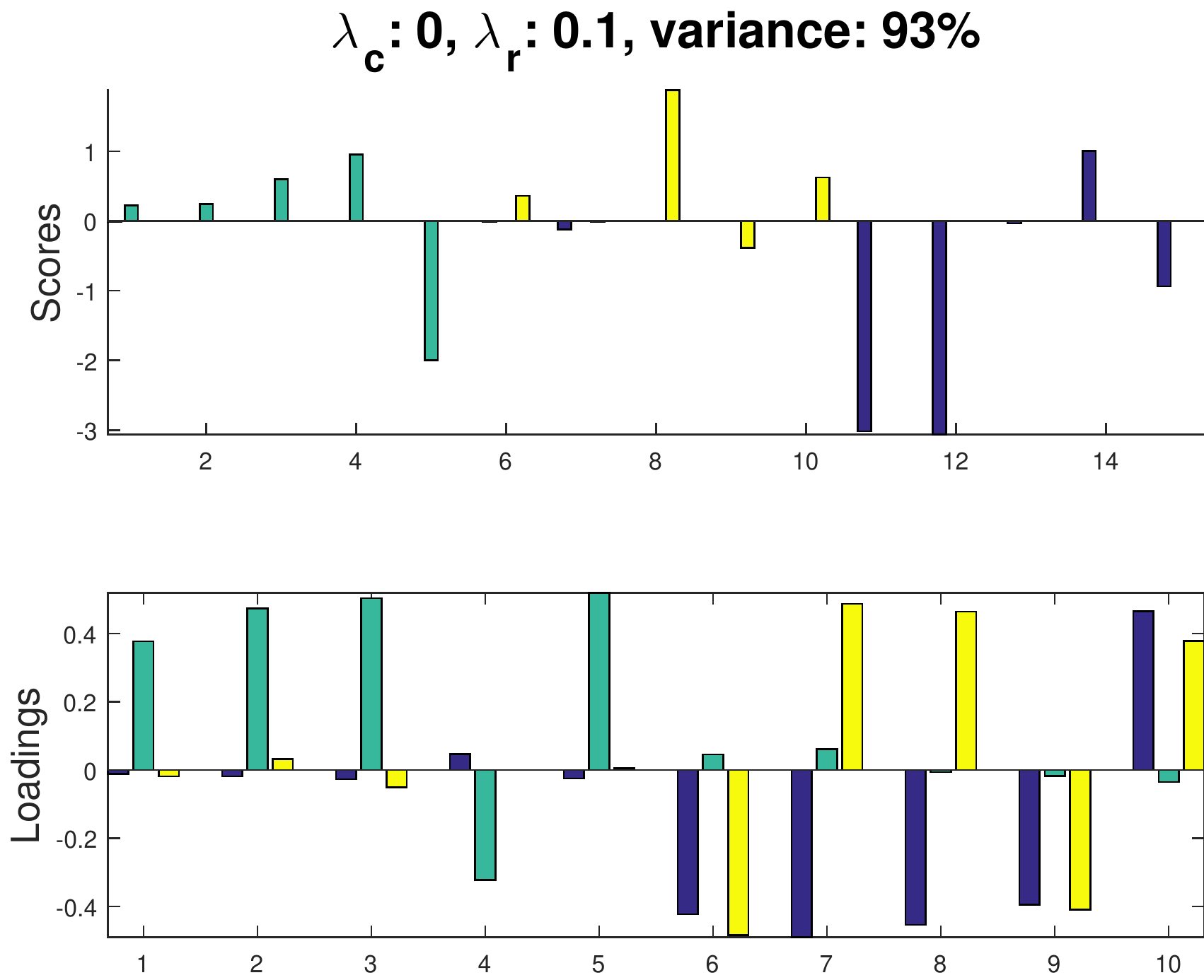}} 
	\hspace{1cm}
	\subfigure[sparse loadings and scores]{\includegraphics[width=0.4\textwidth]{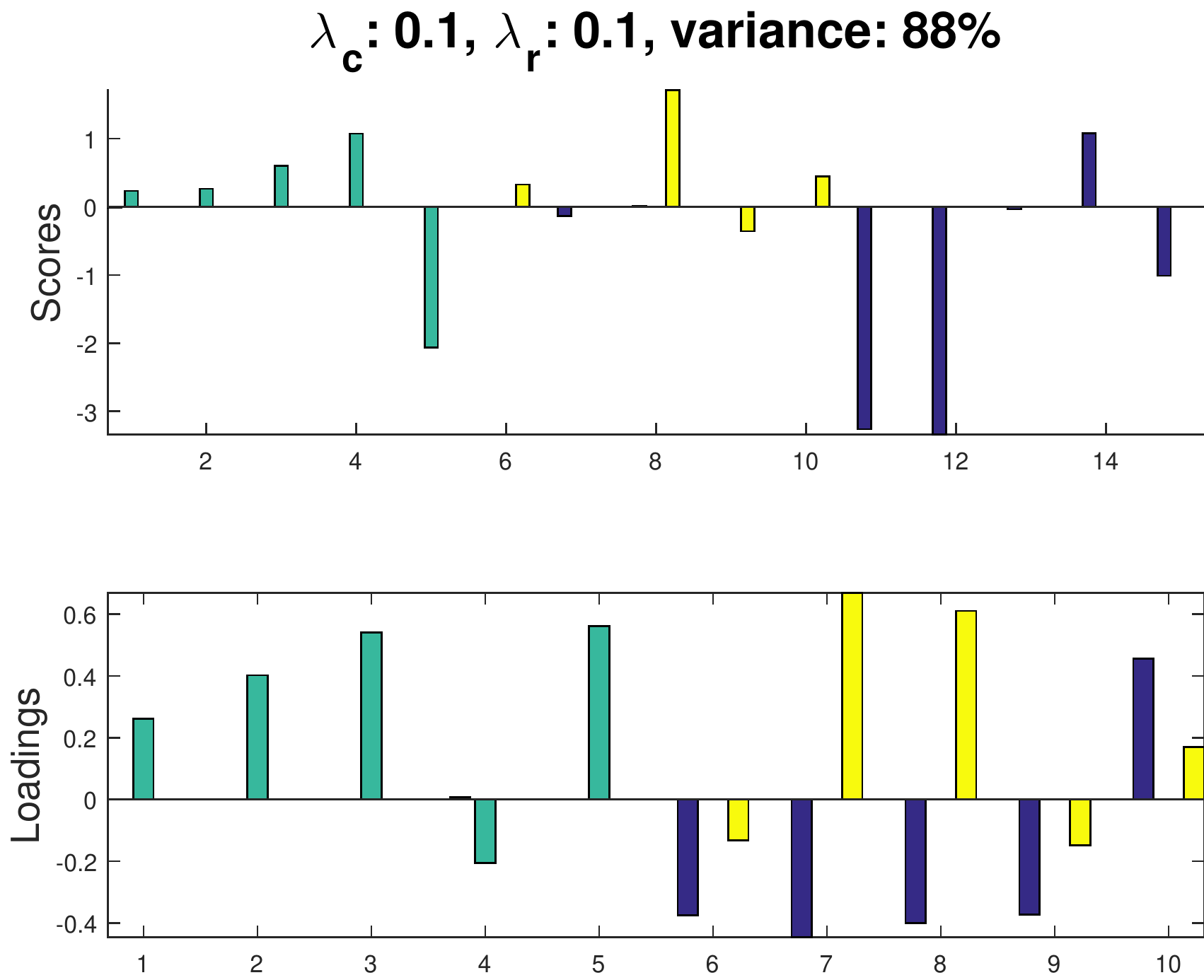}}
	\caption{XCAN models for the simulated data (using three components).}
	\label{fig:Sim2}
\end{figure*}

	Figure \ref{fig:Sim2} shows the result of applying XCAN with three components and for different values of the meta-parameters $\lambda_c$ and $\lambda_r$. The figure shows an upper bar plot with the scores, computed as $\mathbf{\hat{T}}^{X} =   \mathbf{\hat{U}}^{X} \mathbf{\hat{S}}^{X}$, and a lower bar plot with the loadings $\mathbf{\hat{P}}^{X}$. Figure \ref{fig:Sim2}(a) shows regular PCA, since the structural penalties are deactivated, which is used as a baseline. In such setting, each loading/score vector contains information about all variables/observations, respectively.  Figures \ref{fig:Sim2}(b), \ref{fig:Sim2}(c) and \ref{fig:Sim2}(d) show the application of XCAN with structural penalties in the loadings, scores, and both loadings and scores, respectively. In all cases, the XCAN model works as expected, and the variance remains reasonably close to the PCA model. This shows that the penalties meet the true data structure at a minor price in terms of explained variance. 

\begin{figure*}
	\centering
	\subfigure[1-component XCAN]{\includegraphics[width=0.4\textwidth]{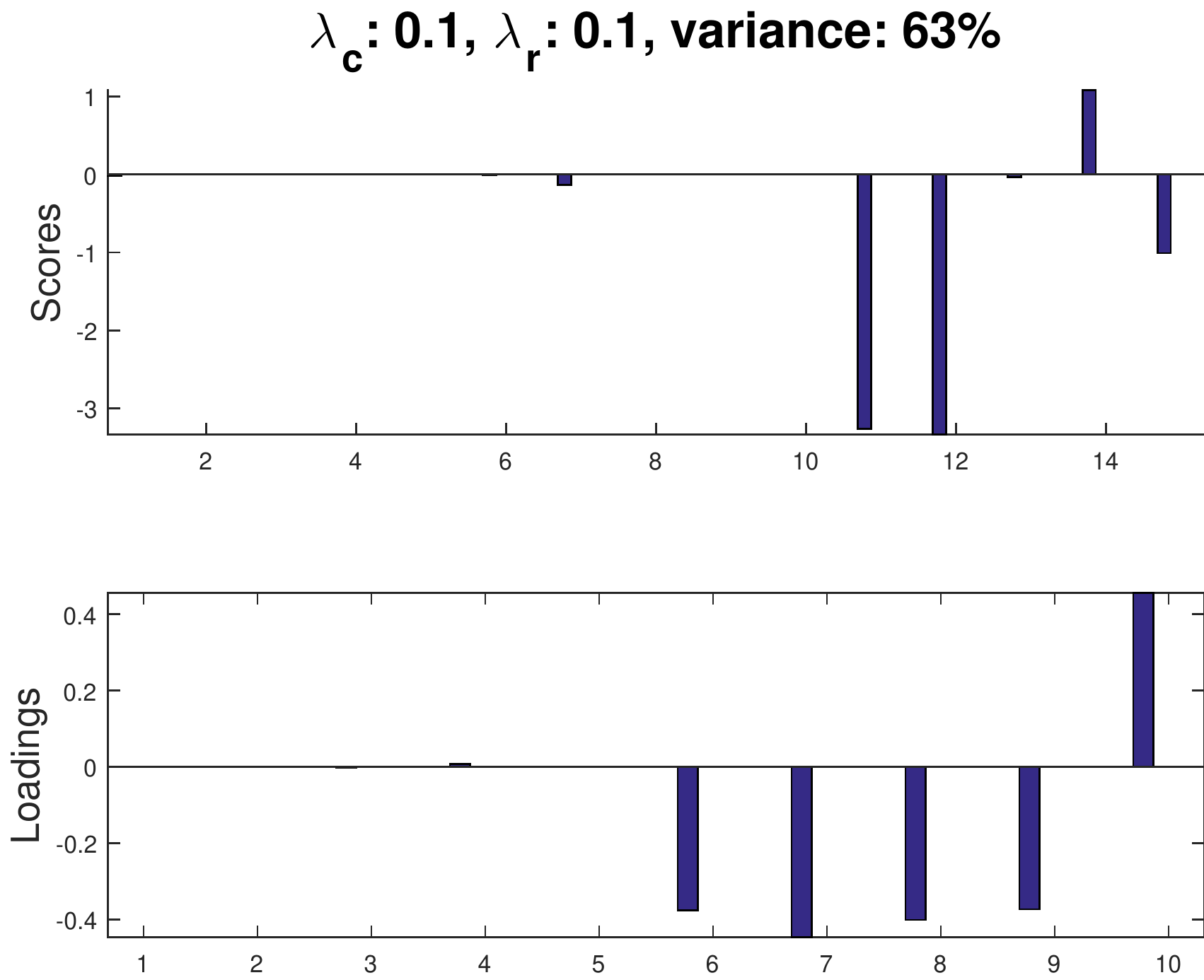}} 
	\hspace{1cm}
	\subfigure[2-component XCAN]{\includegraphics[width=0.4\textwidth]{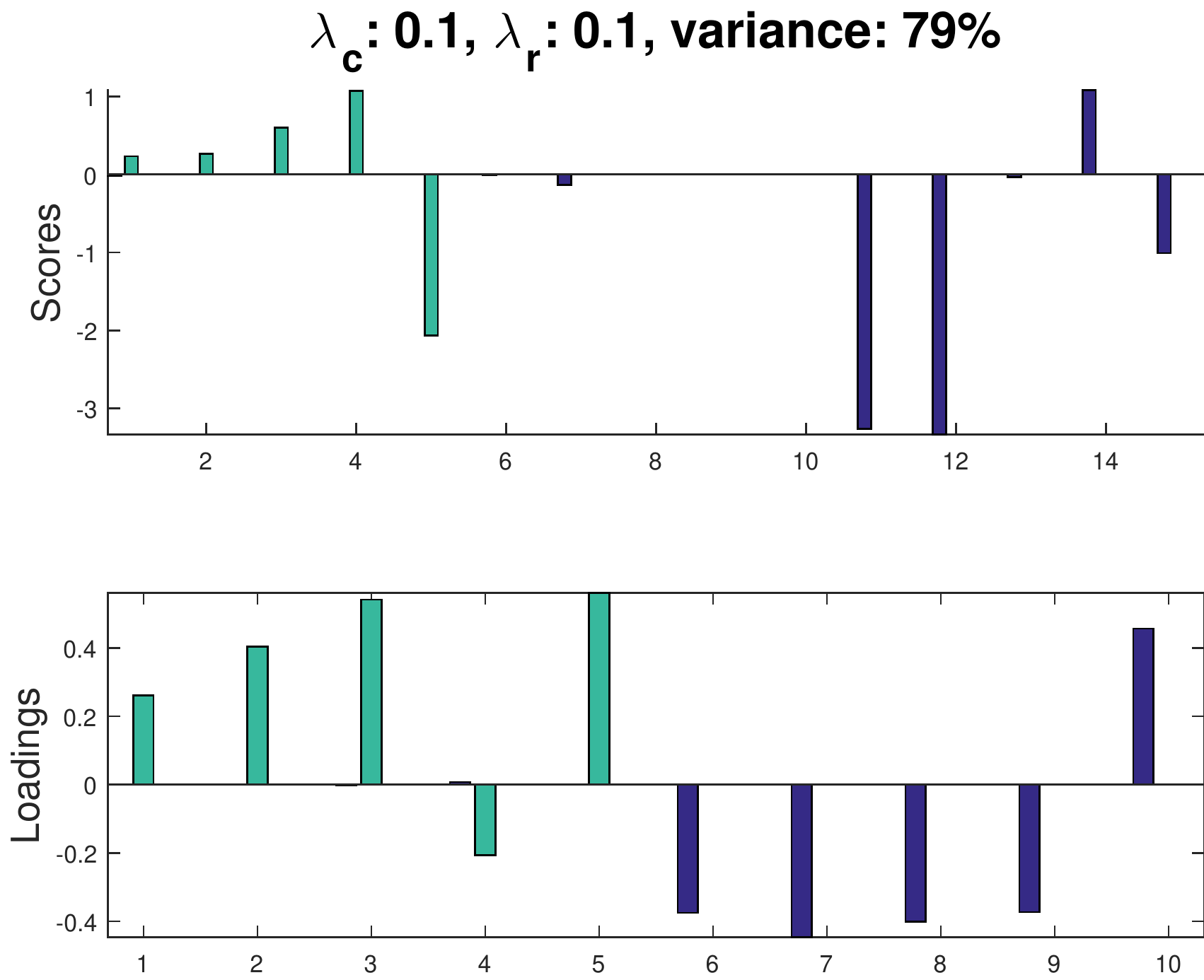}}
	\caption{XCAN models for the simulated data (using one and two components).}
	\label{fig:Sim2b}
\end{figure*}

In order to see the performance of XCAN with different number of components, we have also compared the models with one, two, and three XCs in the example, with the structural penalties activated in loadings and scores. Results can be inspected by comparing Figures \ref{fig:Sim2b} and \ref{fig:Sim2}(d). We observe that the XCAN model of the simulated data is very stable.

	\begin{figure*}
	\centering
	\subfigure[]{\includegraphics[width=0.4\textwidth]{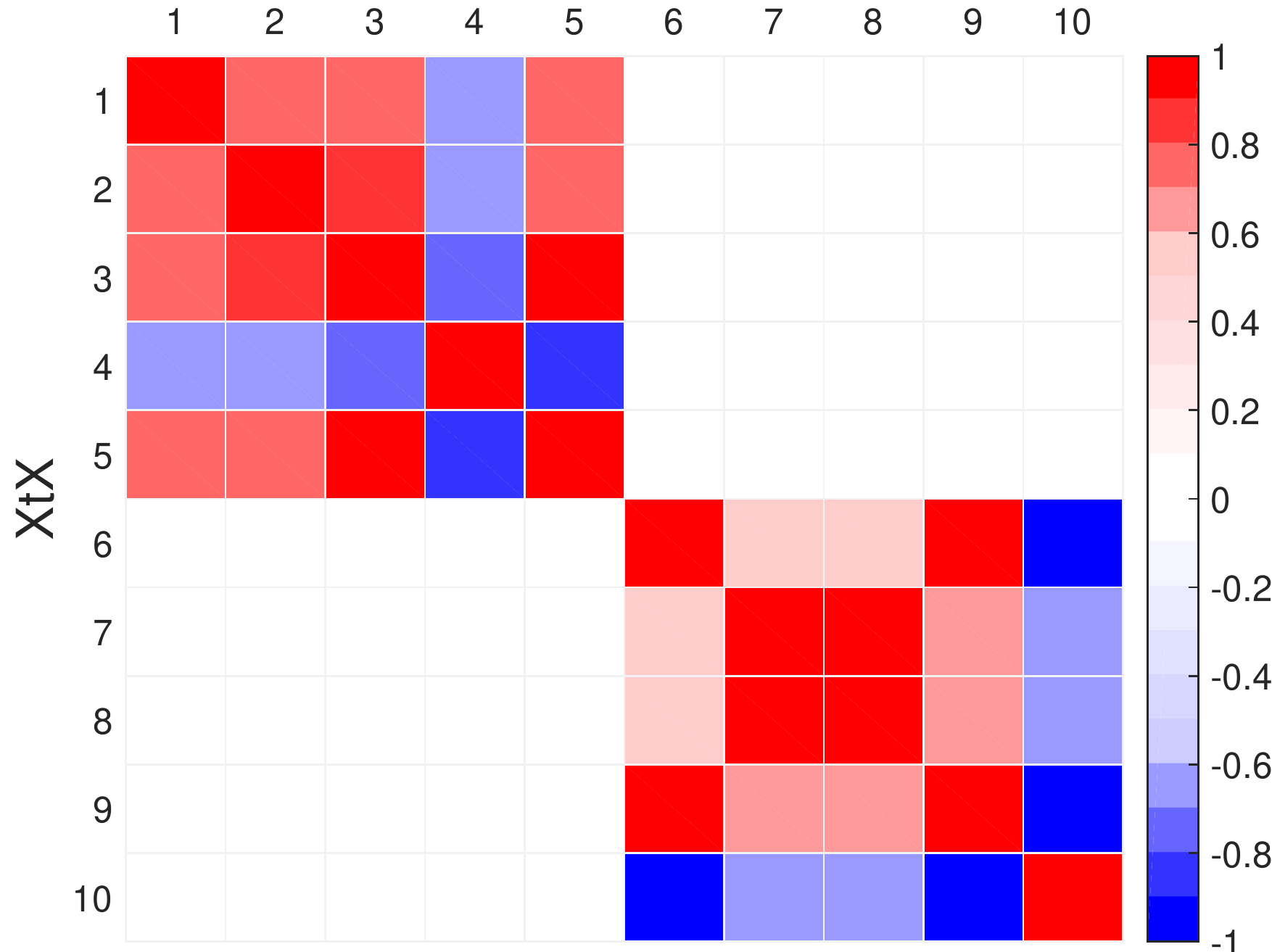}}
	\hspace{1cm}
	\subfigure[]{\includegraphics[width=0.4\textwidth]{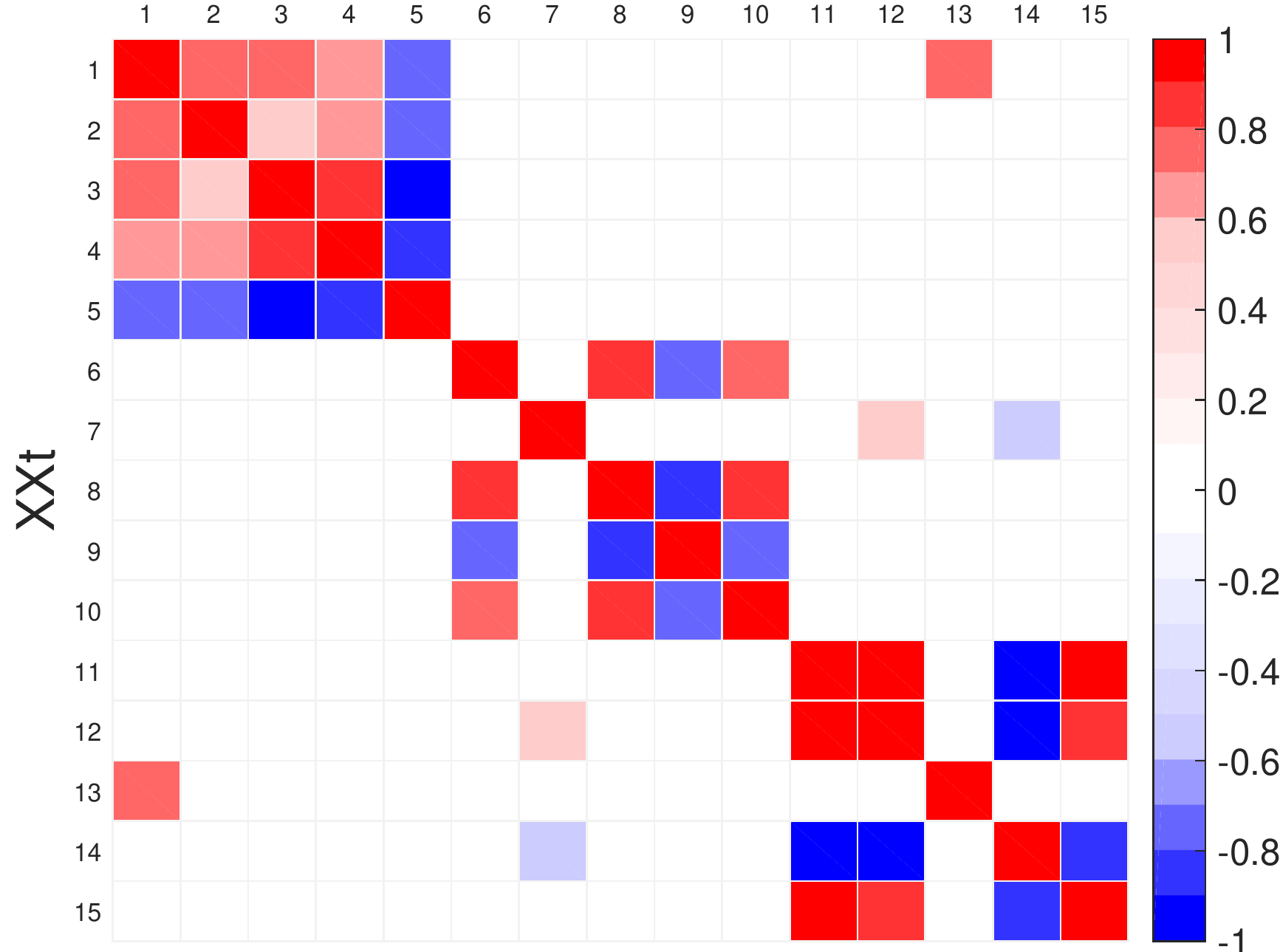}}
	\caption{Cross-product matrices to impose structural penalties in XCAN in the simulated data after thresholding (0.5): (a) $\mathbf{XtX}$ and (b) $\mathbf{XXt}$.}
	\label{fig:Sim2thr}
\end{figure*}

\begin{figure*}
	\centering \subfigure[Non-thresholded]{\includegraphics[width=0.4\textwidth]{Figures/ej_xpca2_res_xp4.eps}}
		\hspace{1cm}
		\subfigure[Thresholded]{\includegraphics[width=0.4\textwidth]{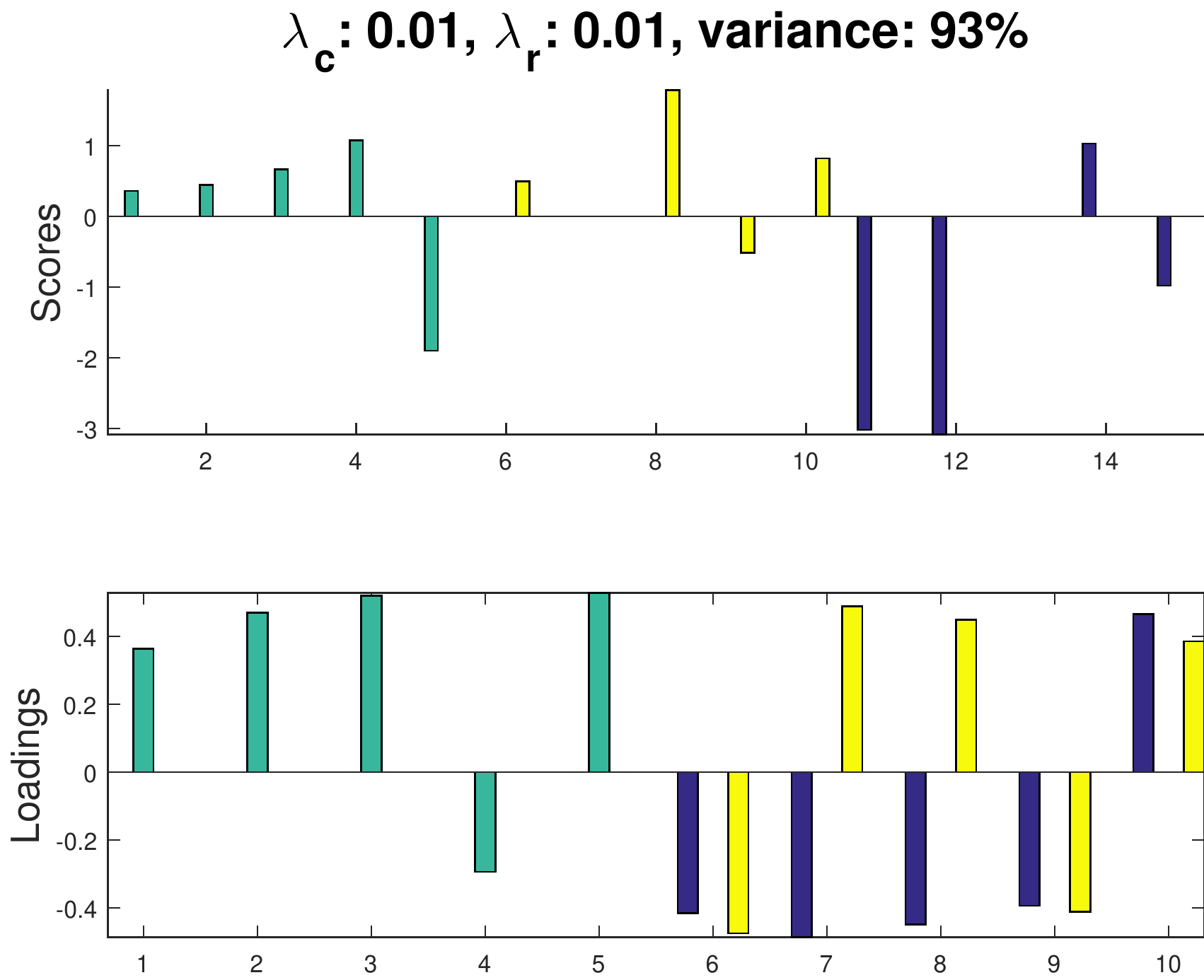}}
		\caption{XCAN models for the simulated data.}
		\label{fig:Sim1thr}
	\end{figure*}

		As discussed before, we can apply thresholding in XCAN. If we set all values in  $\mathbf{XtX}$ and $\mathbf{XXt}$ with magnitude less than 0.5, i.e., within the interval $(-0.5, 0.5)$, to zero, we obtain the matrices in Figure \ref{fig:Sim2thr}, where most of the spurious correlations are discarded. Figure 	\ref{fig:Sim1thr} compares 3-component XCAN models with and without thresholding in the cross-product matrices. We can see that thresholding can be effective in terms of imposing sparsity. In the current example, we can achieve a sparser model using lower values of the meta-parameters and capture higher variance.

	\subsection{Animal Data}
	
	This case study makes use of a toy data set created to illustrate the co-clustering algorithm in \cite{coclustering}. The data contains 34 observations, most of them representing animals, each with 17 attributes, including binary and continuous attributes. {The data values are} non-negative. Binary attributes are used to describe if an animal has (1) or does not have (0) a specific feature. For instance, in the feature `Carnivore', `Lion' contains a 1 and `Cow' a 0.
	
\begin{figure*}
		\centering
		\subfigure[]{\includegraphics[width=0.45\textwidth]{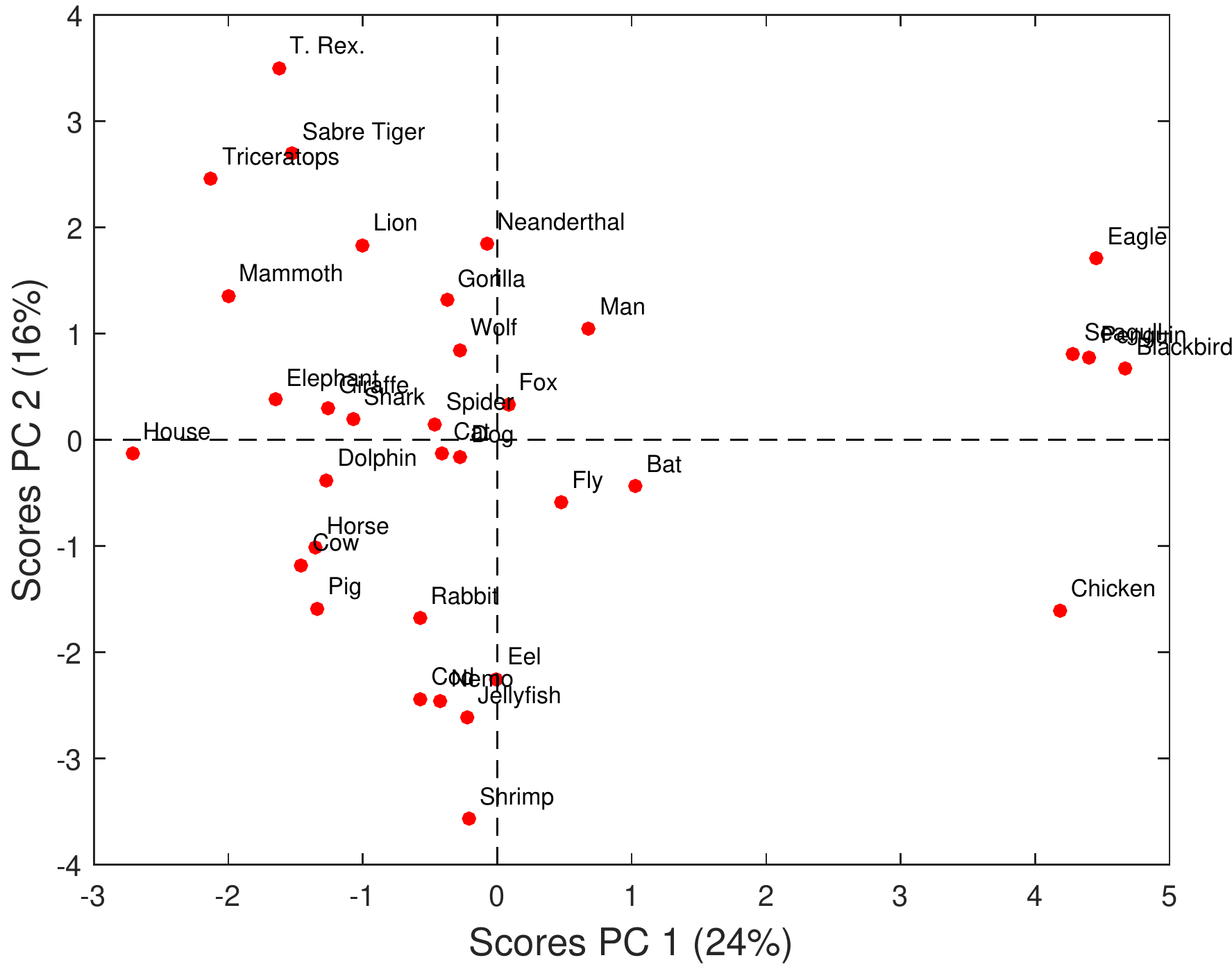}}  \hfill
		\subfigure[]{\includegraphics[width=0.45\textwidth]{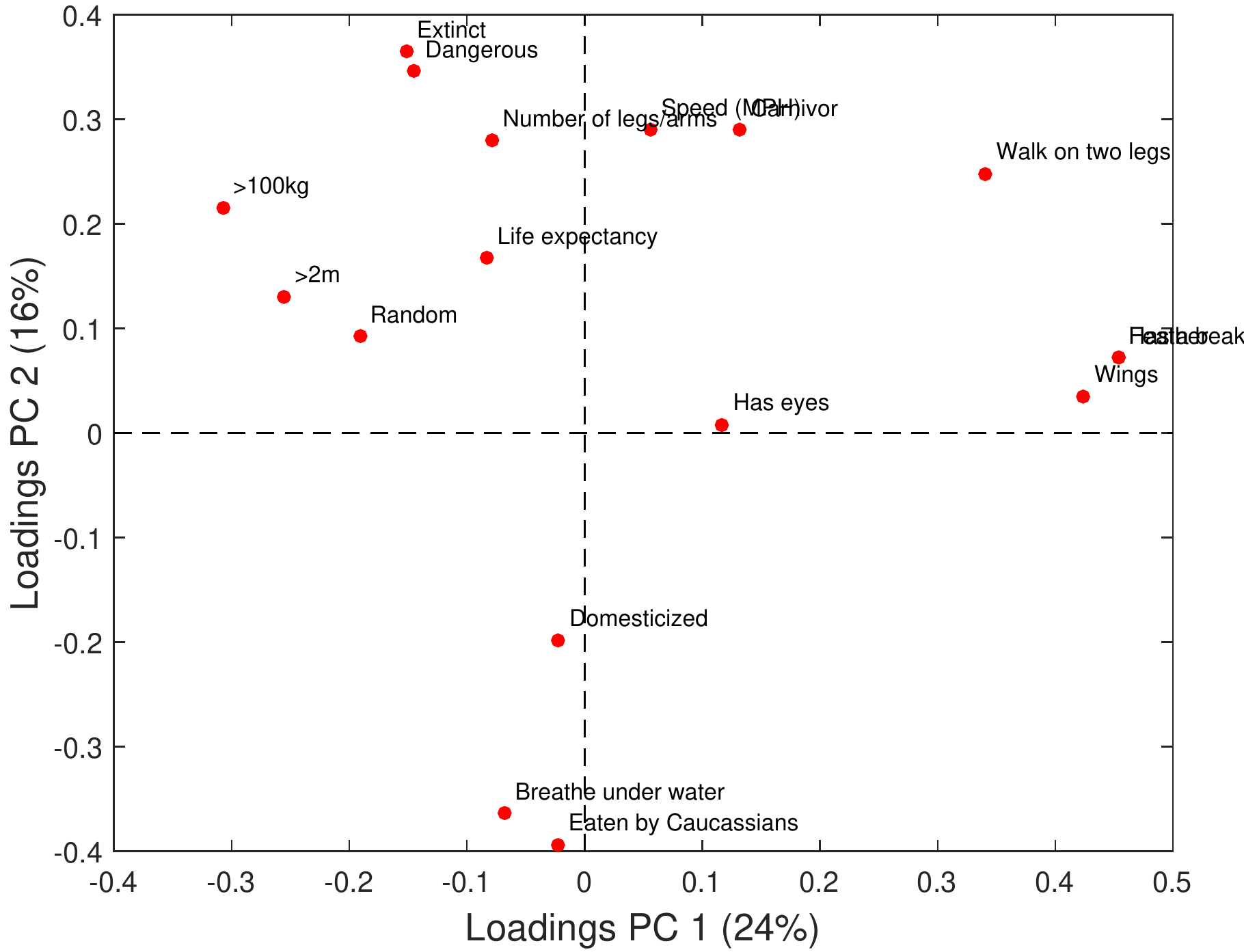}} 
		\caption{(a) Score plot, and (b) loading plot for the Animal Data.}
		\label{fig:An}
	\end{figure*}

	Scores and loadings plots for the first two PCs of PCA are shown in Figure \ref{fig:An}, for auto-scaled data, that is, mean-centered across the animals mode and scaled to unit variance within the attributes. The first PC is dominated by birds, e.g., `Eagle', `Blackbird' or `Chicken', which share features like `Wings', `Feathers' or `Has a beak'. Birds are small, and for this reason they are located opposite to big animals, in particular, to the `House', which was included in the data set as an outlier. The second PC contrasts `Dangerous' and `Extinct' animals, most notably the `T. Rex', with those `Domesticized' and `Eaten by Caucasians', which are also correlated with `Breathe under water'. Both components are interpretable, but still complex in the sense that they mix different concepts: birds + small vs big, dangerous vs domesticated + fish. As such, this data set is a perfect example of the limitations of PCA reported in the introduction. 
	
\begin{figure*}
	\centering
	\subfigure[]{\includegraphics[width=0.4\textwidth]{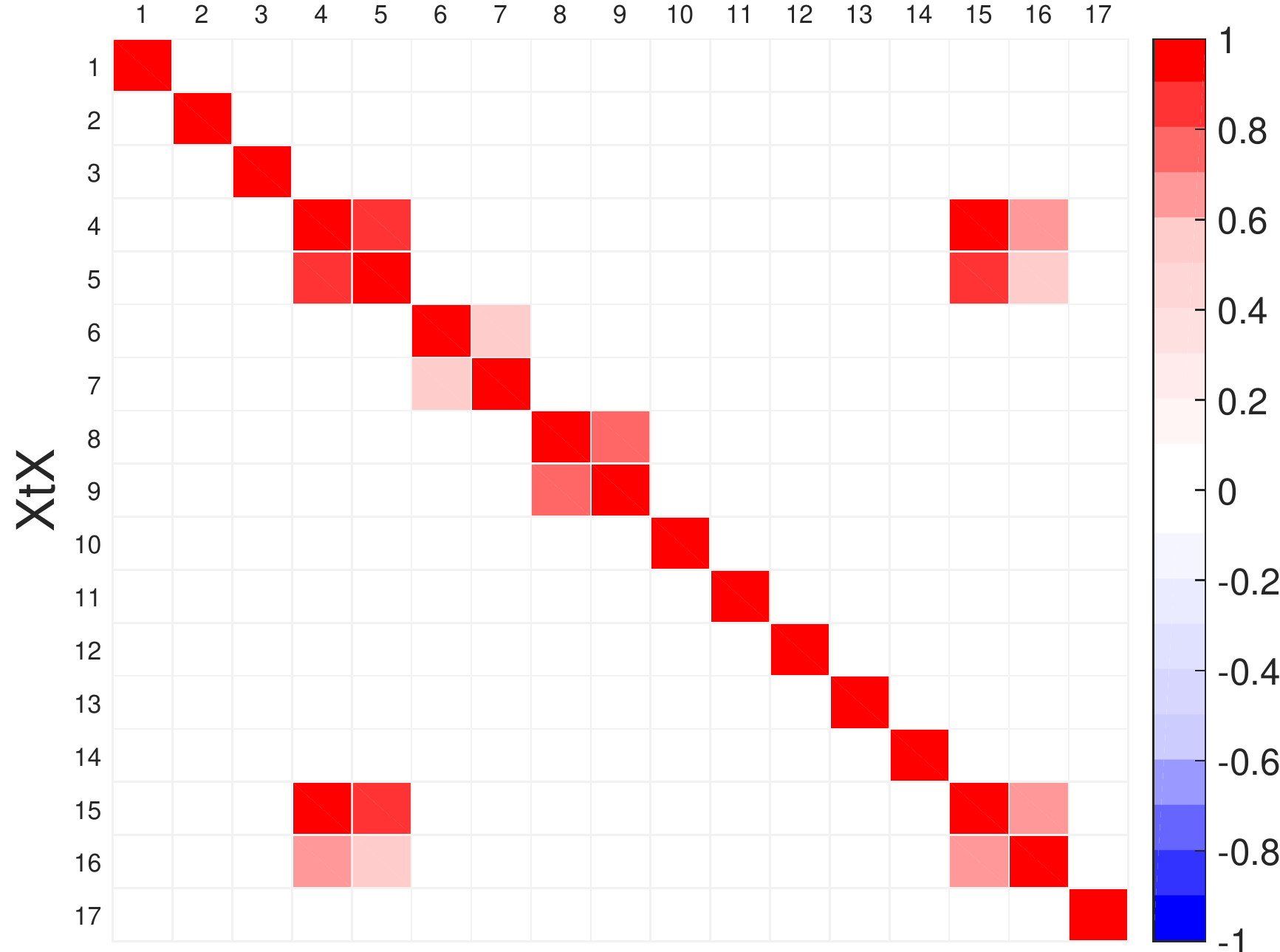}} \hfill
	\subfigure[]{\includegraphics[width=0.4\textwidth]{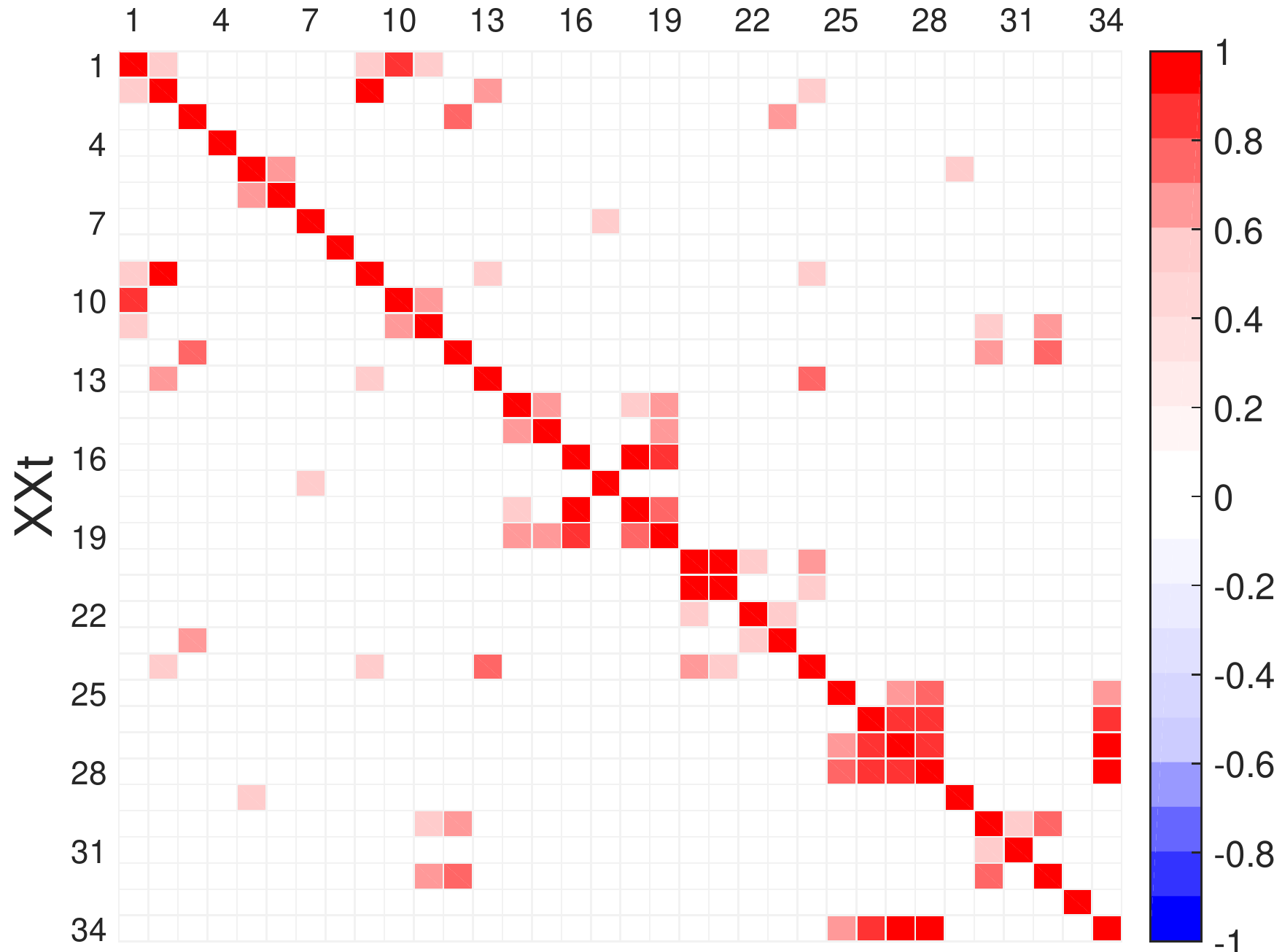}}
	\caption{Thresholded cross-product matrices to impose structural penalties in XCAN for the Animal Data: (a) $\mathbf{XtX}$ and (b) $\mathbf{XXt}$.}
	\label{fig:AnXX}
\end{figure*}

	To perform the XCAN analysis, we auto-scaled the data. 
	Since the goal is to understand differences among individuals and group them as in (co-)clustering, it makes sense to center the data so that we study the variability among the set of individuals, instead of with respect to some center of coordinates. {Scaling seems to equalize the influence of the variables in the model}. Since we want each component to be a cluster of similar (and not antagonist) individuals, we constrain $\mathbf{XXt}$ to be non-negative. In particular, every entry in $\mathbf{XXt}$ with a value below 0.5 is hard thresholded to 0, i.e., values within the interval $(-\infty,0.5)$ are thresholded. That way, components will be extra sparse and yield only positive (or only negative) scores. In case a component contains negative scores we simply change its sign. For this analysis, we are not concerned with the signs of the loadings, but we still want loadings to be extra sparse, so we threshold the entries of $\mathbf{XtX}$ using the threshold value of 0.5, i.e., setting every entry with a value within the interval $(-0.5,0.5)$ to 0. This allows negative correlations of -0.5 or lower. Resulting cross-product matrices are shown in Figure \ref{fig:AnXX}.

\begin{figure*}
\centering
{\includegraphics[width=0.32\textwidth]{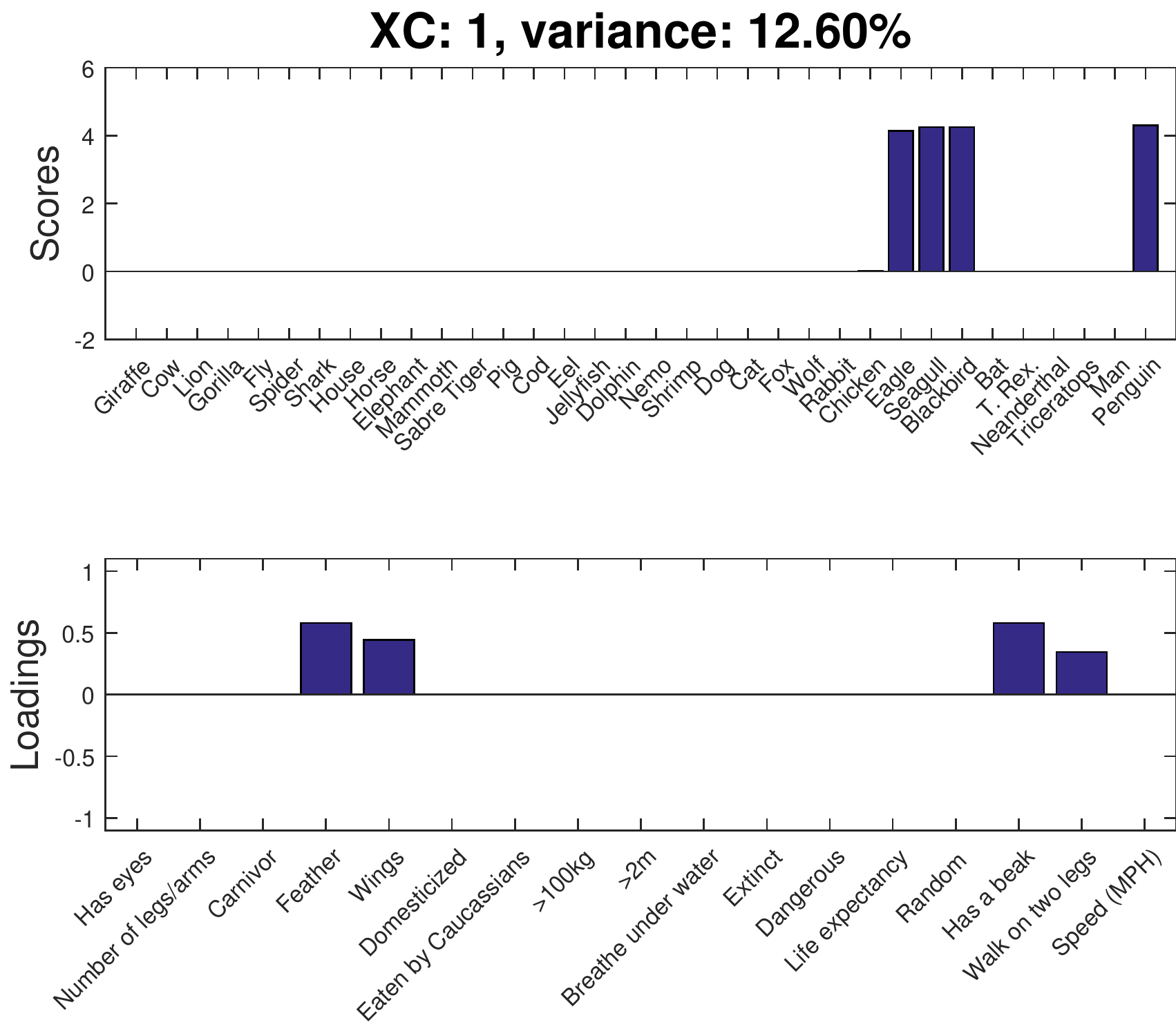}}  \hfill
{\includegraphics[width=0.32\textwidth]{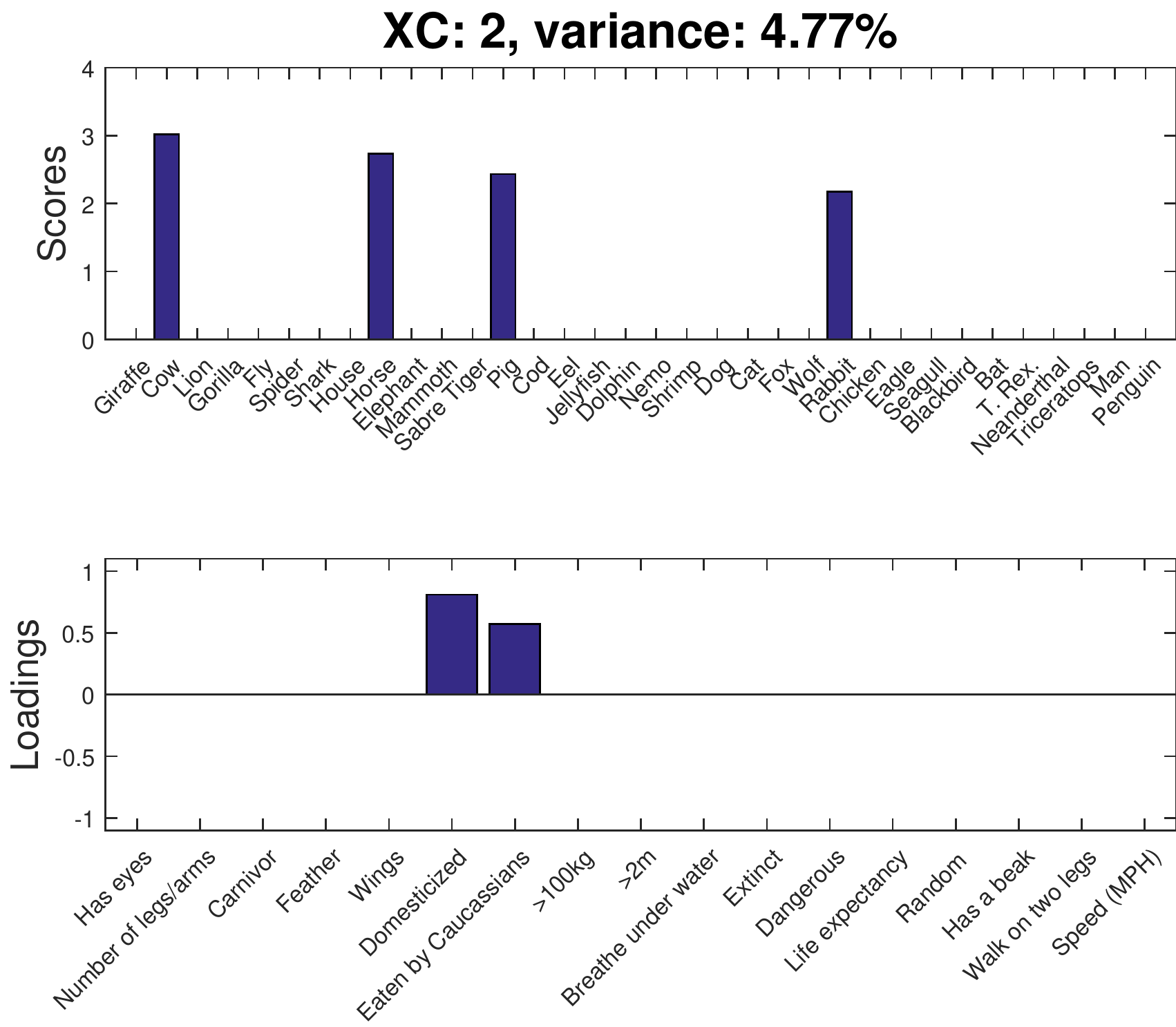}}  \hfill
{\includegraphics[width=0.32\textwidth]{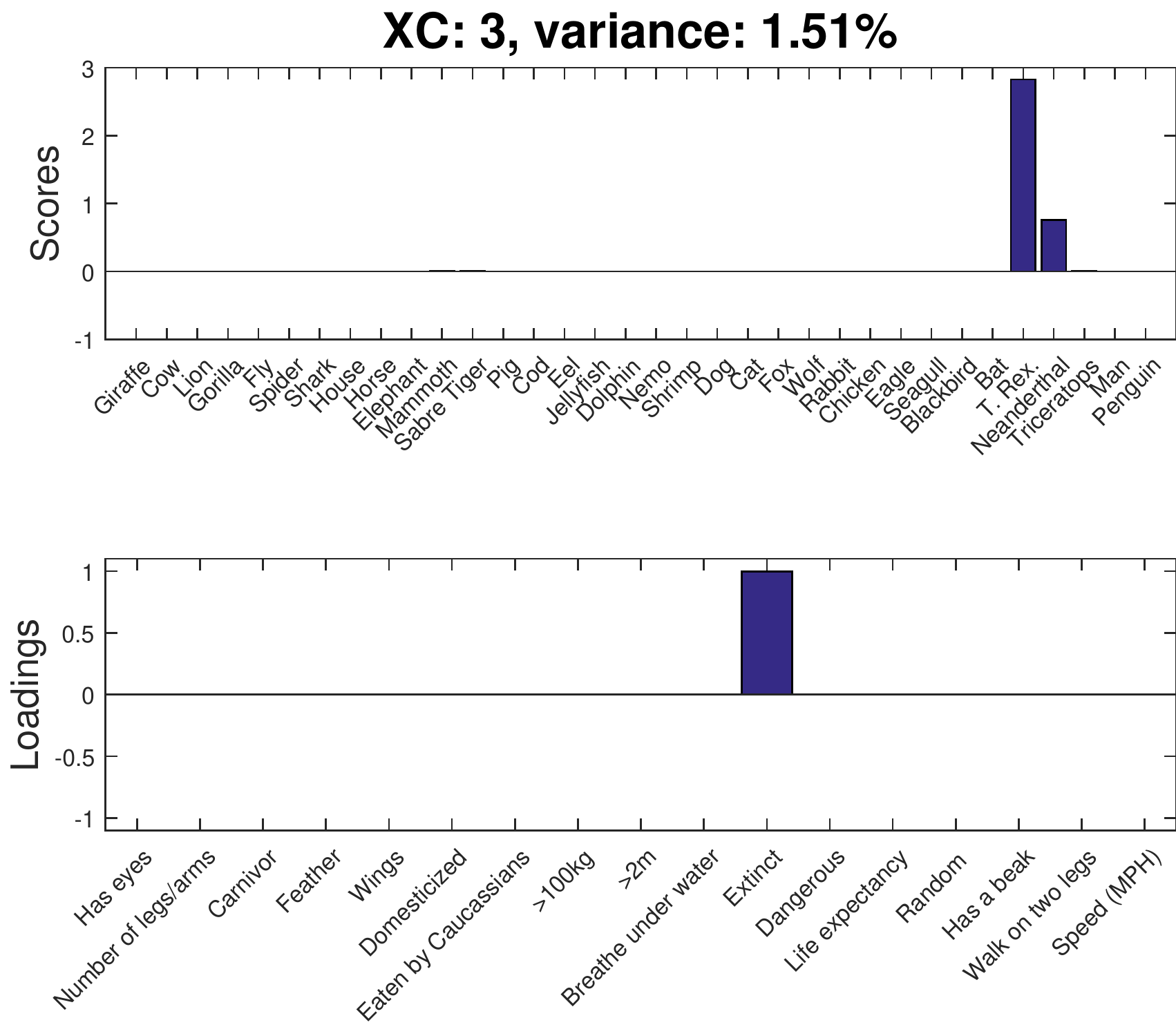}}  \hfill \\
{\includegraphics[width=0.32\textwidth]{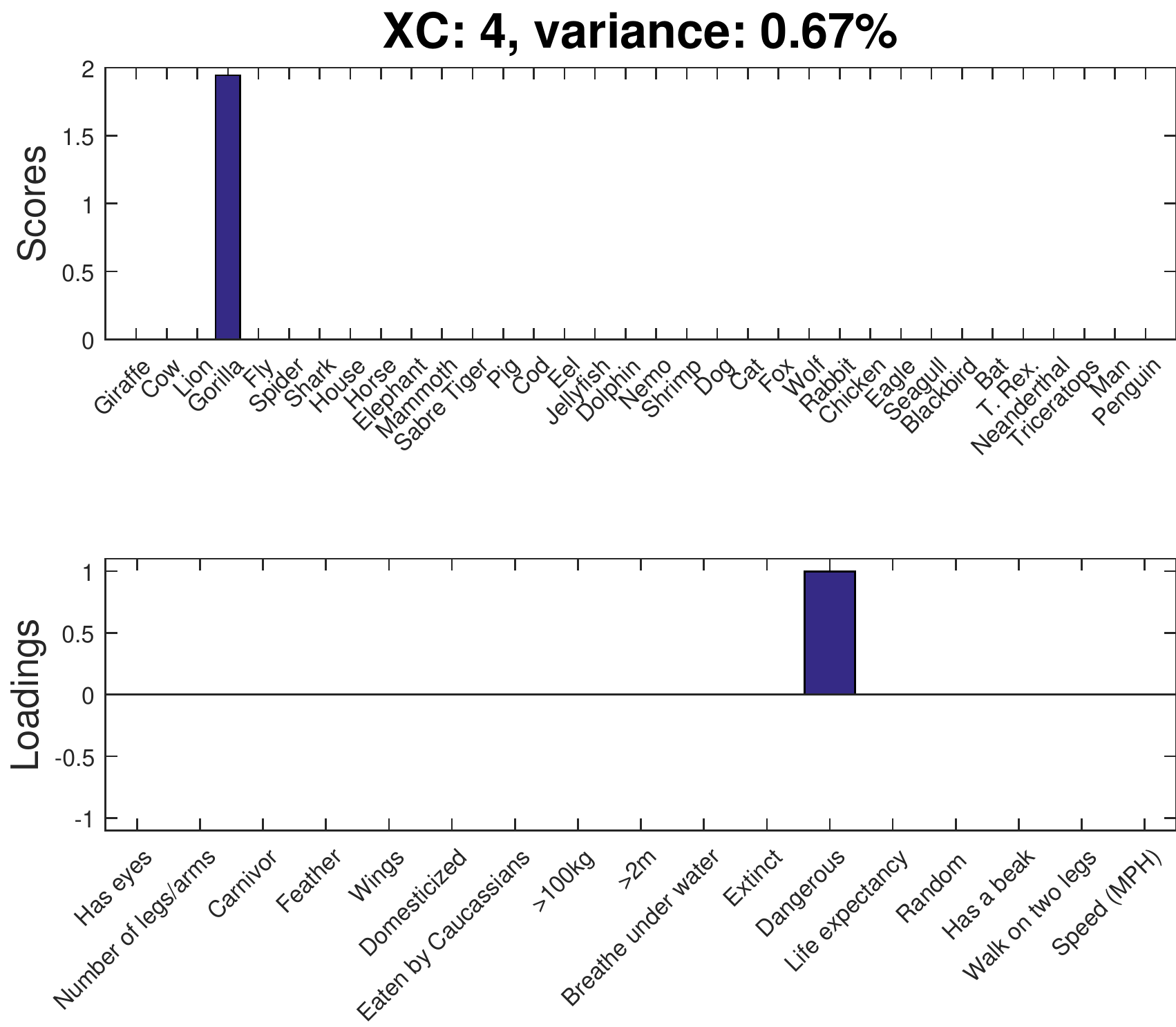}} \hfill
{\includegraphics[width=0.32\textwidth]{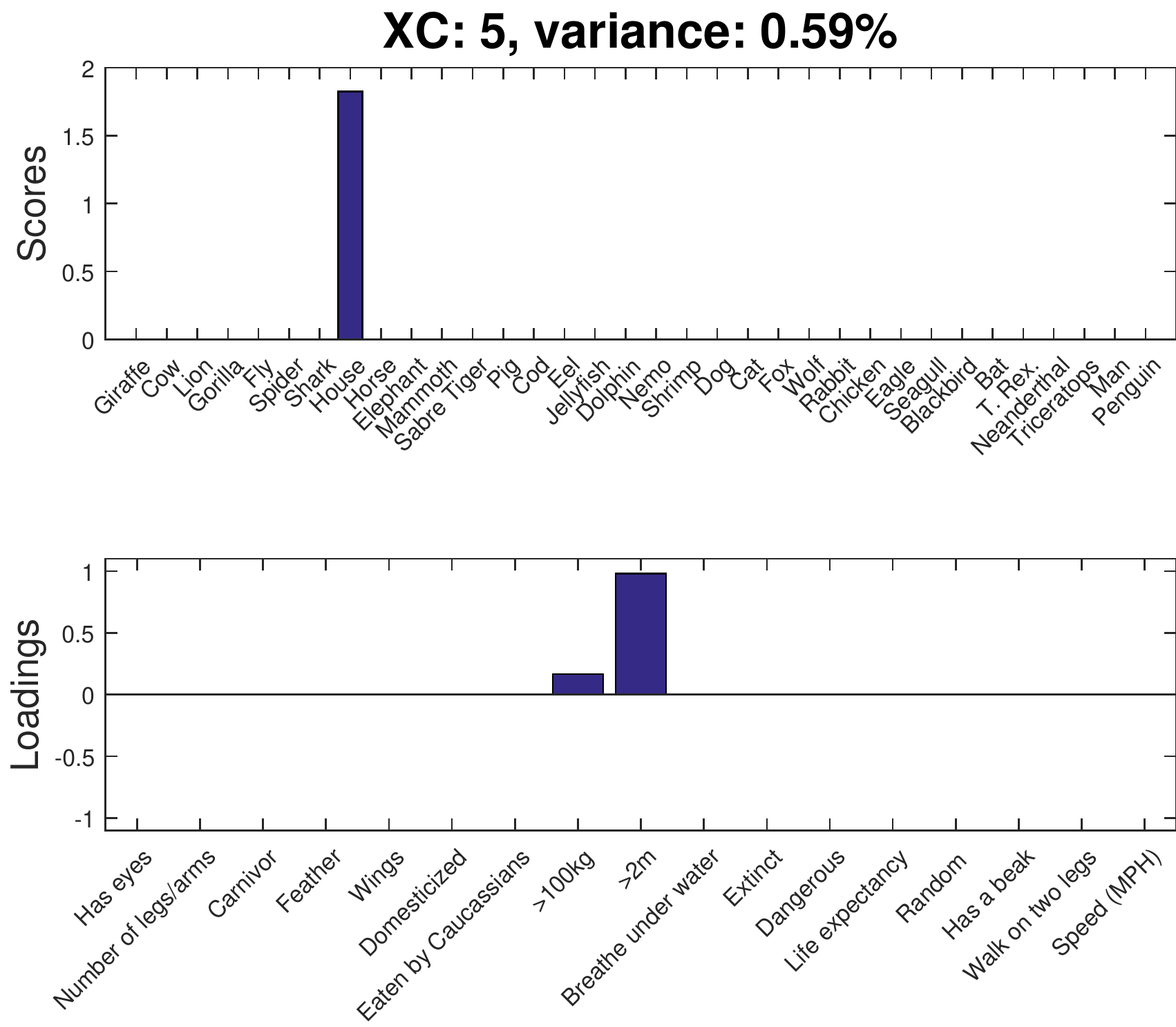}} \hfill
{\includegraphics[width=0.32\textwidth]{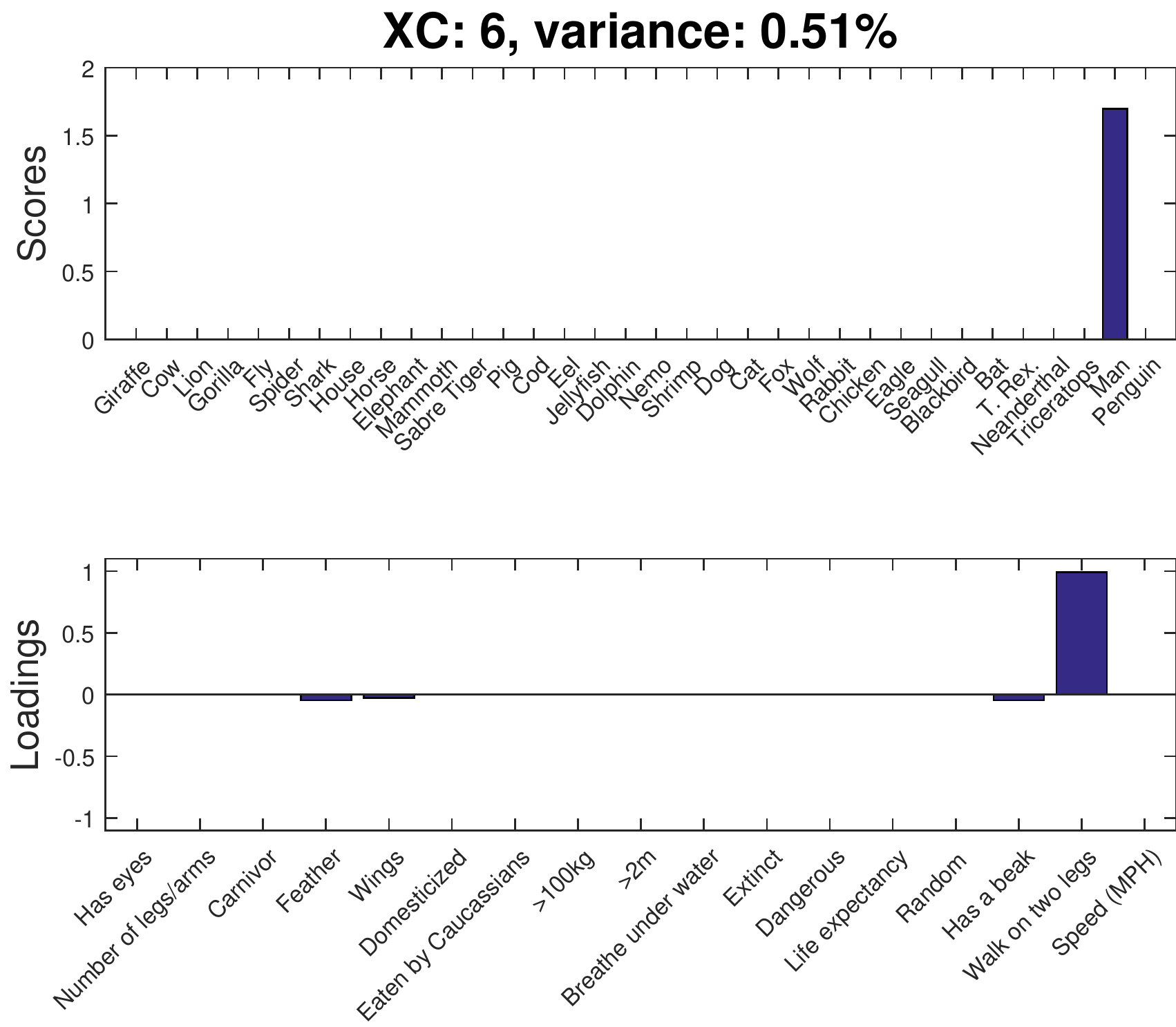}} \\
{\includegraphics[width=0.32\textwidth]{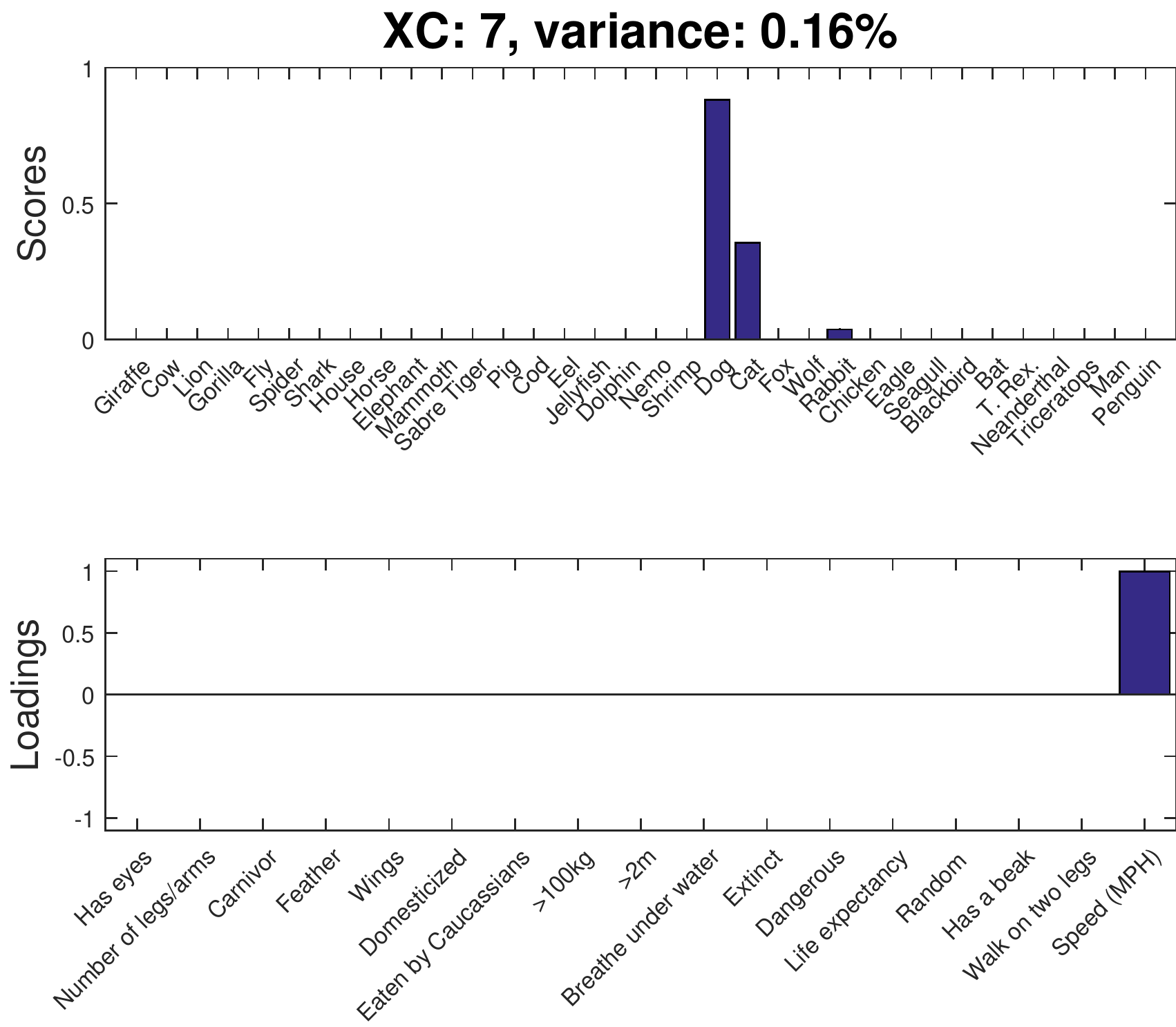}} \hfill
{\includegraphics[width=0.32\textwidth]{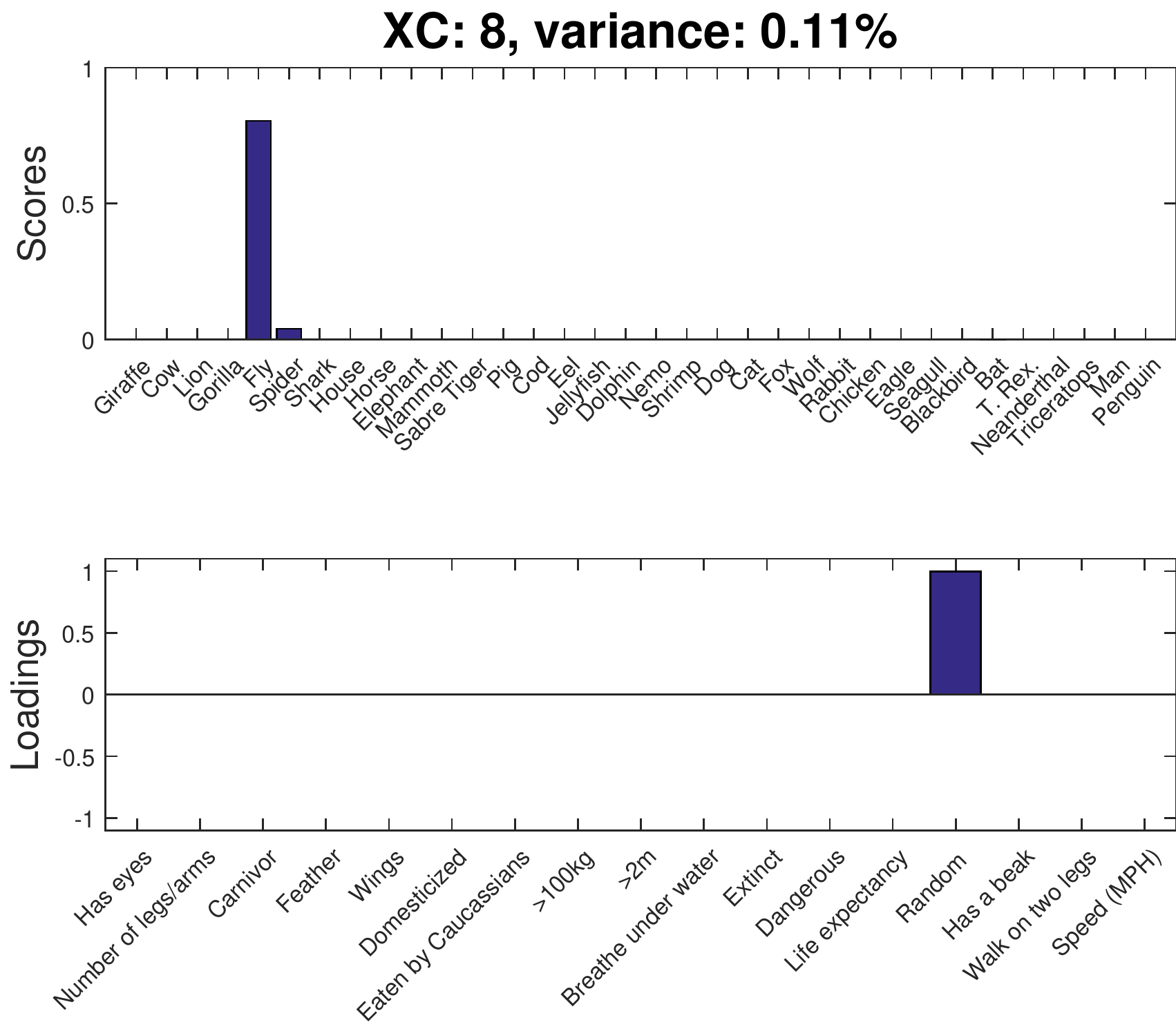}} \hfill
{\includegraphics[width=0.32\textwidth]{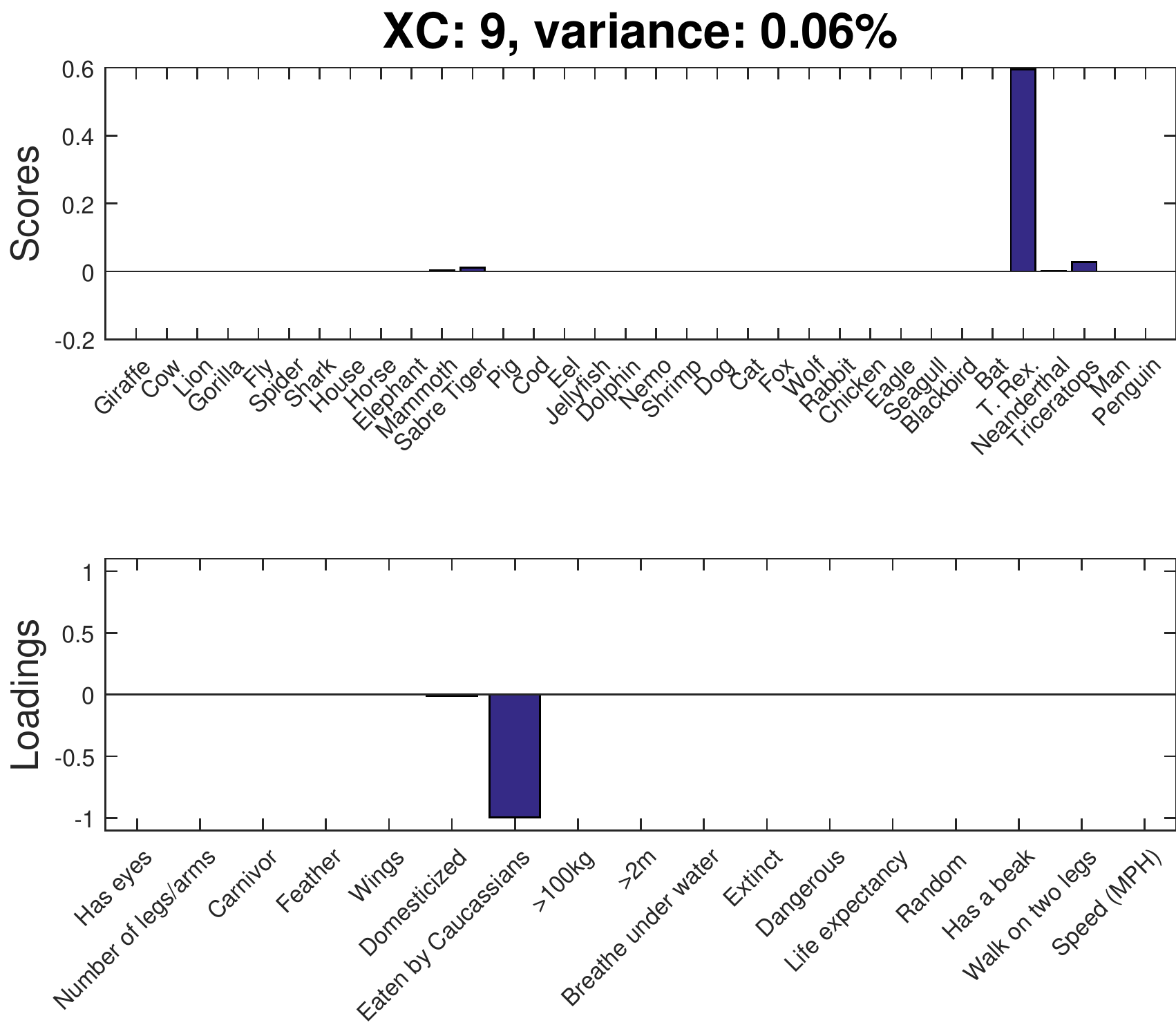}} \\
\caption{The 9-component XCAN model on the Animal data.}
\label{fig:An2}
\end{figure*}

\begin{figure}
	\centering
	\includegraphics[width=0.4\textwidth]{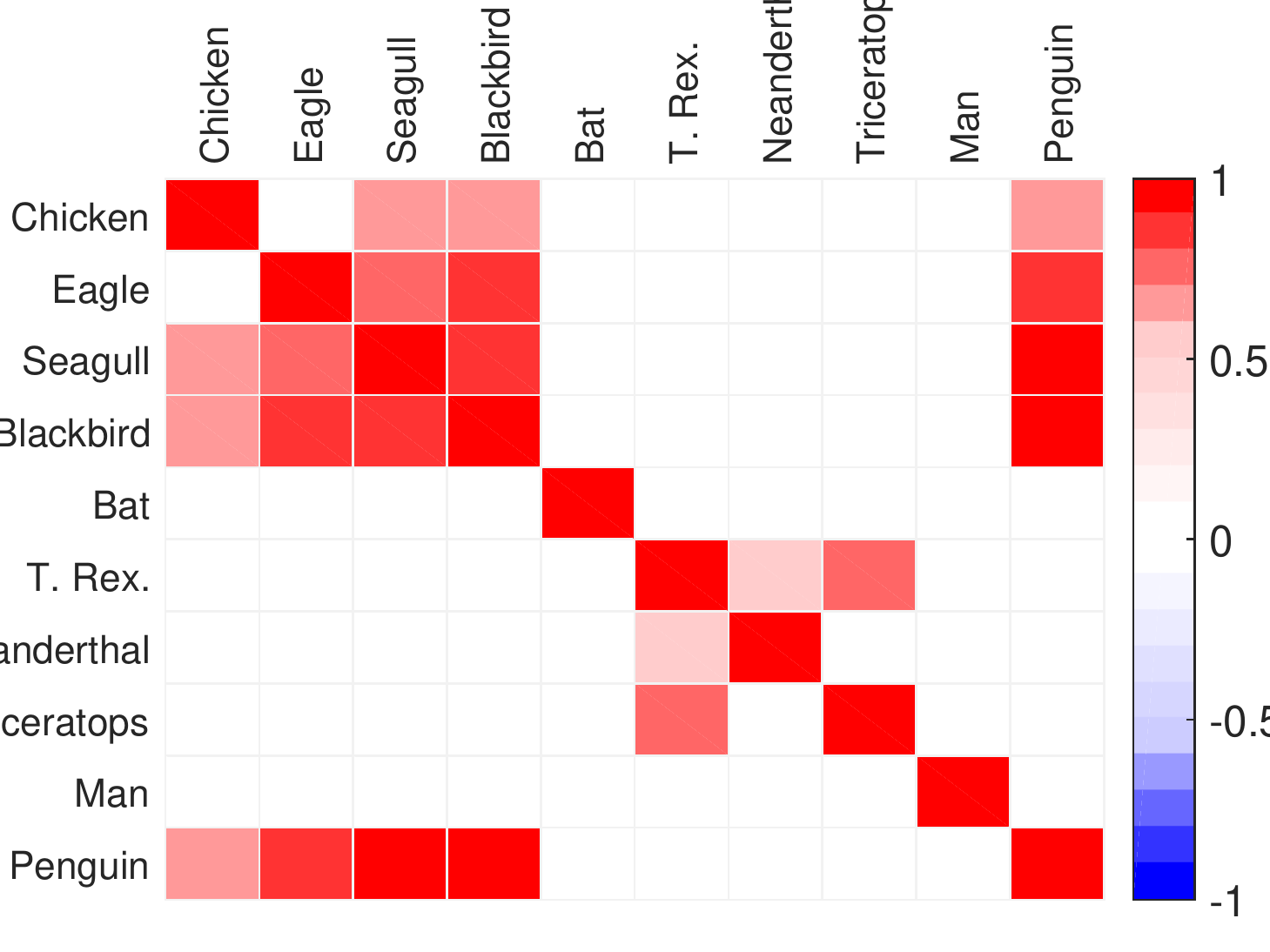}
	\caption{Zoom of matrix $\mathbf{XXt}$.}
	\label{fig:AnXXz}
\end{figure}

The 9-component XCAN model is shown in Figure \ref{fig:An2}. We can see that all XCs are sparse in both scores and loadings, and all scores are non-negative, as expected. The model extracts meaningful components, that can be interpreted as co-clusters. The first component represents wild birds, with their corresponding features. Interestingly, `Chicken' is not included in the component. The reason for that can be seen in Figure \ref{fig:AnXXz}, which represents a zoom of matrix $\mathbf{XXt}$: `Chicken' and `Eagle', even though they share the features in the first component, they are completely different animals according to matrix $\mathbf{XXt}$, showing a correlation close to 0. Thus, XCAN does not place them in the same component. Since `Eagle' is strongly correlated with the other birds, the model selects the former among the group of scores of the first component. The second component represents domesticated animals consumed for food. The third component focuses on the feature of extinct, but only two (`T.Rex' and `Neanderthal') out of the five extinct animals (including `Mamouth', `Sabre Tiger' and `Triceratops') show relevant scores. {The reason for that is similar to before and also apparent in Figure \ref{fig:AnXXz}. 'Neanderthal' and 'Triceratops' are uncorrelated, and for this reason they cannot share the same component.} The rest of the components are more or less self-explanatory.

The combination of sparsity and grouping in components 1 and 3 is coherent with GPCA, but not with PCA/SPCA models, where one would expect that all individuals with feathers and wings, or those extinct, should be placed in the same component. If, for some reason, we are interested only in sparsity but not on the grouping characteristic of XCAN, then traditional sparse methods should be used for the analysis. With XCAN, we explicitly avoid, within a component, objects or features that are not associated in the cross-product matrices.



In conclusion, we observe that the XCAN model performs exactly as expected: components are sparse and reflect accurately the structure in the cross-product matrices. 

\subsection{Real Data}

\subsubsection{Vast Challenge for Cybersecurity}

The increase in the number of cybersecurity incidents, coupled with the shortage of specialized professionals, has created a need for efficient data analysis tools to support the detection, triaging and analysis of incidents \cite{SIEMMarket}. In particular, anomaly-based Intrusion Detection Systems (IDS) \cite{Garcia09} are fundamental resources to unveil new attack strategies. A large number of intrusion detection approaches based on PCA have been proposed in the last two decades  \cite{Lakhina2004a,Delimargas2014,Callegari2014,Aiello2016,MSNM2016}.	In a recent paper \cite{theron_vizsec_2017}, the multivariate detection approach was extended to GPCA. 

With GPCA, we can identify anomalies in the data following a straightforward approach. 
components simplify the interpretation by reducing the number of variables to examine. Components can be interpreted one at a time, following a very similar workflow to the one security analysts use in traditional software tools. Since XCAN inherits the features of GPCA, we will explore its performance in the cybersecurity domain. In comparison to GPCA, XCAN can also impose sparsity in the rows, which in the cybersecurity domain typically corresponds to time-resolved data. This is useful to speed up the analysis of an incident, so that the analyst can focus on a few points in time to troubleshoot the problem.

\begin{figure*}
	\centering
	\subfigure[]{\includegraphics[width=0.4\textwidth]{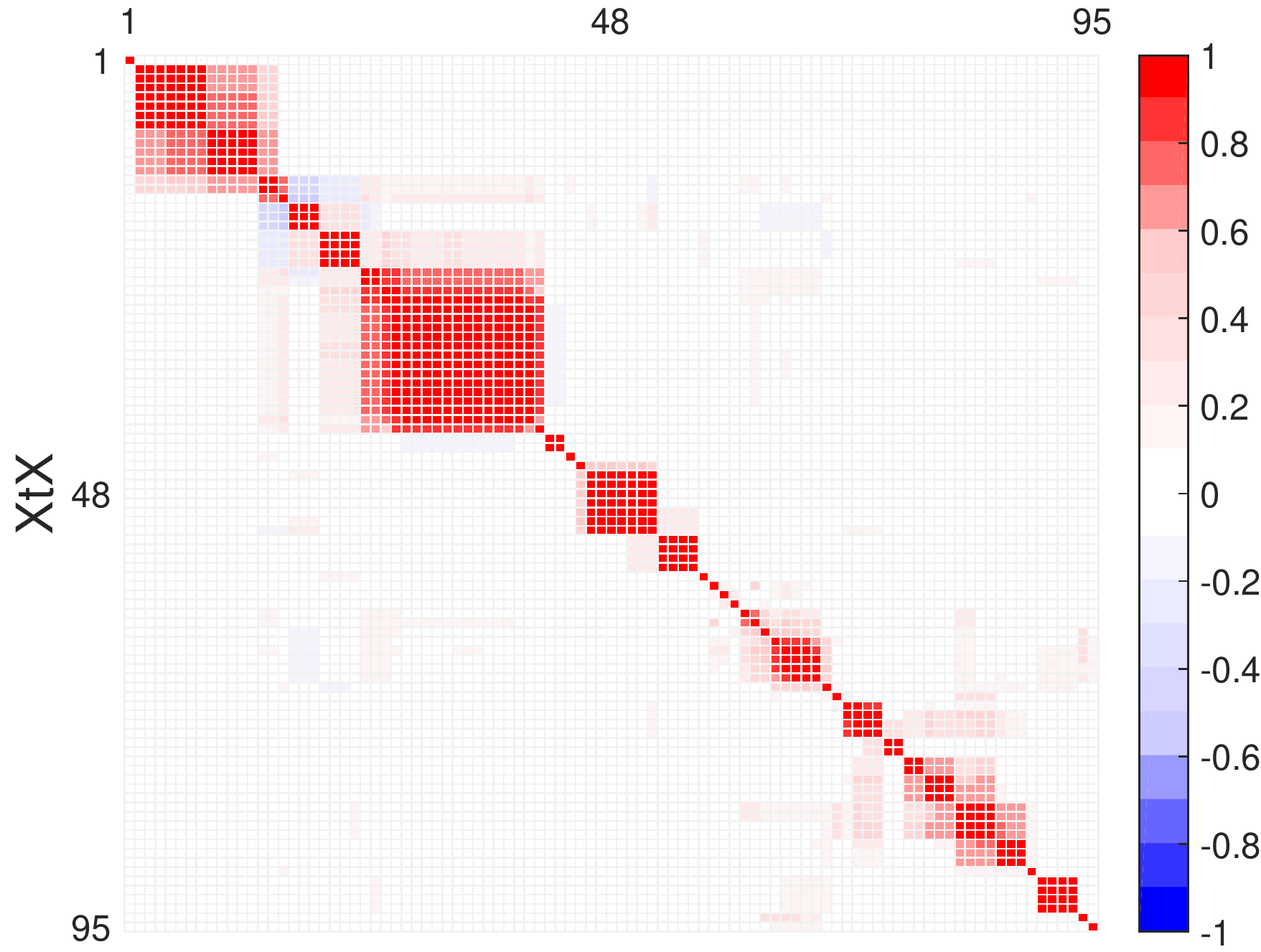}} \hfill
	\subfigure[]{\includegraphics[width=0.45\textwidth]{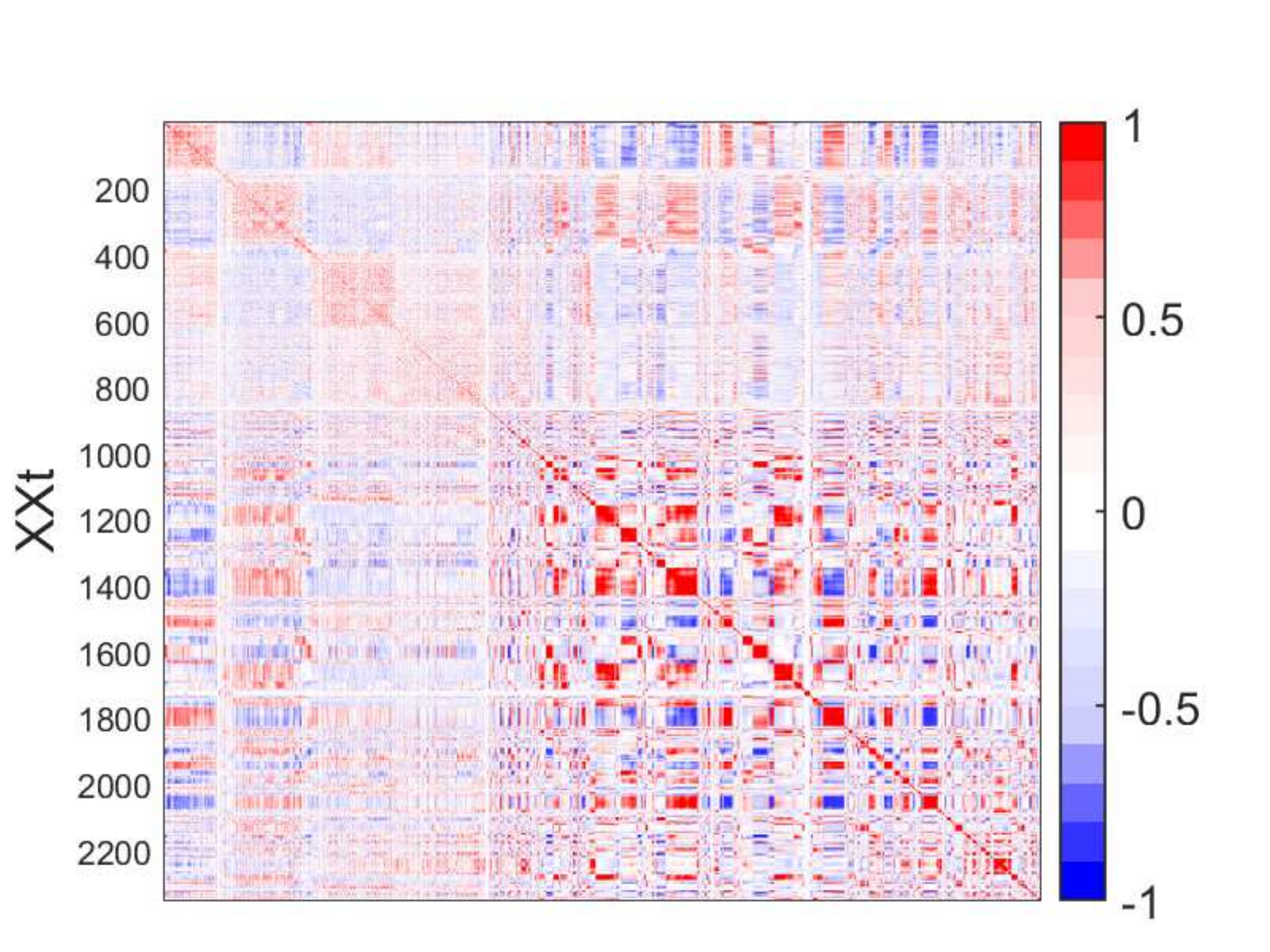}}
	\caption{Cross-product matrices to impose structural penalties in XCAN for the Vast Data: (a) $\mathbf{XtX}$ and (b) $\mathbf{XXt}$.}
	\label{fig:vaXX}
\end{figure*}

The data of the present case study comes from the VAST 2012 2nd mini-challenge \cite{VAST2012} and was captured in a corporate network during a time frame of two days. During that time period, a botnet compromised the network, causing performance problems and the emergence of spyware. The raw data is parsed in a total of 265 features at 1 minute intervals, yielding a 2345 x 265 data matrix. More details can be found in \cite{MBDA}.  Cross-product matrices following eq. (\ref{XtX}) and eq. (\ref{XXt}) are shown in Figure 
\ref{fig:vaXX}. The plots illustrate that applying a group-wise constrained model is reasonable, both in observations and features, since the cross-product matrices contain squares of high correlation.

We used the following strategy for the application of XCAN in this problem domain. Cross-product matrices were thresholded to positive values above 0.7 ($\mathbf{XtX}$) and 0.3 ($\mathbf{XXt}$). Data was auto-scaled for similar reasons as in the previous case study. The mean centering is useful to detect anomalies as deviations from the mean. The scaling normalizes the relevance of the variables in the model.

\begin{figure*}
	\centering 
	{\includegraphics[width=0.3\textwidth]{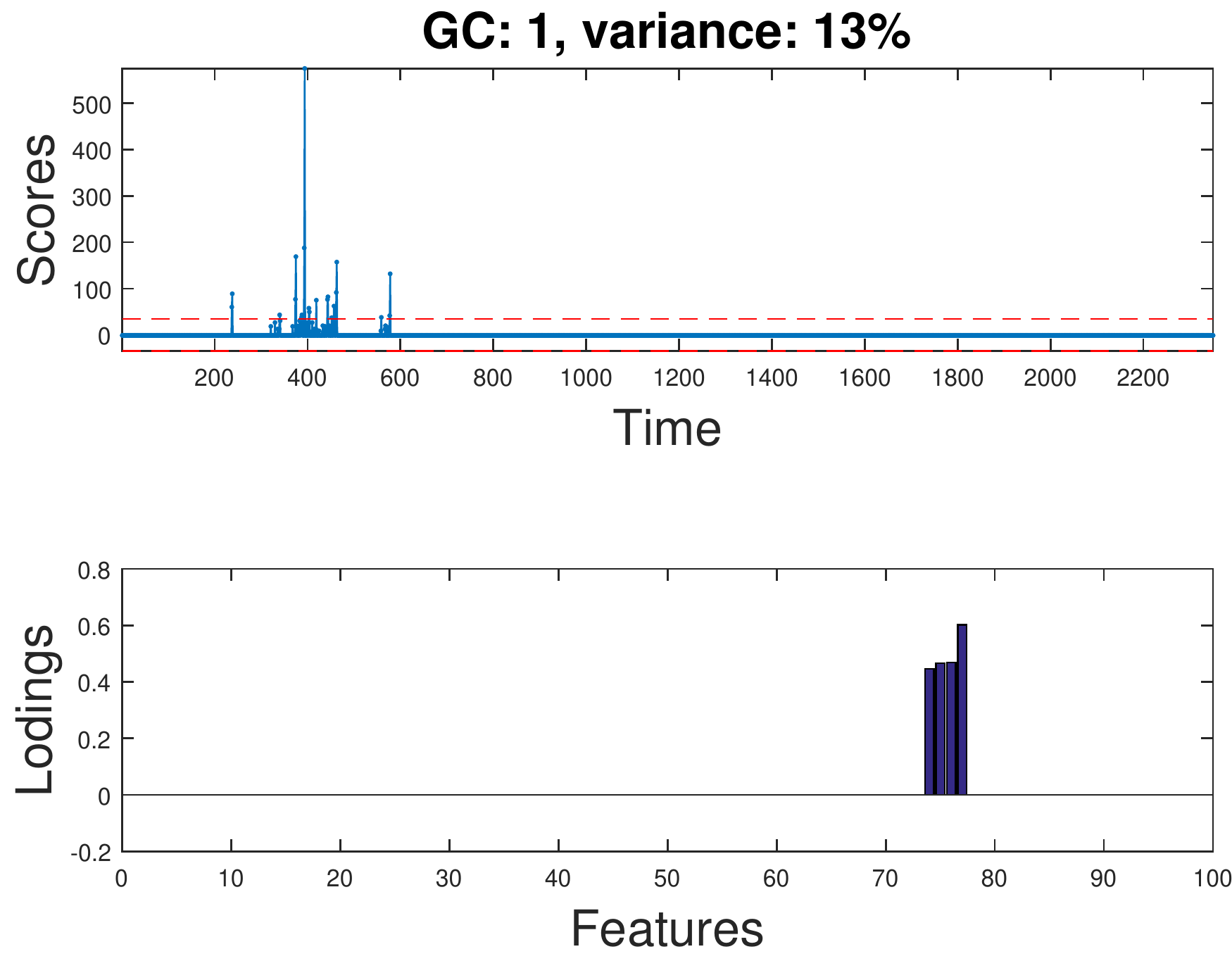}} \hfill
	{\includegraphics[width=0.3\textwidth]{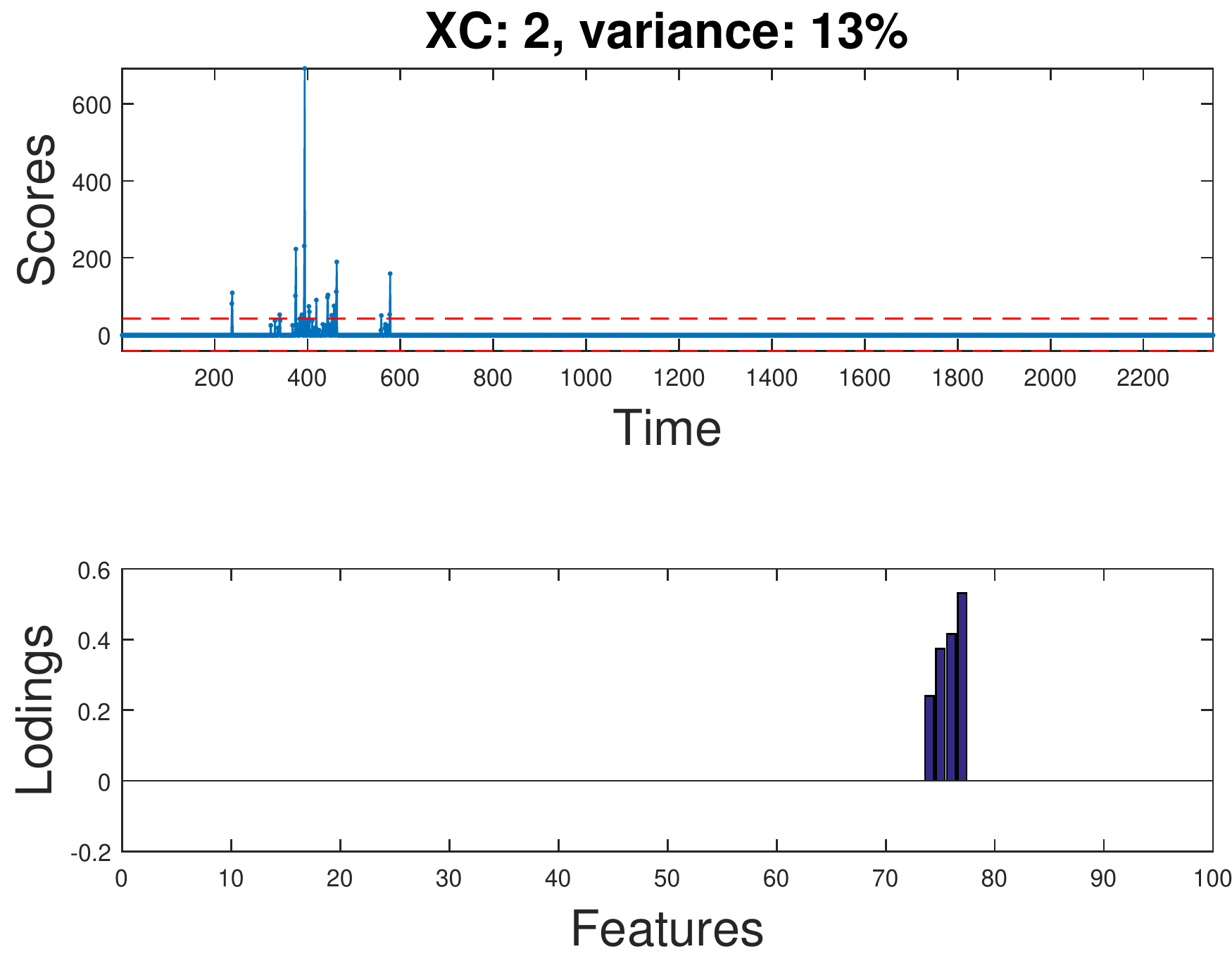}} \hfill
	{\includegraphics[width=0.3\textwidth]{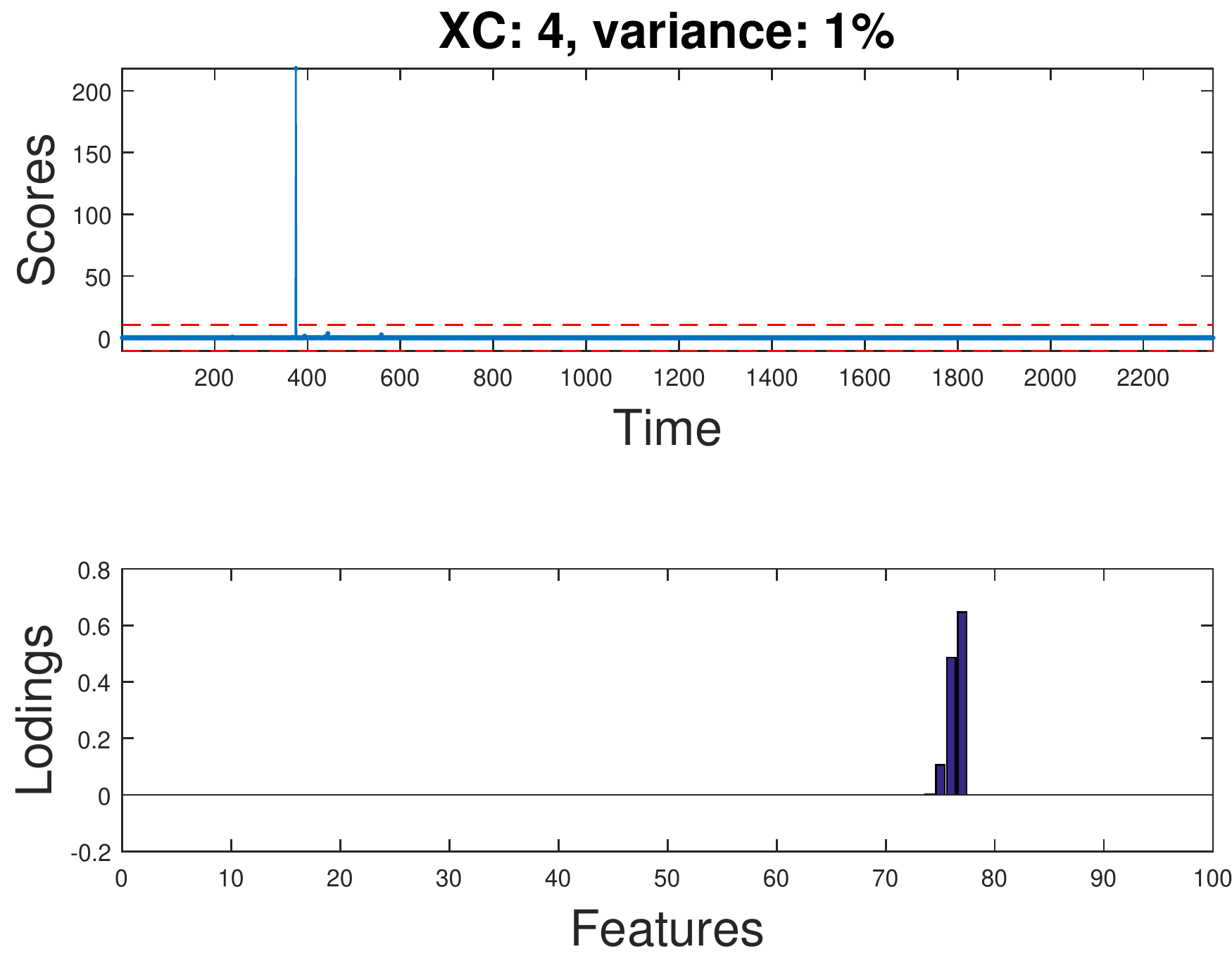}} \\
	{\includegraphics[width=0.3\textwidth]{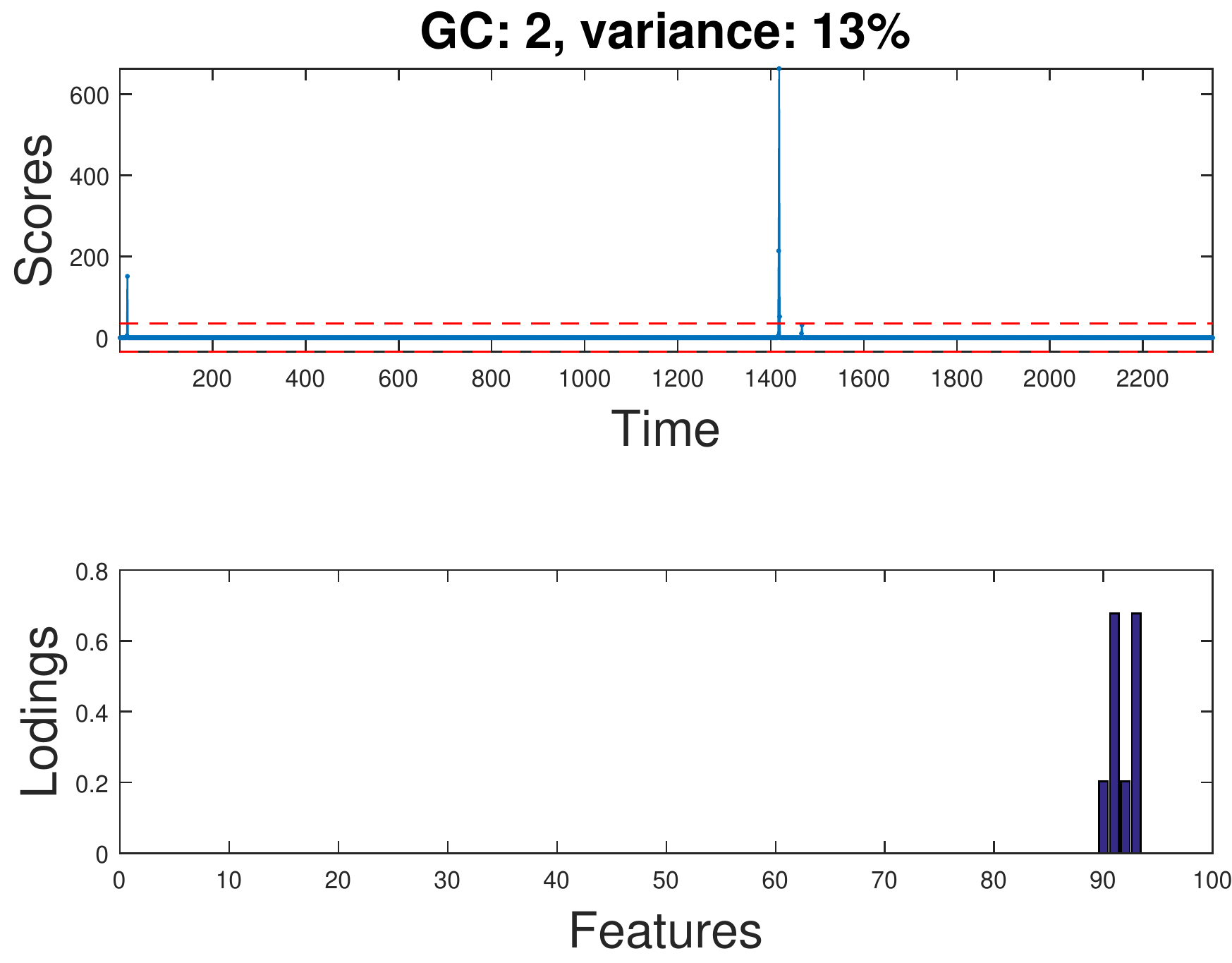}} \hfill
	{\includegraphics[width=0.3\textwidth]{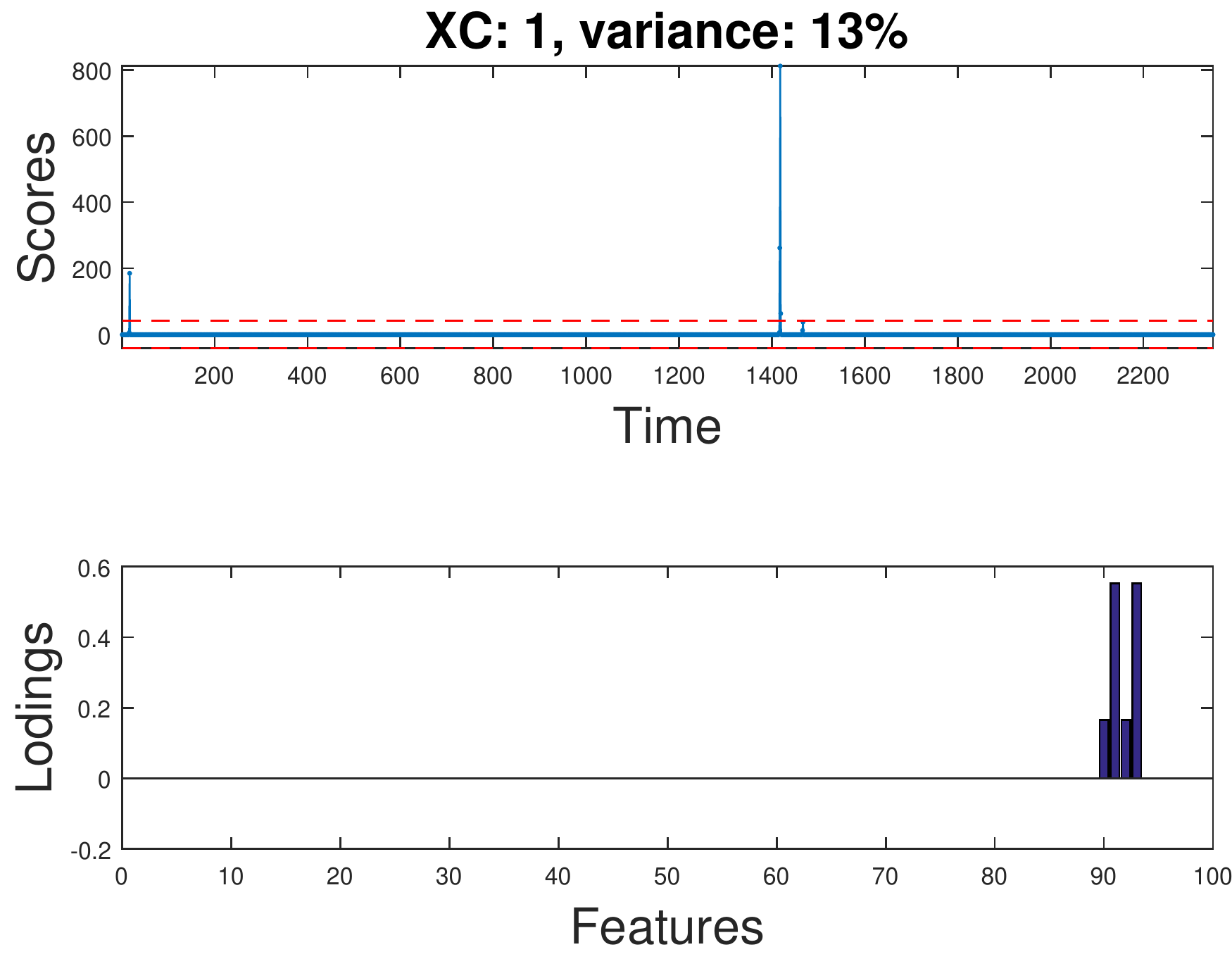}} \hfill
	{\includegraphics[width=0.3\textwidth]{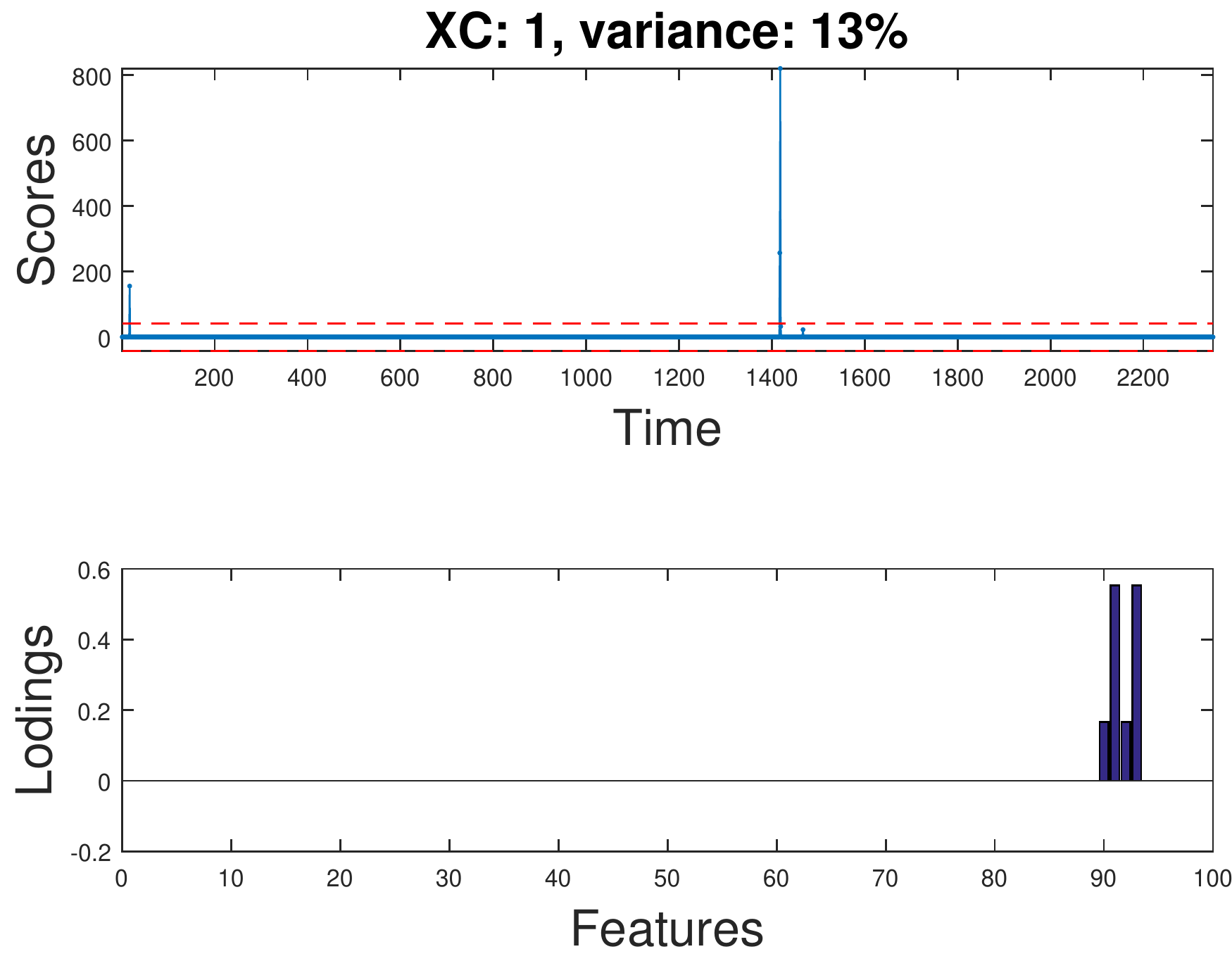}} \\
	{\includegraphics[width=0.3\textwidth]{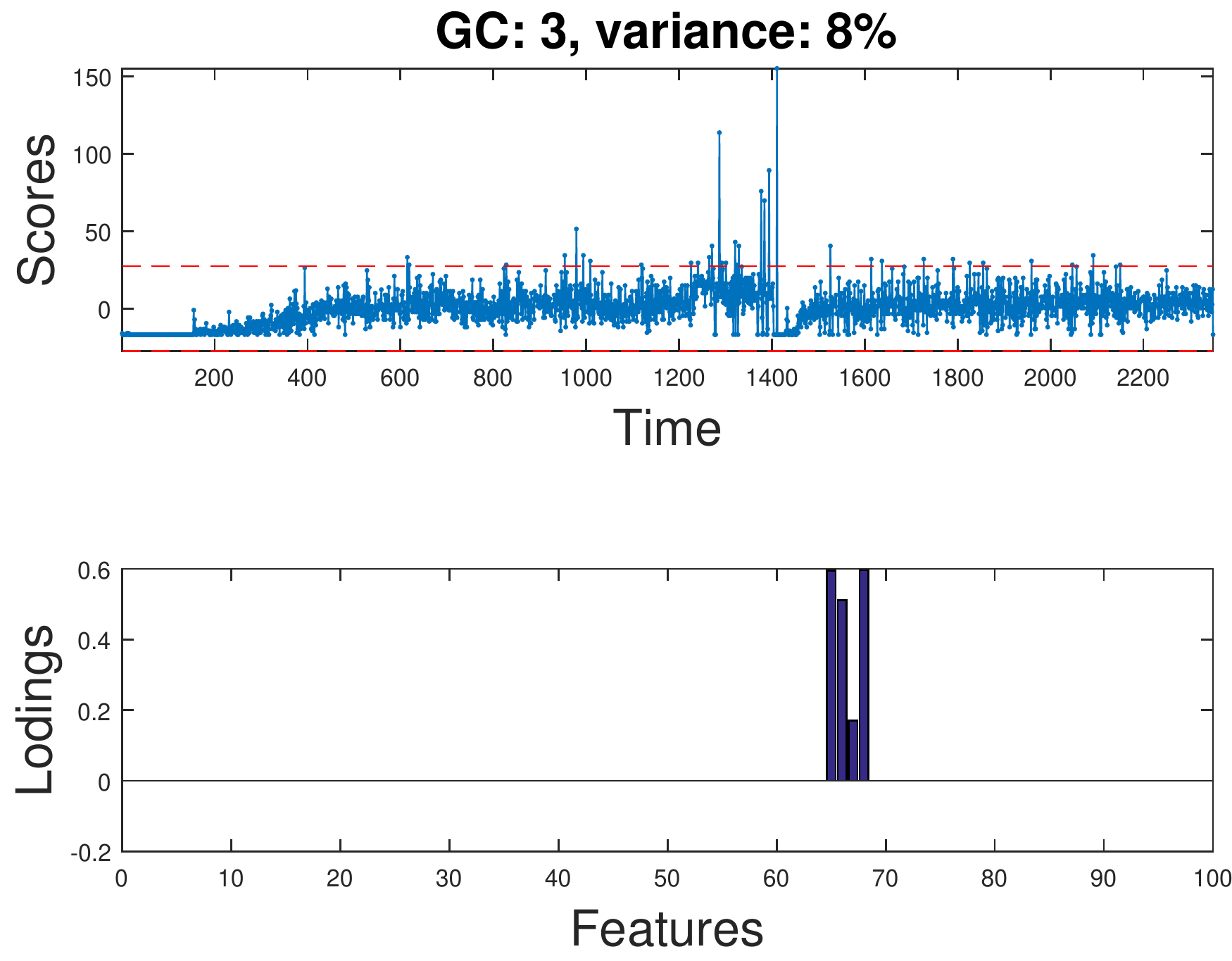}} \hfill
	{\includegraphics[width=0.3\textwidth]{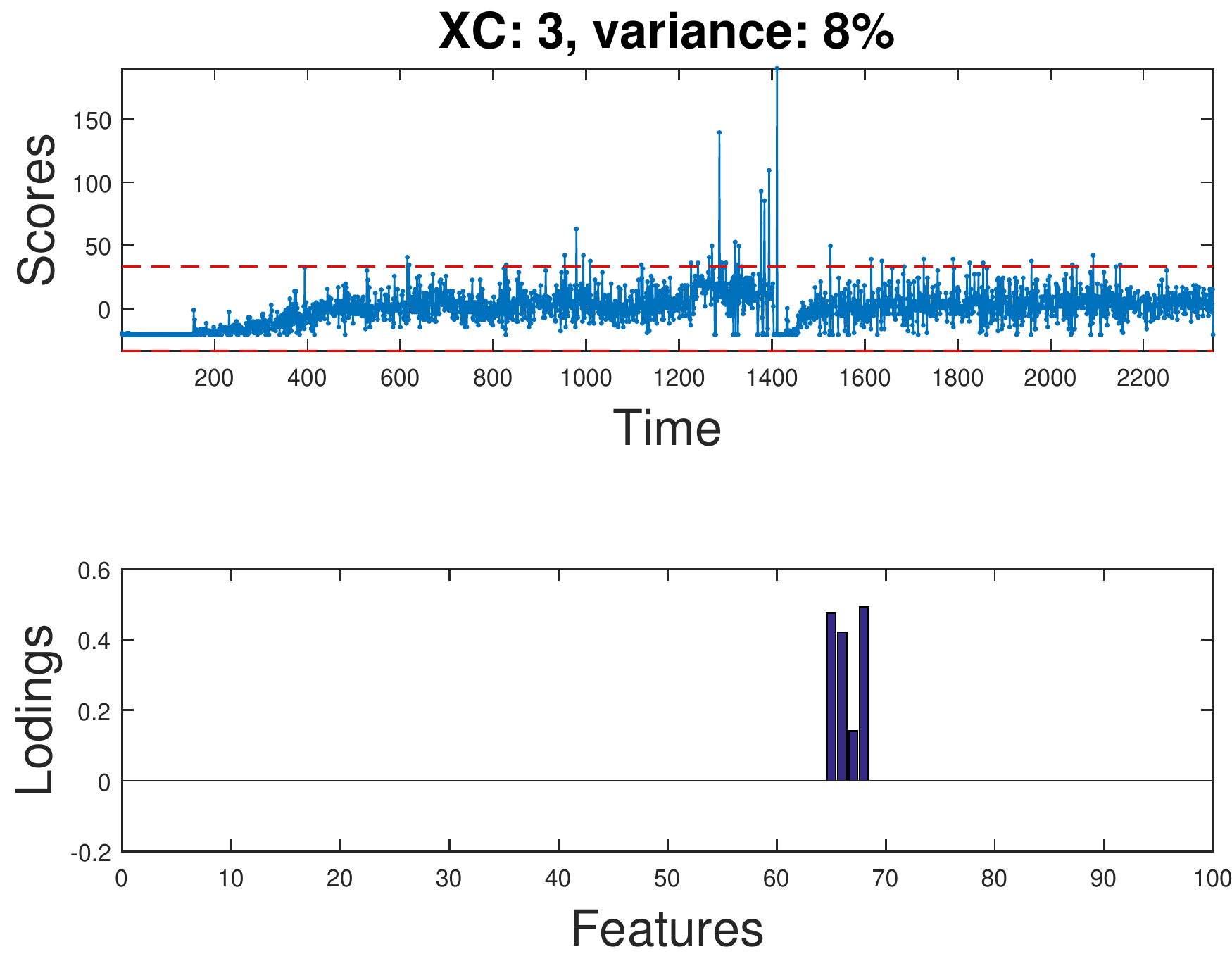}} \hfill
	{\includegraphics[width=0.3\textwidth]{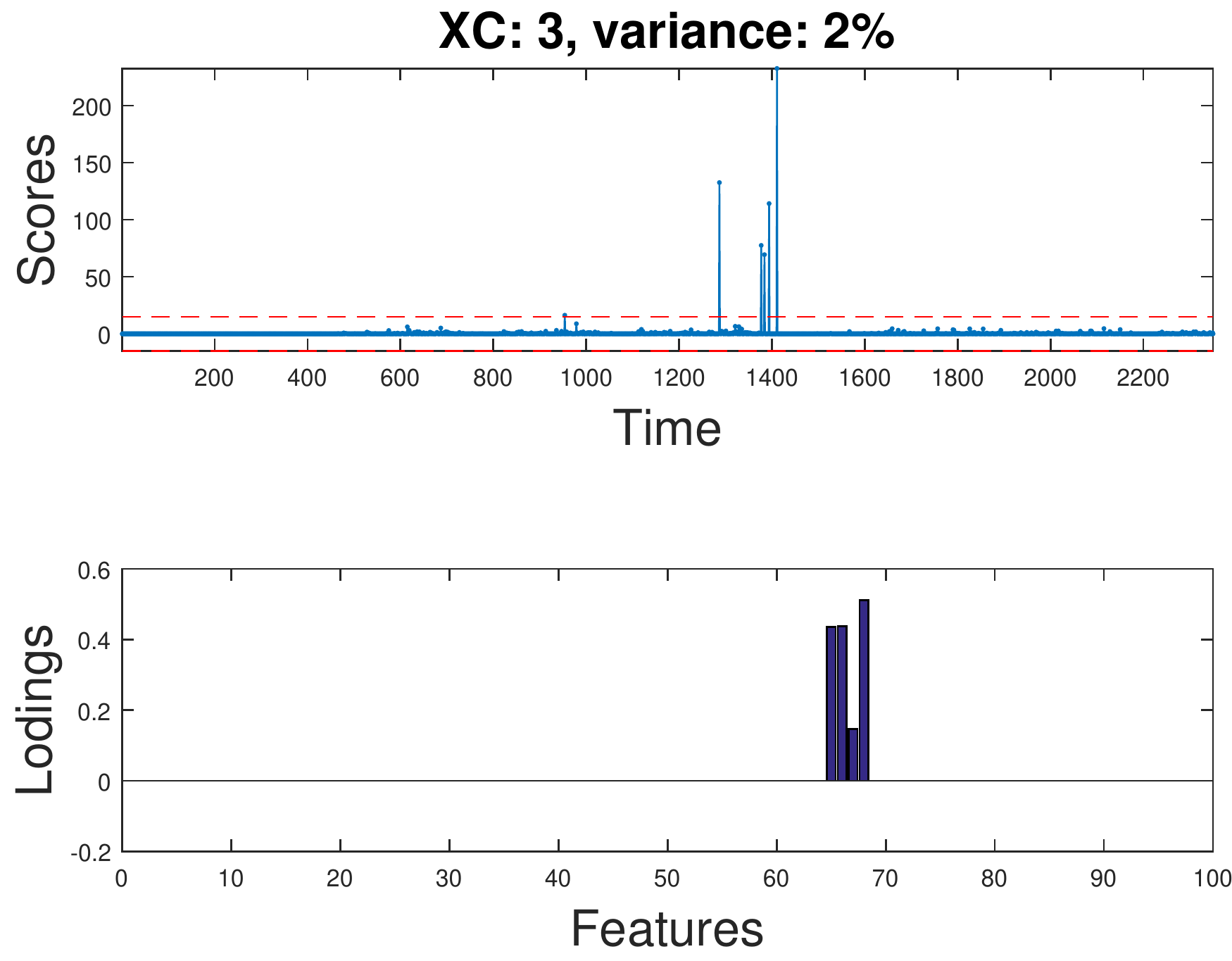}} \\
	\subfigure[GPCA 41\%]{\includegraphics[width=0.3\textwidth]{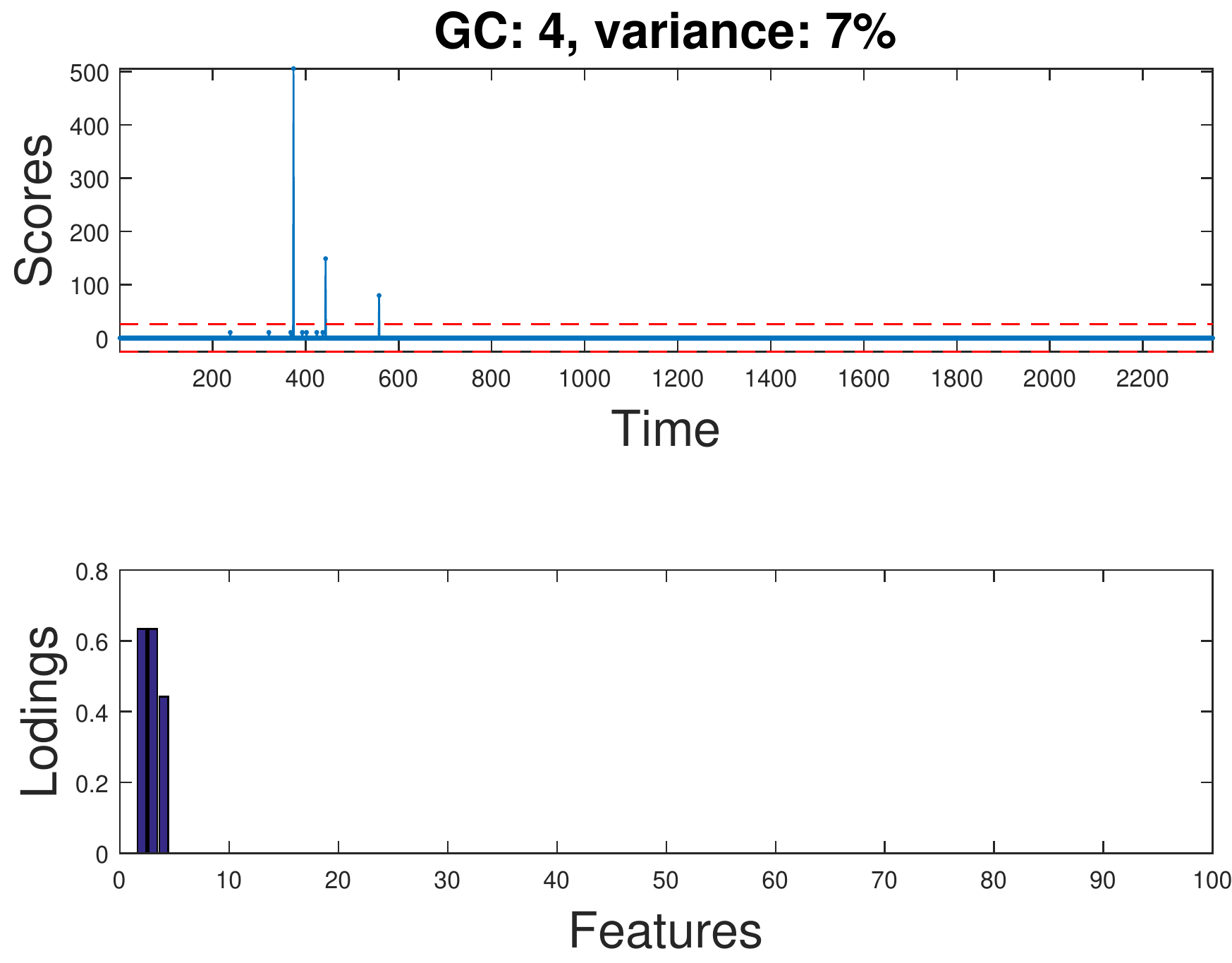}} \hfill
	\subfigure[XCAN (rows) 41\%]{\includegraphics[width=0.3\textwidth]{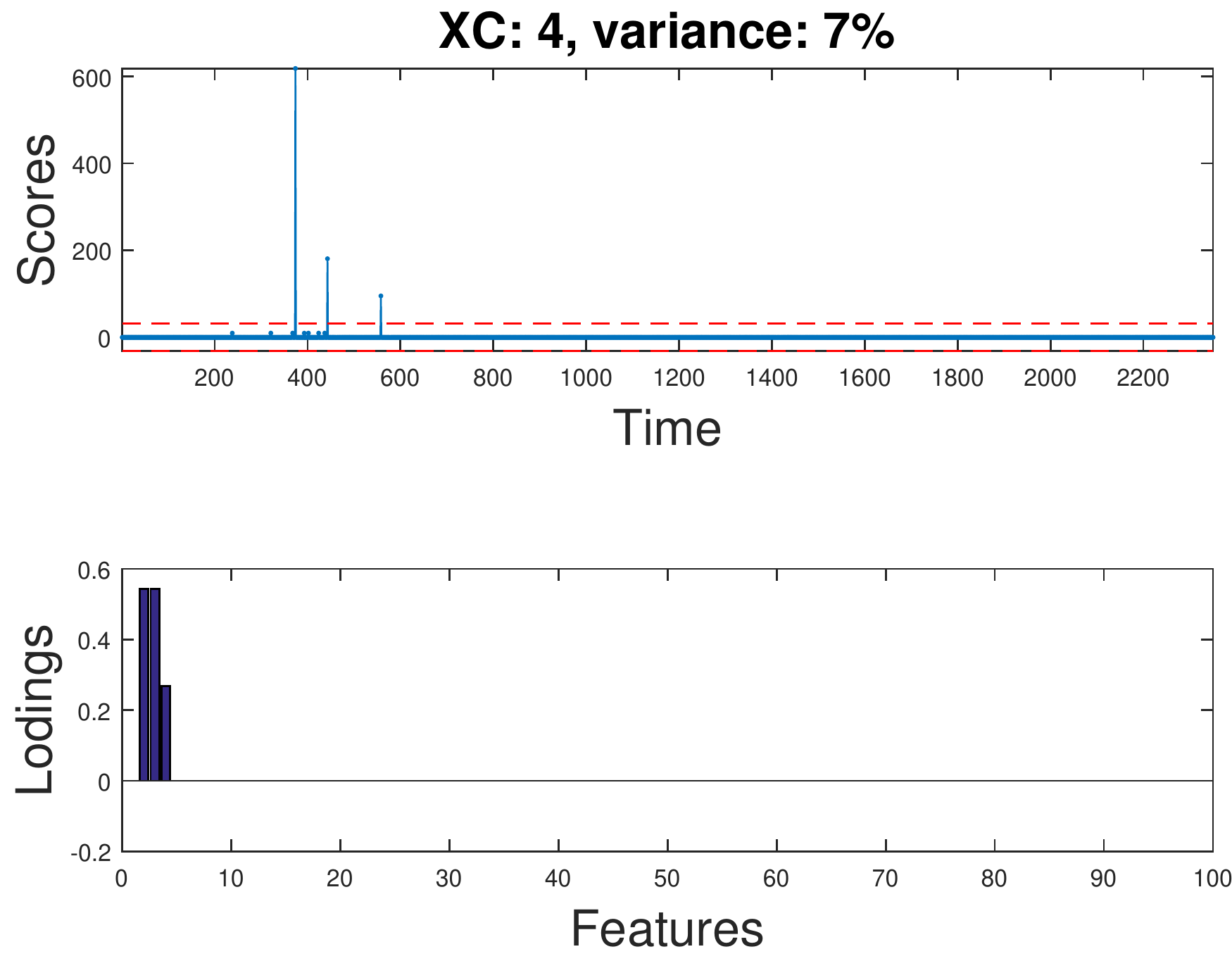}} \hfill
	\subfigure[XCAN (r\&c) 23\%]{\includegraphics[width=0.3\textwidth]{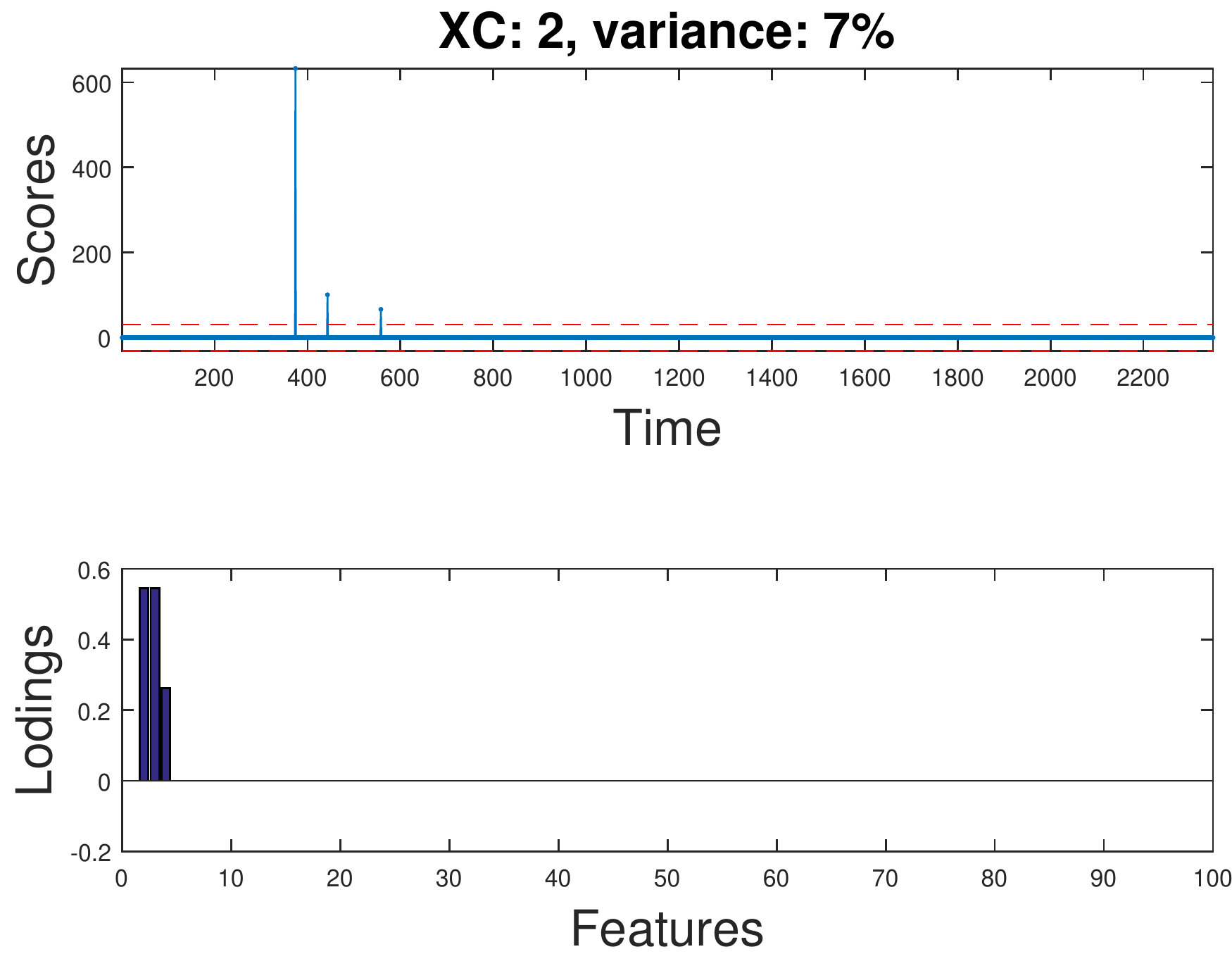}} \\
	\caption{First four GCs for GPCA (a), 4-component XCAN with sparse loadings (b), 4-component XCAN with sparse loadings and scores (c) using the Vast data.}
	\label{fig:gpca_va}
\end{figure*}

Figure \ref{fig:gpca_va} compares the first four components by GPCA (left column) and the 4-component model by XCAN with sparse loadings (middle column) and both sparse scores and loadings (right column). We re-ordered the components of XCAN to match those in GPCA. To improve the visualization for security analysts, we included statistical control limits in the scores of each component, so that anomalies can be easily identified. Both GPCA and XCAN provide useful results for anomaly detection, and components can be easily interpreted one-at-a-time. For instance, the first component of GPCA (upper-left figure) describes a number of anomalies around sampling time 400, which are related to the variables in the loadings. So a brief description of the location in time of the anomalies, and their diagnosis (the variables related to the abnormal behavior) is obtained in a single plot. In comparison to PCA (e.g., see \cite{MSNM2016}), this approach for anomaly detection is much simpler and easy to understand for security analysts.    

We can also see that GPCA and XCAN with sparse loadings are very similar. 
If we apply also sparsity in the rows in XCAN, a reduced set of observations are identified for each component, allowing the analyst to focus on a subset of time points to proceed with a more detailed forensic analysis.

\subsubsection{Olive Data}

\begin{figure}
	\centering
	\includegraphics[width=.55\textwidth]{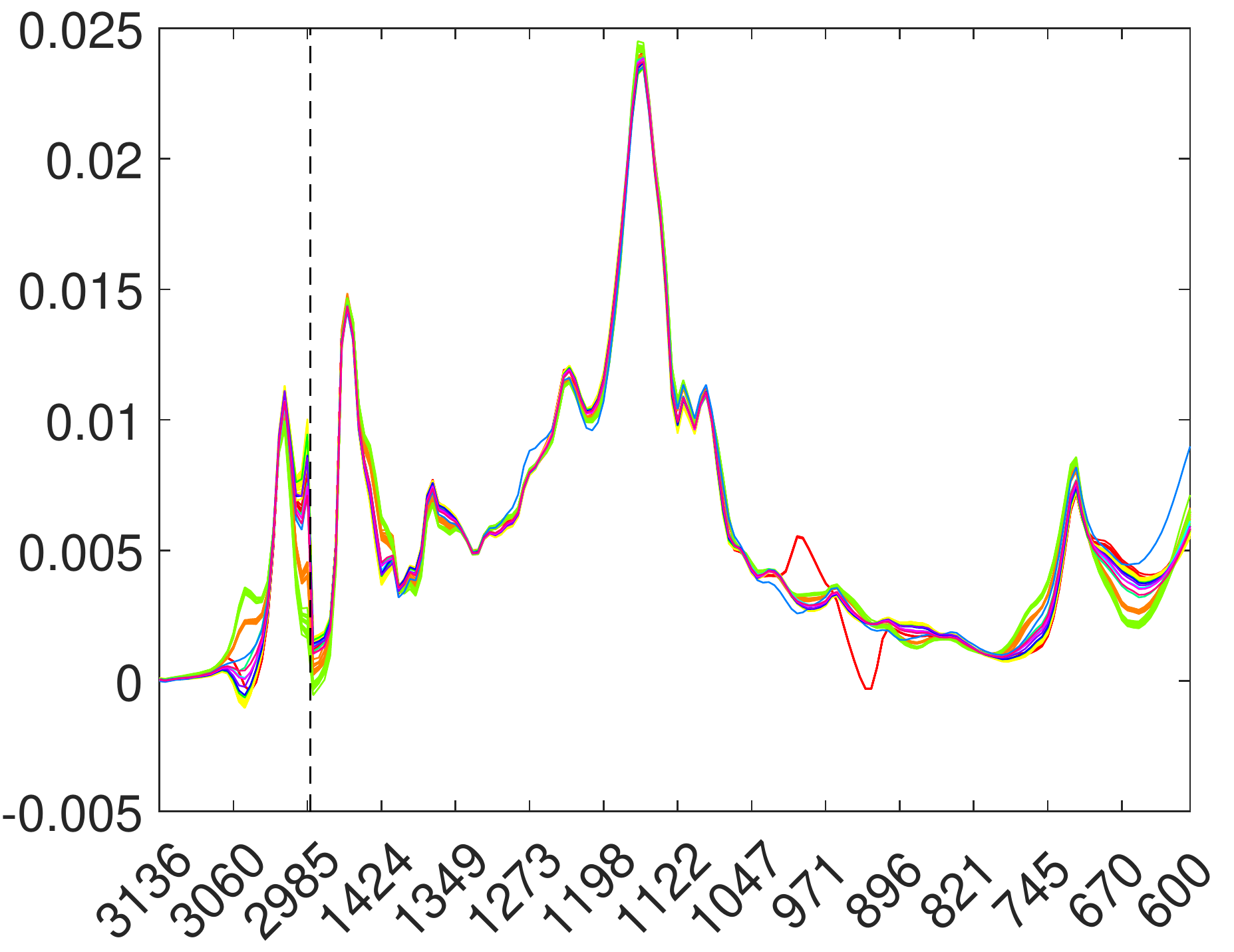}
	\caption{Spectra of different olive oil brands.}
	\label{fig:Olsp}
\end{figure}

{Oil samples from 24 brands and several types (olive, corn, sesame, etc.) were obtained. Infrared spectra were measured using a Nicolet 5-DX FT-IR system. Each spectrum consisted of 1556 measurements from 3600 to 600 cm-1 of which two regions were used in the analysis here in accordance with the original publication \cite{doi:10.1366/0003702971941935}.} The resulting spectra, with dimension $80 \times 188$, are shown in Figure 
\ref{fig:Olsp}, with different colors representing different brands of oil.

\begin{figure*}
	\centering
	{\includegraphics[width=0.45\textwidth]{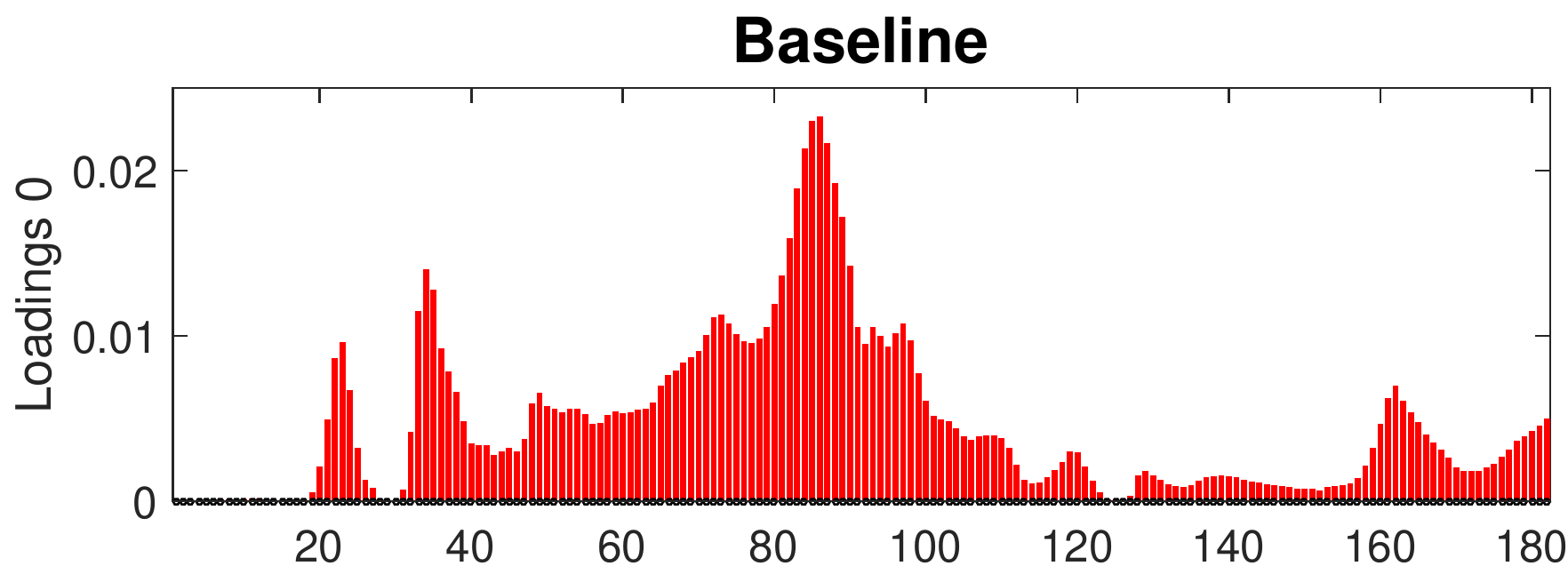}} \\
	{\includegraphics[width=0.45\textwidth]{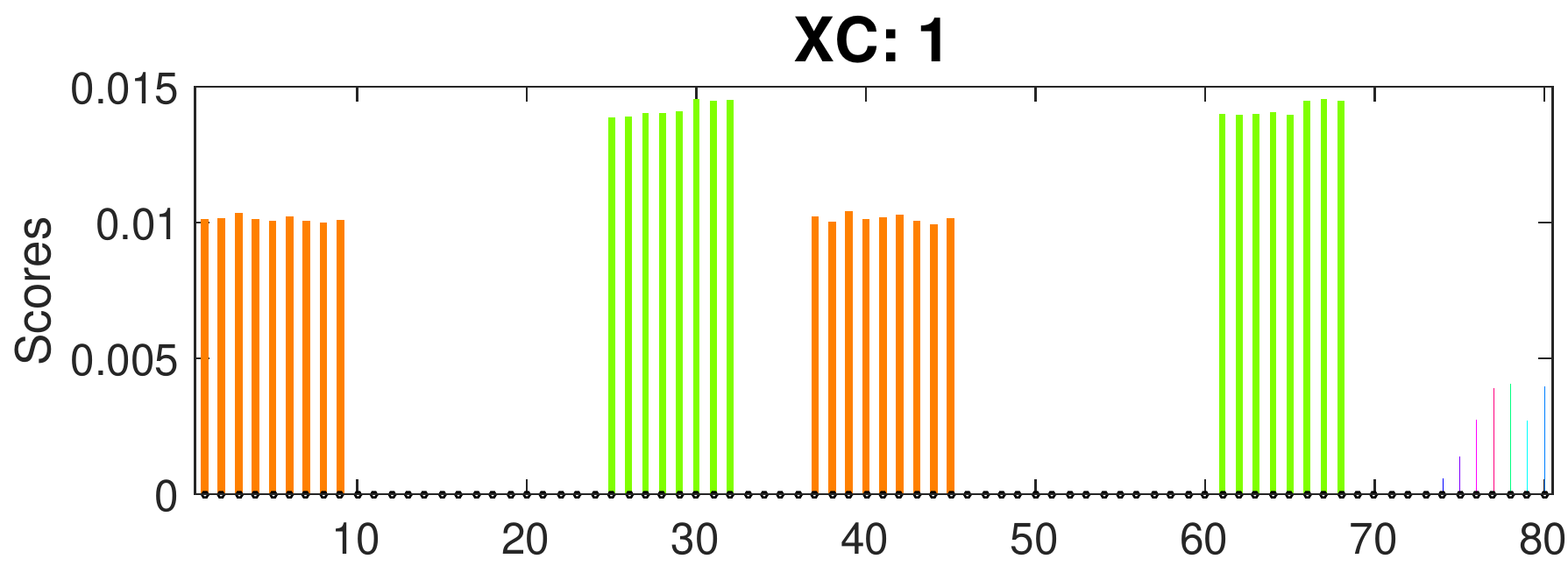}}  \hfill
	{\includegraphics[width=0.45\textwidth]{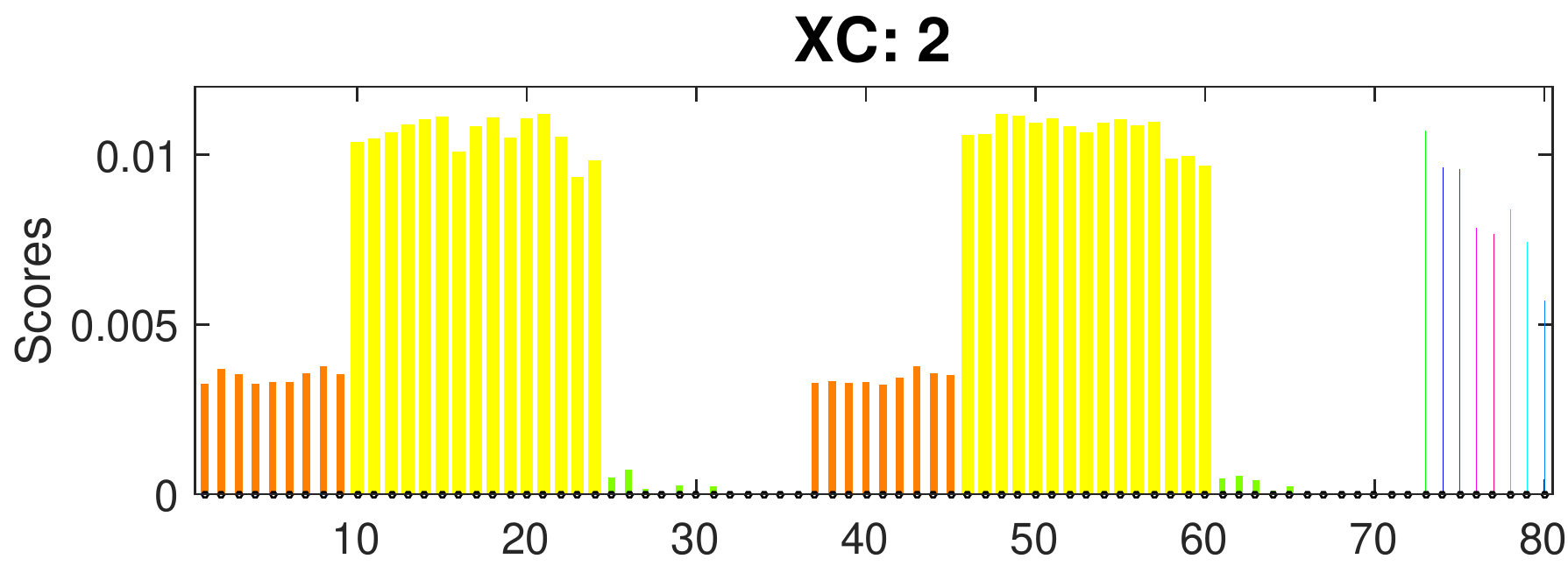}}  \\ 
	{\includegraphics[width=0.45\textwidth]{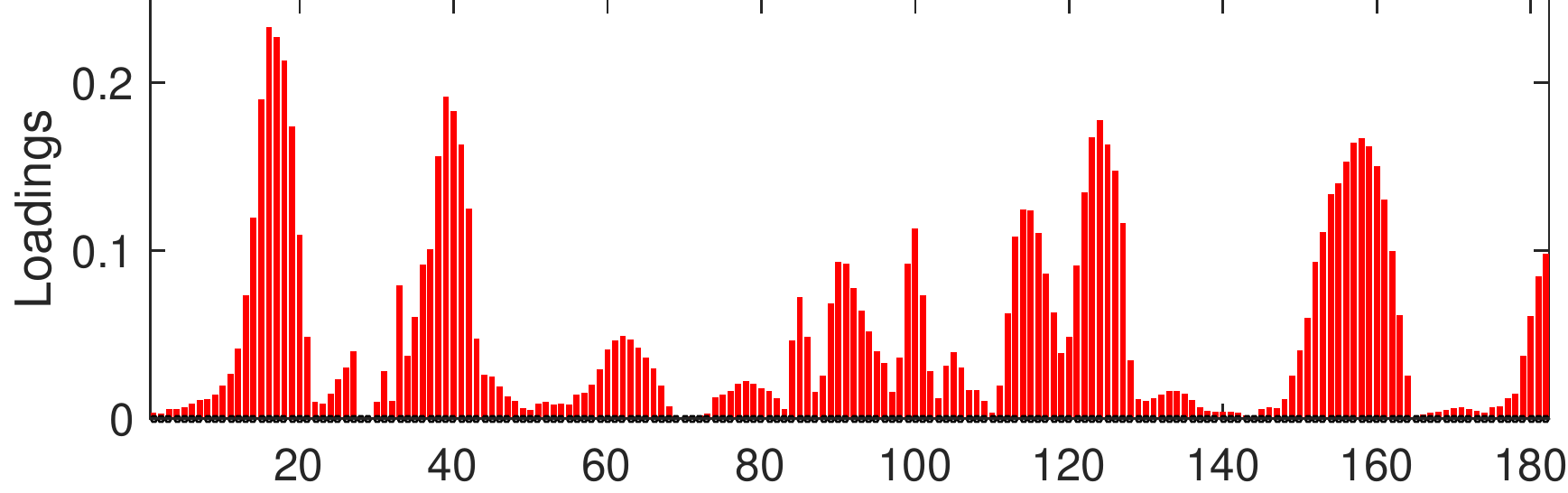}} \hfill
	{\includegraphics[width=0.45\textwidth]{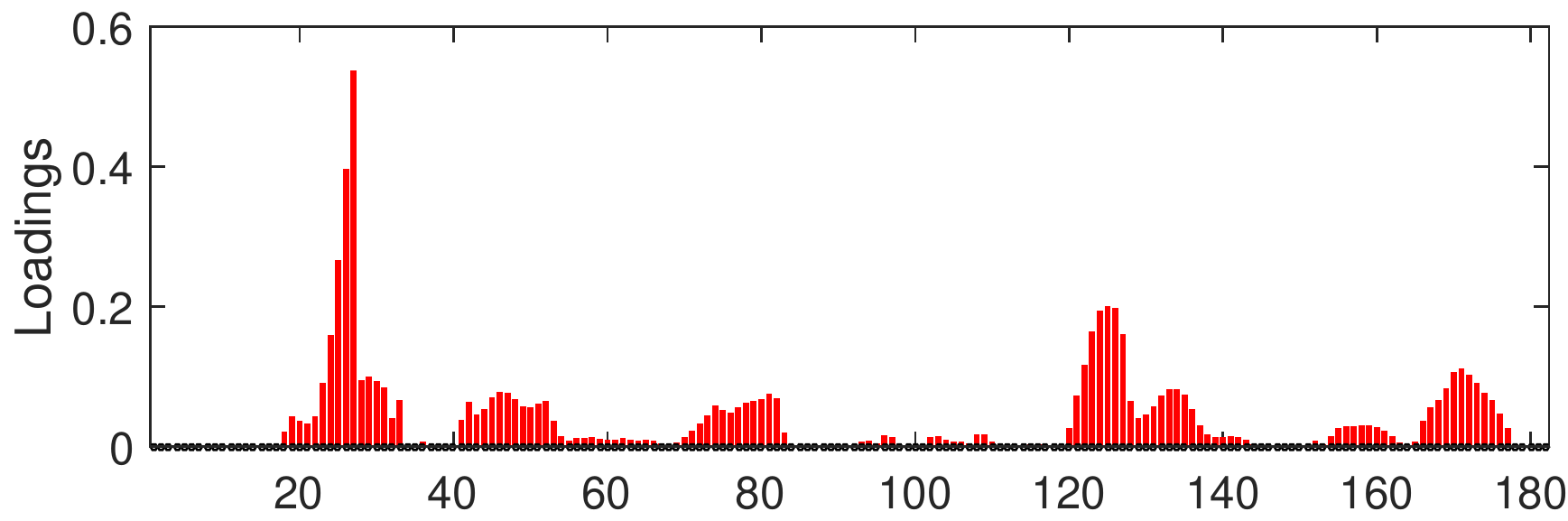}} \\
	{\includegraphics[width=0.45\textwidth]{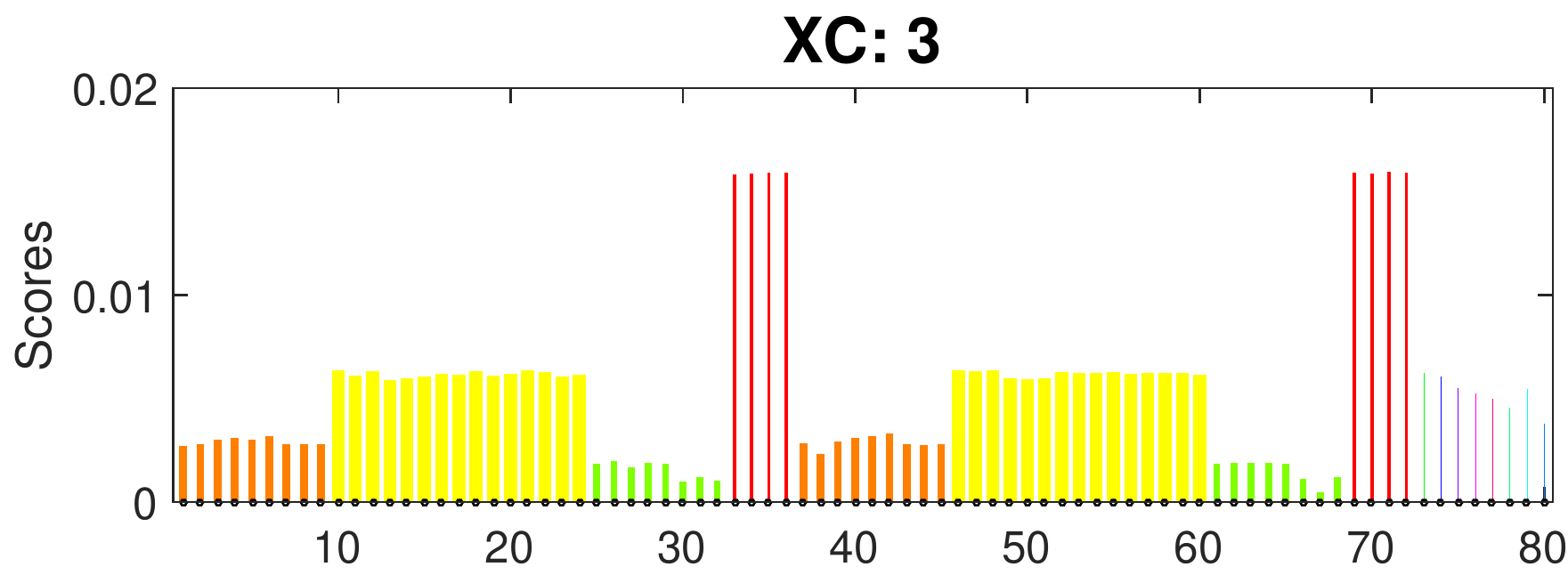}}  \hfill
	{\includegraphics[width=0.45\textwidth]{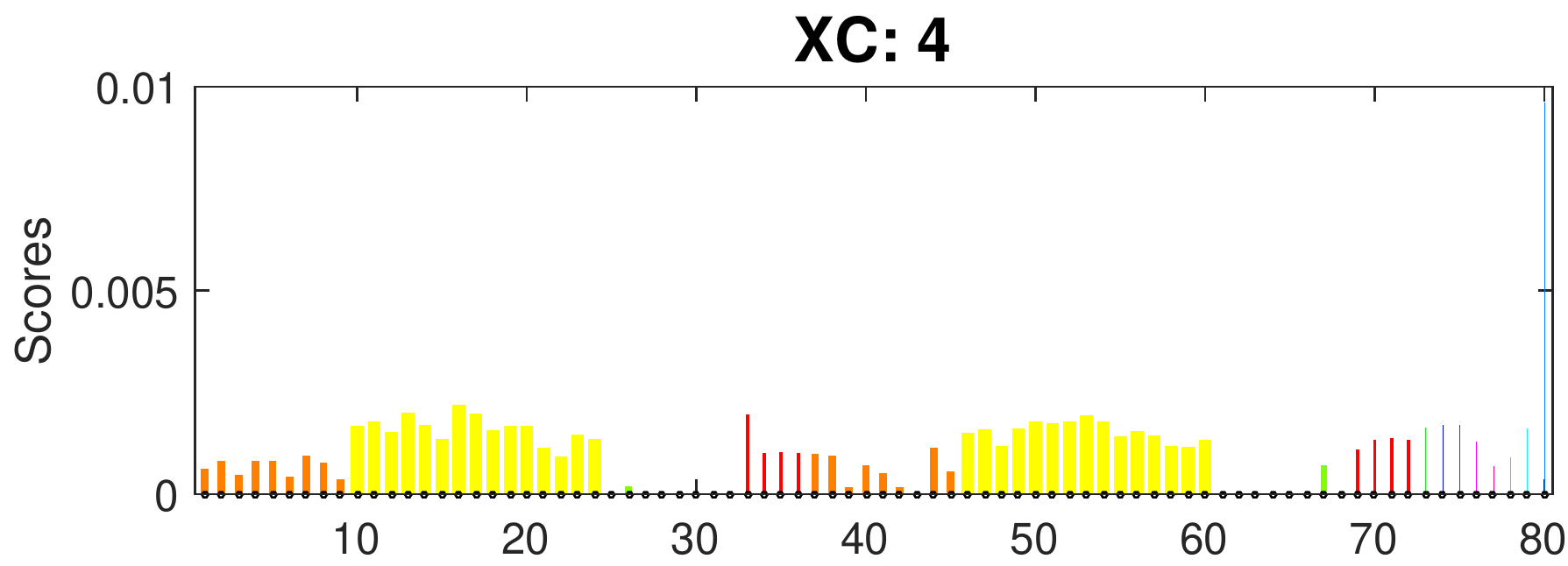}} \\ 
	{\includegraphics[width=0.45\textwidth]{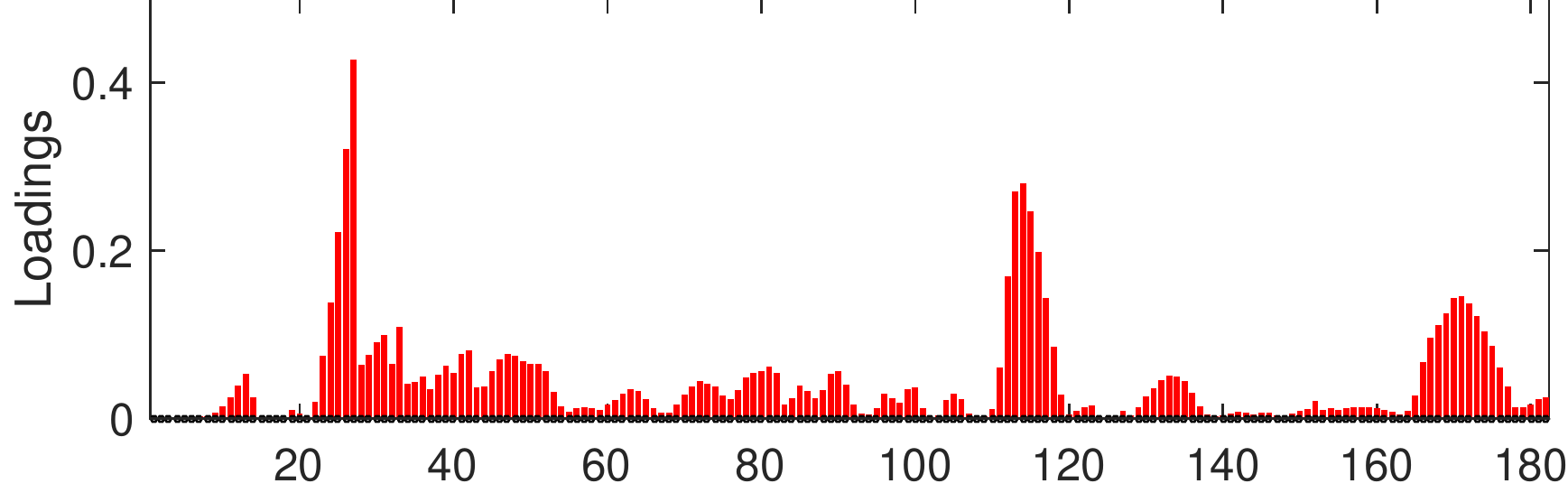}} \hfill
	{\includegraphics[width=0.45\textwidth]{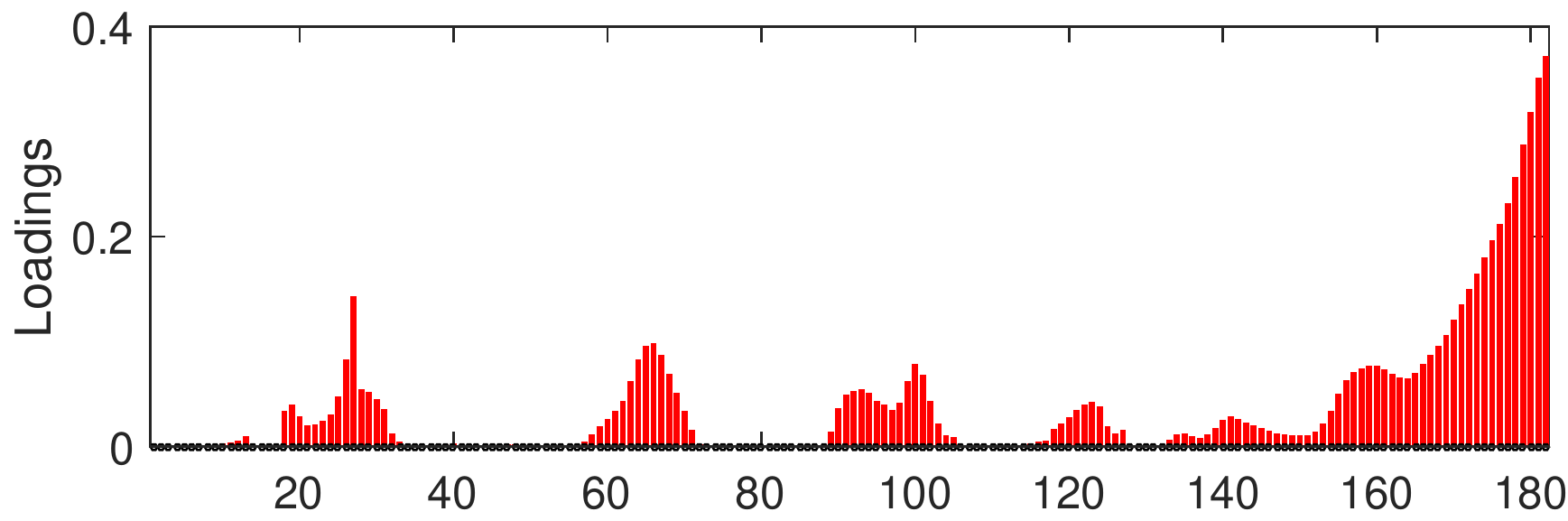}} 
	\caption{Olive data. XCAN model with baseline and non-negativity constraints on $\mathbf{\hat{U}}^X$, $\mathbf{\hat{S}}^X$ and $\mathbf{\hat{T}}^X$, and deactivated structural penalties.}
	\label{fig:PCA_NN}
\end{figure*}

One meaningful way to model spectra with matrix factorization is:

\begin{equation} \label{eq:PCAm_at}
\mathbf{X}_{nc} = \mathbf{1}\mathbf{\hat{p}}_0^T + \mathbf{\hat{T}}\mathbf{\hat{P}}^T +
\mathbf{E}\ s.t. \  \mathbf{\hat{T}} \ge 0, \ \mathbf{\hat{P}} \ge 0
\end{equation}

\noindent where $\mathbf{\hat{T}}$ and $\mathbf{\hat{P}}$ are non-negative, and $\mathbf{\hat{p}}_0$ takes the roll of a baseline. For this example, we will follow the same approach with XCAN, and also incorporate the baseline and non-negativity constraints on $\mathbf{\hat{U}}^X$, $\mathbf{\hat{S}}^X$ and $\mathbf{\hat{T}}^X$. The result with $\lambda_c=\lambda_r=0$ is shown in Figure 
\ref{fig:PCA_NN}. The top shows the baseline, contained in $\mathbf{\hat{p}}_0$. We observe that components are generally not sparse.

\begin{figure*}
	\centering
	\subfigure[]{\includegraphics[width=0.4\textwidth]{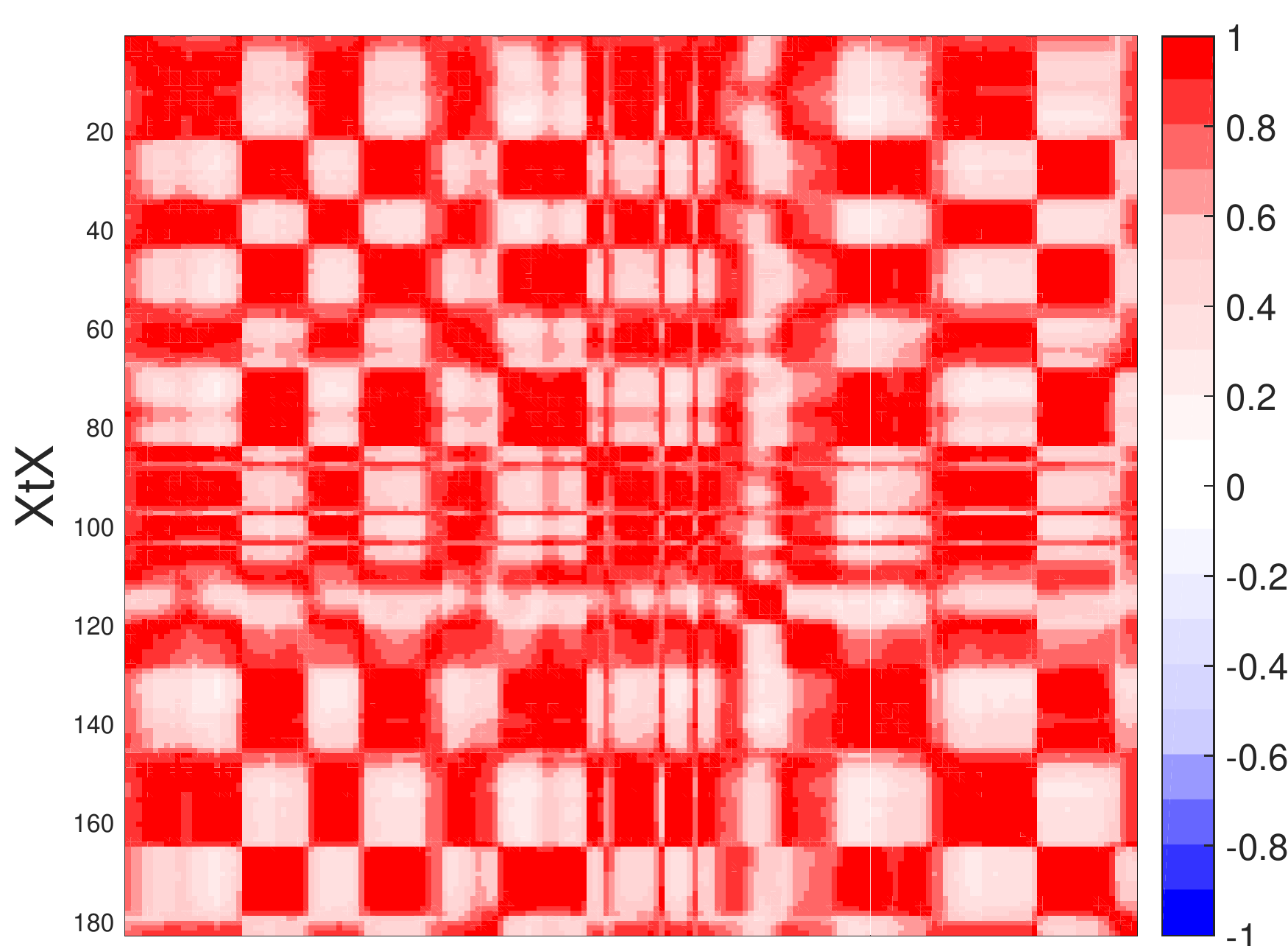}}\hfill
	\subfigure[]{\includegraphics[width=0.4\textwidth]{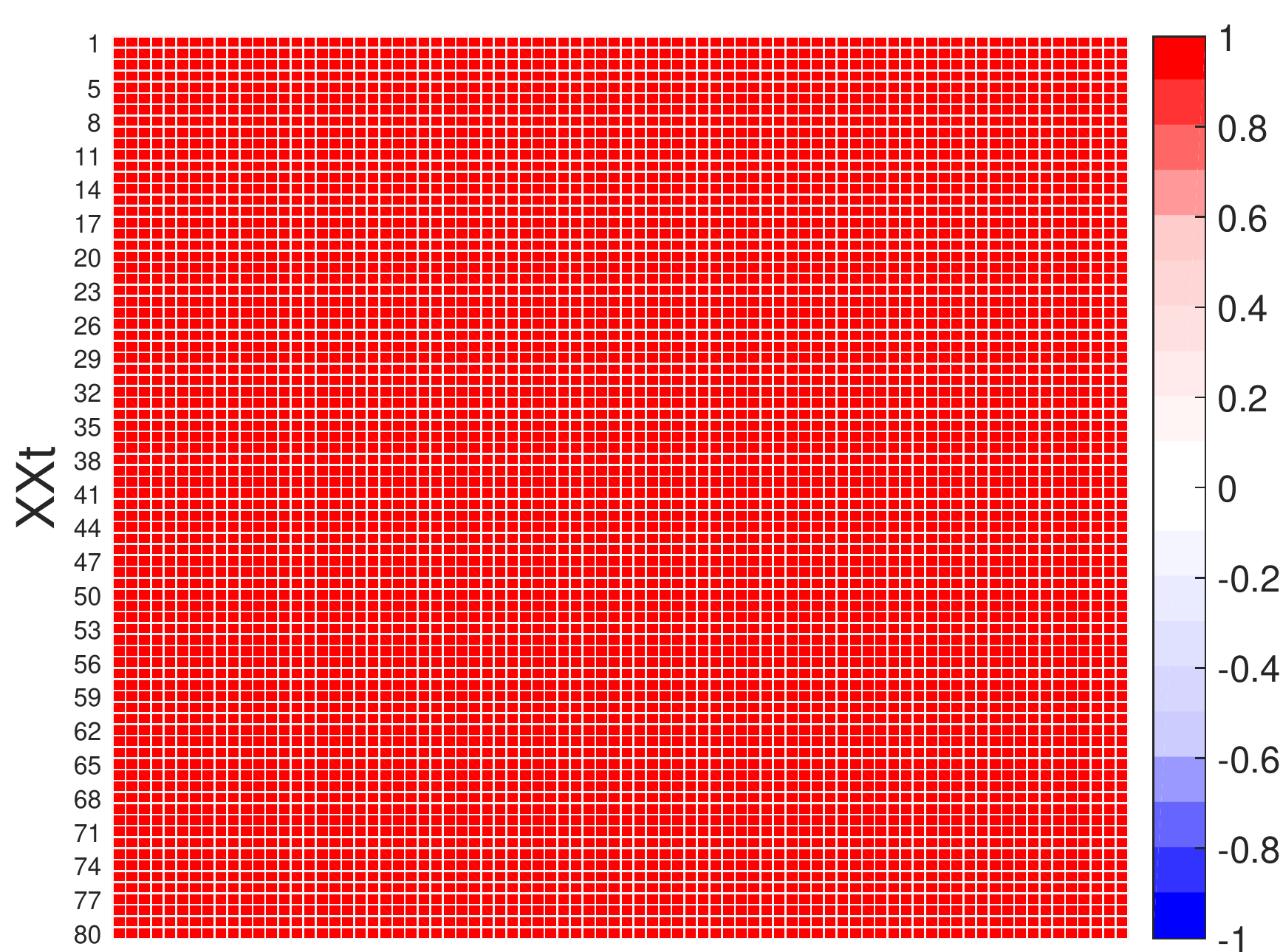}} \hfill
	\subfigure[]{\includegraphics[width=0.4\textwidth]{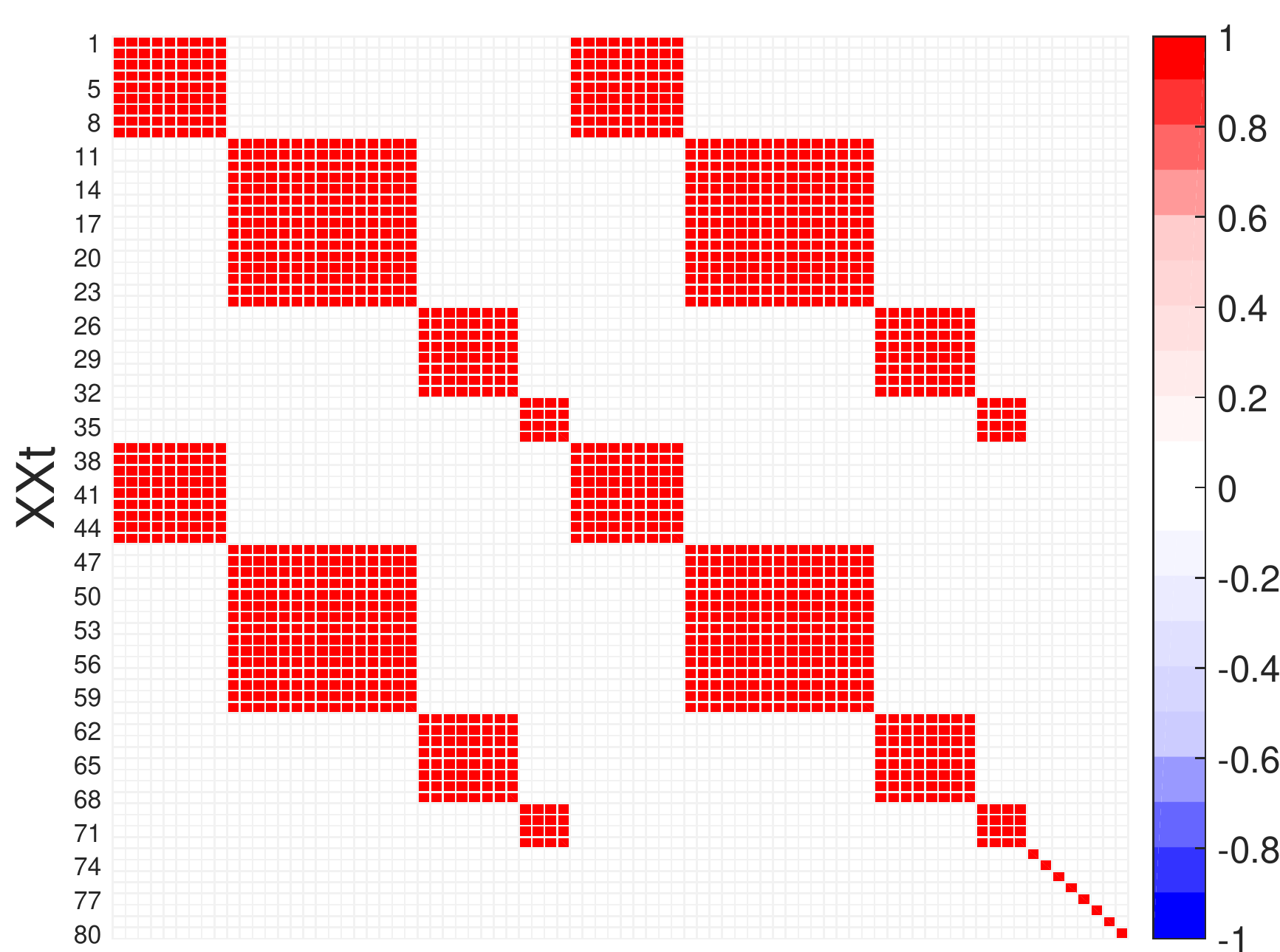}} 
	\caption{Cross-product matrices to impose structural penalties in XCAN for the Olive Data: (a) $\mathbf{XtX}$, (b) $\mathbf{XXt}$, and (c) observation class map (1: equal class, 0: different class).}
	\label{fig:OlXX}
\end{figure*}

Since in this case study the data is not centered and it is far from the origin of coordinates, the cross-product expressions in eqs. (\ref{XtX}) and (\ref{XXt}) are impractical: they are almost completely filled with 1s. For this reason, we post-processed cross-product matrices by subtracting a baseline computed with the minimum values of each column (row) for $\mathbf{XtX}$ ($\mathbf{XXt}$). This approach gives us the cross-product matrices in Figure \ref{fig:OlXX}(a) and (b).  We can see that the cross-product matrix $\mathbf{XXt}$ still does not indicate any groups of observations. For this reason, we will not use this cross-product matrix with XCAN. Instead, here, we will perform a different analysis to illustrate another potential use of XCAN with data from samples, which has class information, and define $\mathbf{XXt}$ based on the classes of the samples as in Figure \ref{fig:OlXX}(c). Each matrix entry contains 1 if the corresponding observations belong to the same class, or 0 otherwise. This illustrates how we can use external information to influence the XCAN outcome, instead of using the cross-product matrix definitions given earlier.

\begin{figure*}
	\centering
	{\includegraphics[width=0.45\textwidth]{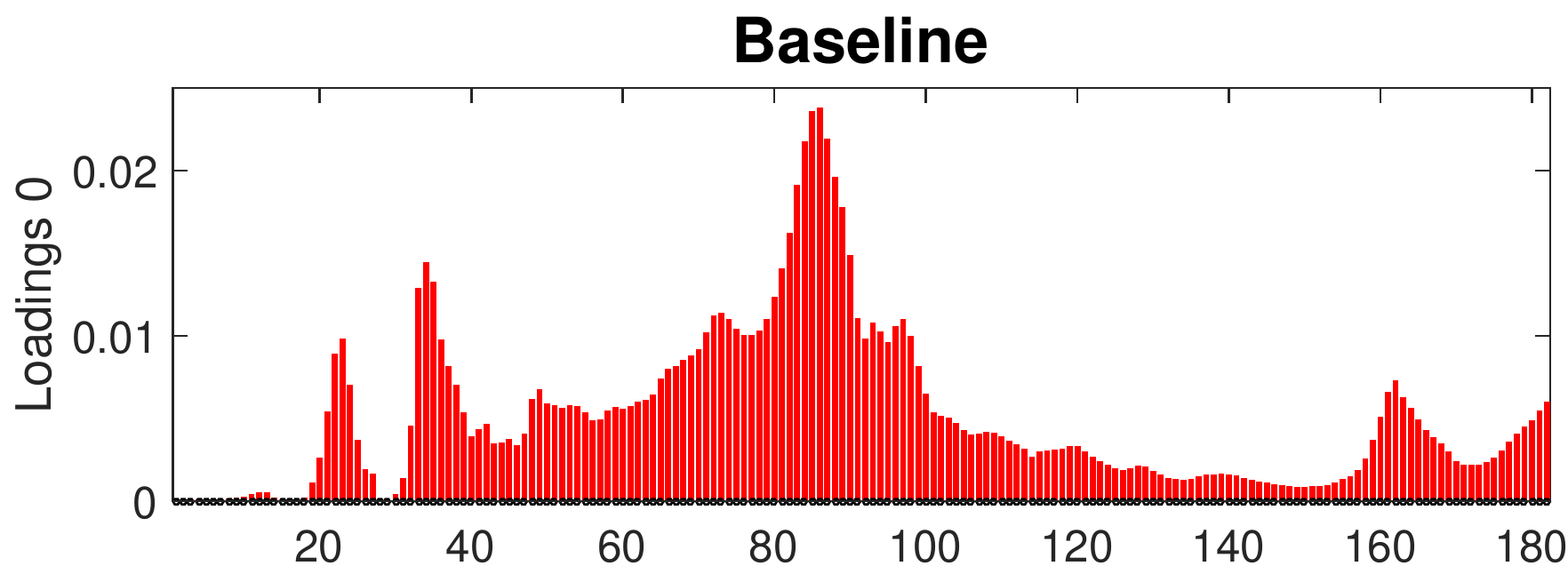}} \\
	{\includegraphics[width=0.45\textwidth]{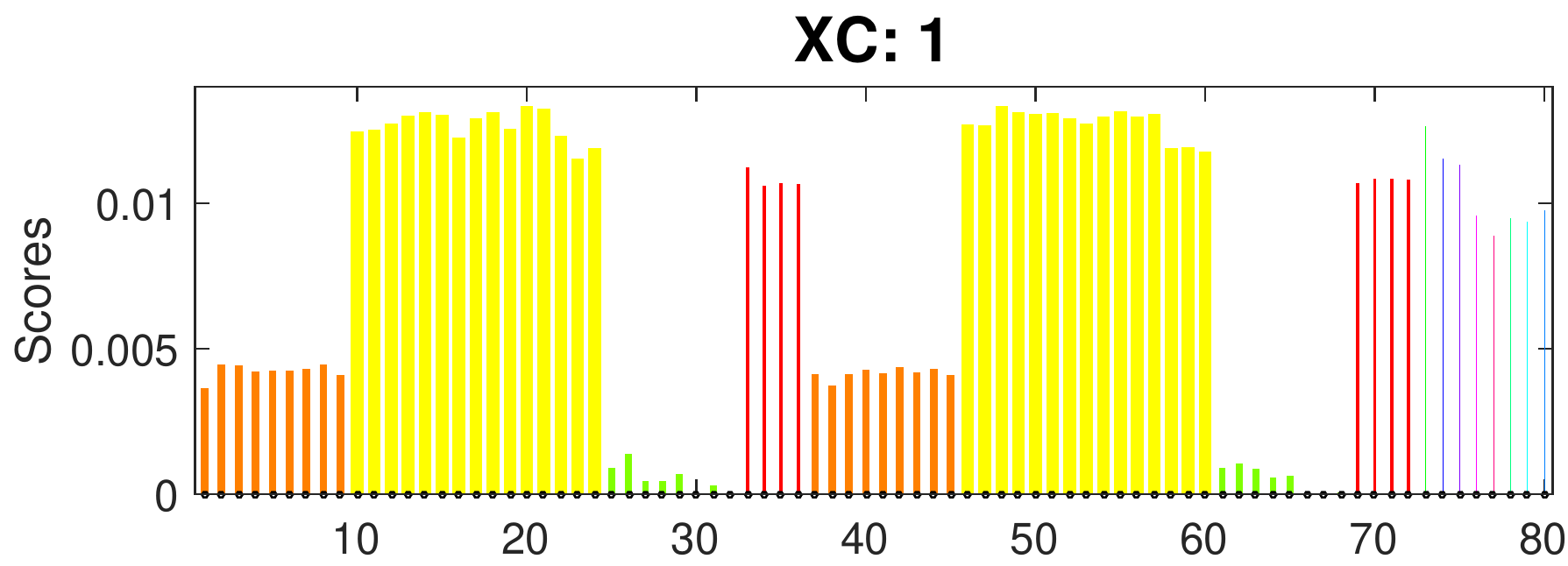}}  \hfill
	{\includegraphics[width=0.45\textwidth]{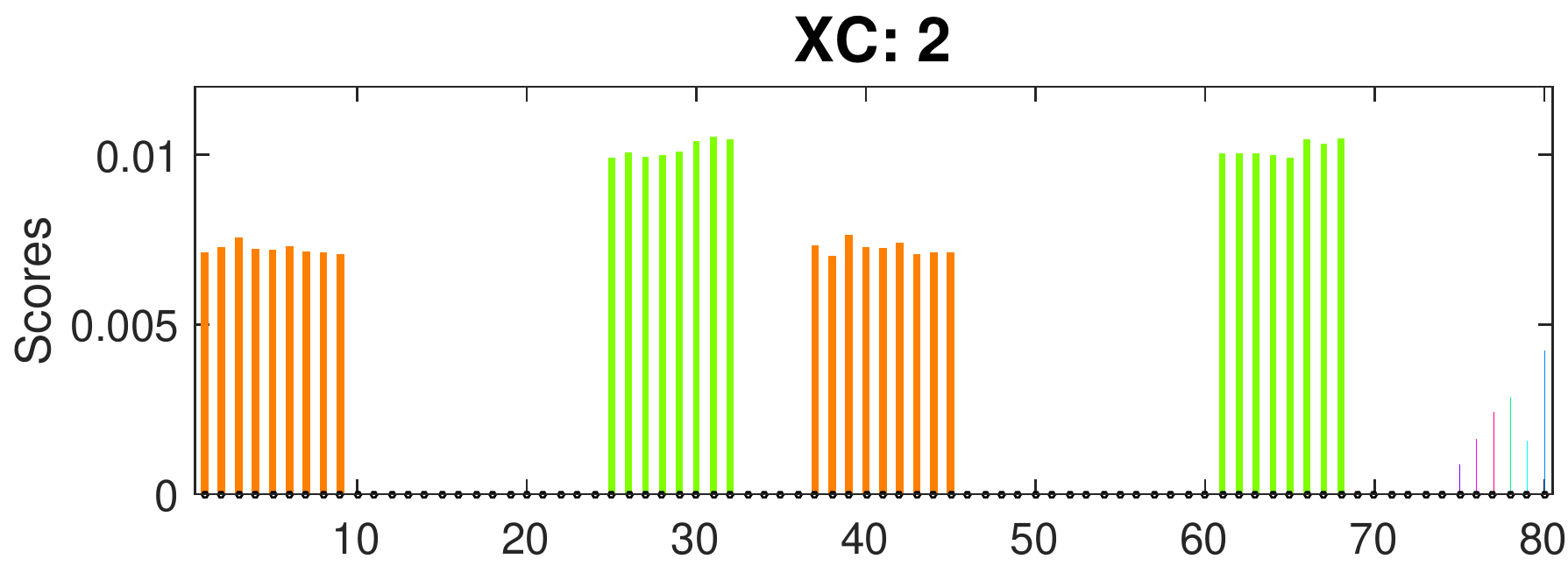}}  \\ 
	{\includegraphics[width=0.45\textwidth]{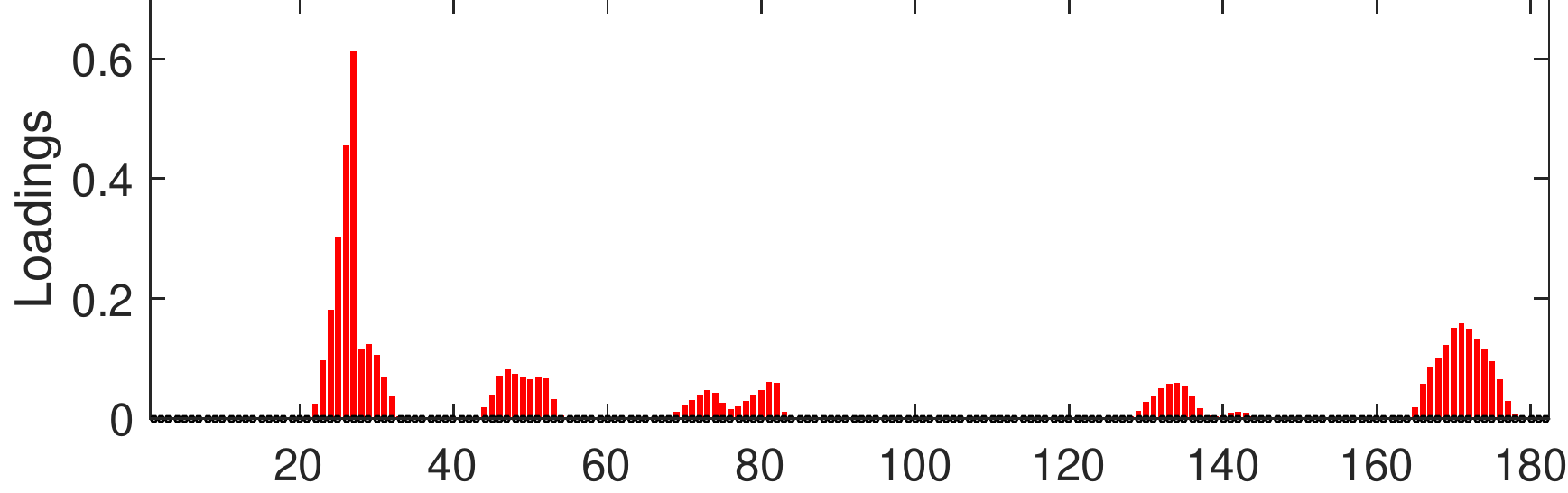}} \hfill
	{\includegraphics[width=0.45\textwidth]{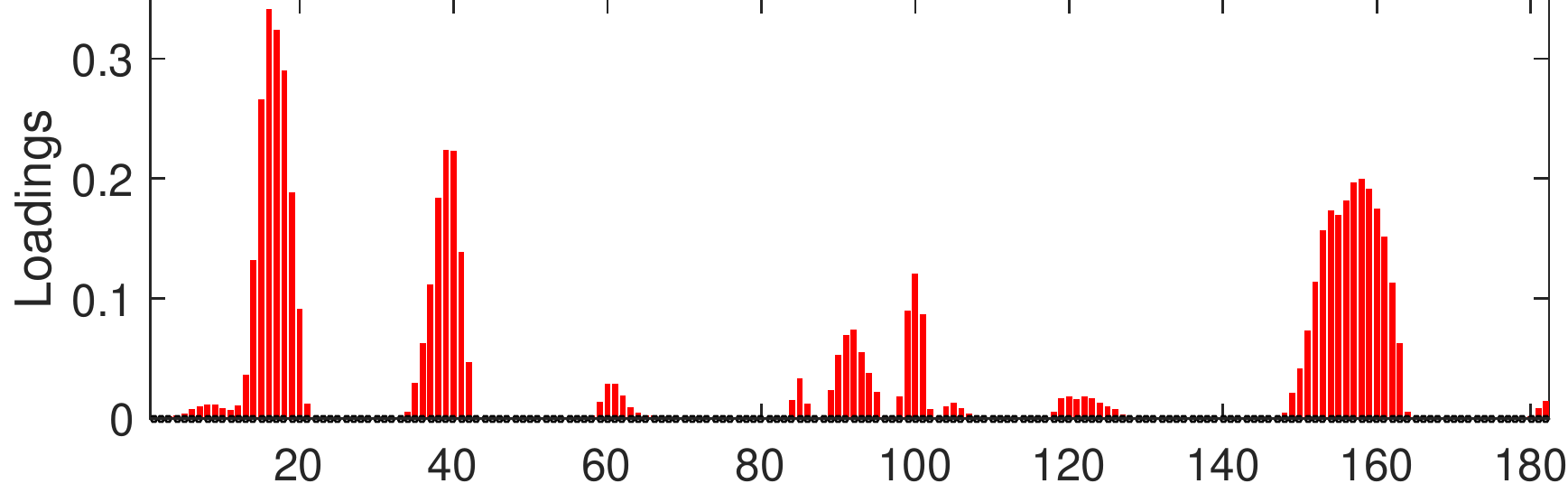}} \\
	{\includegraphics[width=0.45\textwidth]{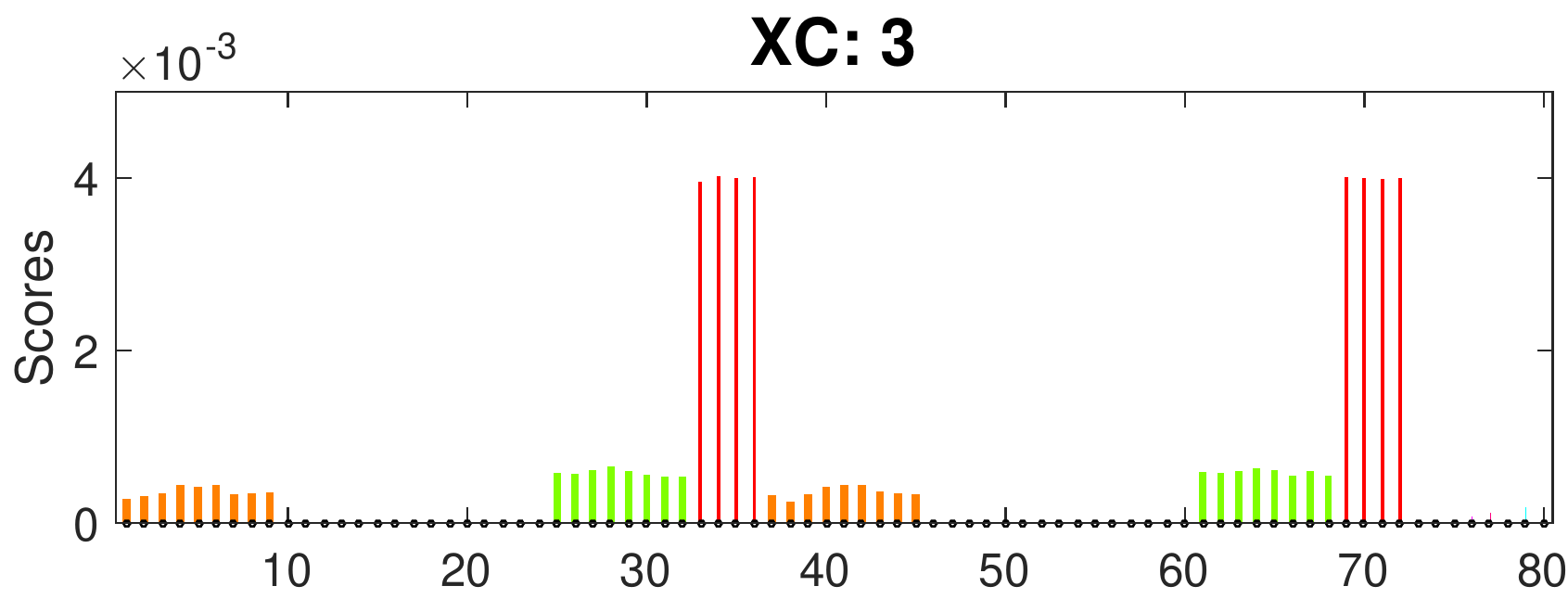}}  \hfill
	{\includegraphics[width=0.45\textwidth]{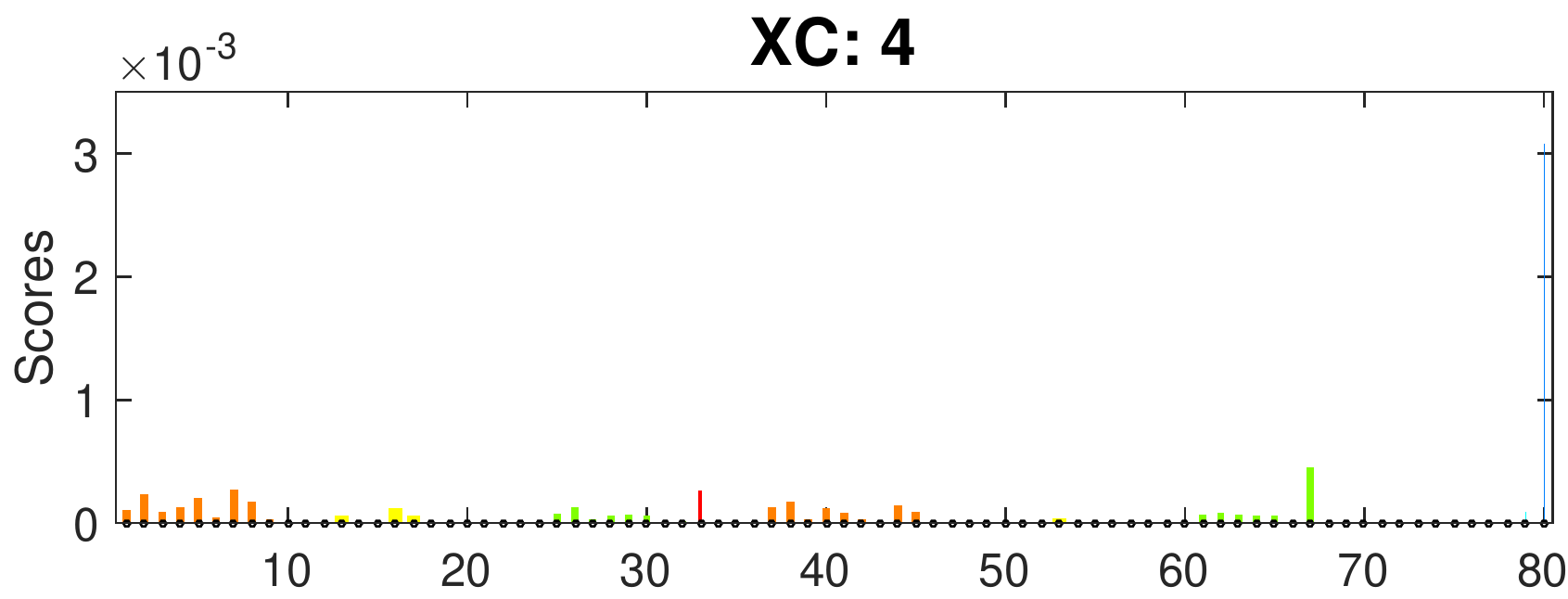}} \\ 
	{\includegraphics[width=0.45\textwidth]{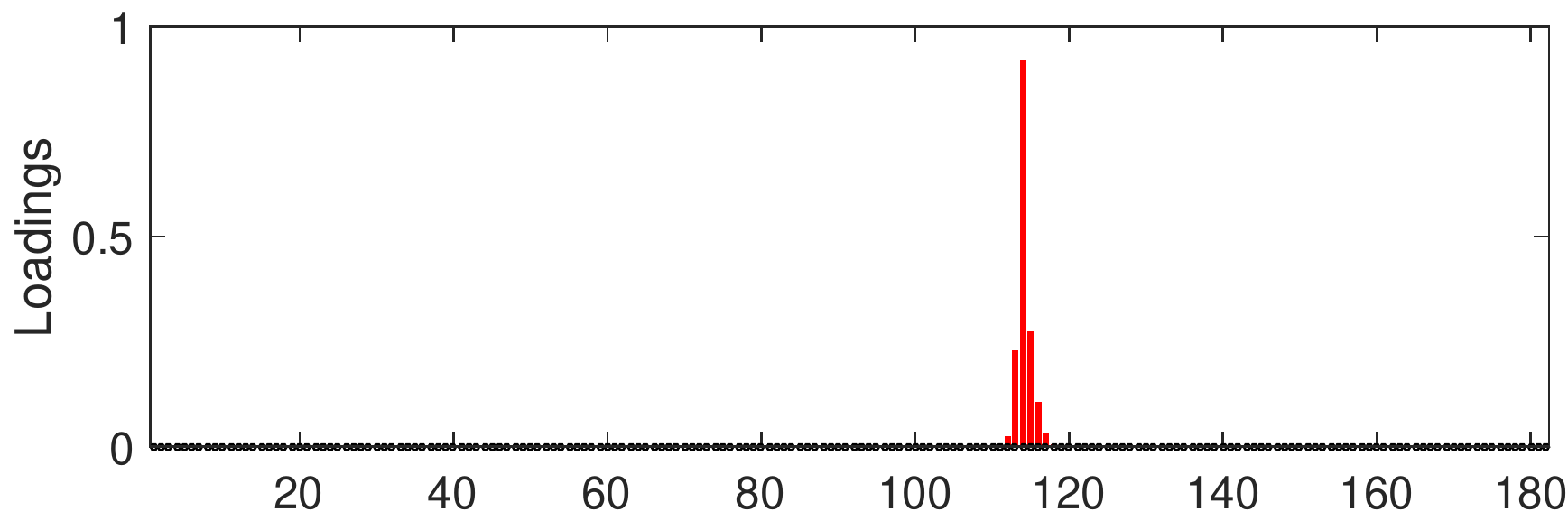}} \hfill
	{\includegraphics[width=0.45\textwidth]{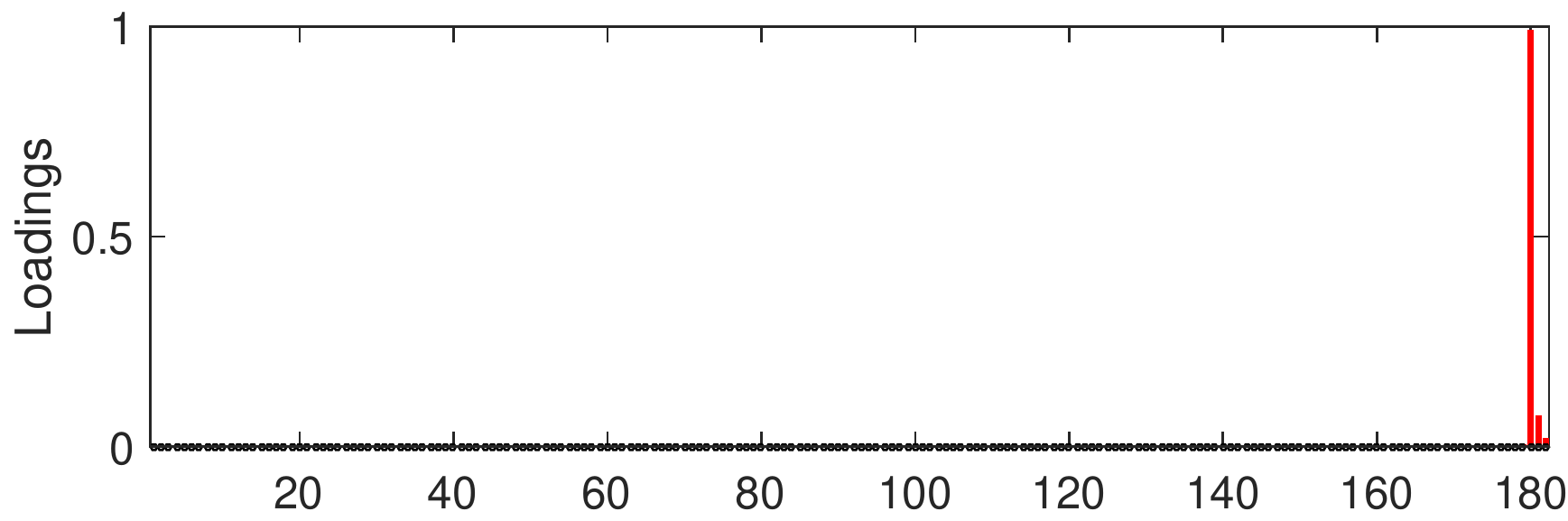}} 
	\caption{Olive data. XCAN model with baseline and non-negativity constraints on $\mathbf{\hat{U}}^X$, $\mathbf{\hat{S}}^X$ and $\mathbf{\hat{T}}^X$, penalized only with the cross-product matrix in Figure \ref{fig:OlXX}(a)).}
	\label{fig:XCAN_NN}
\end{figure*}

\begin{figure*}
	\centering
	{\includegraphics[width=0.45\textwidth]{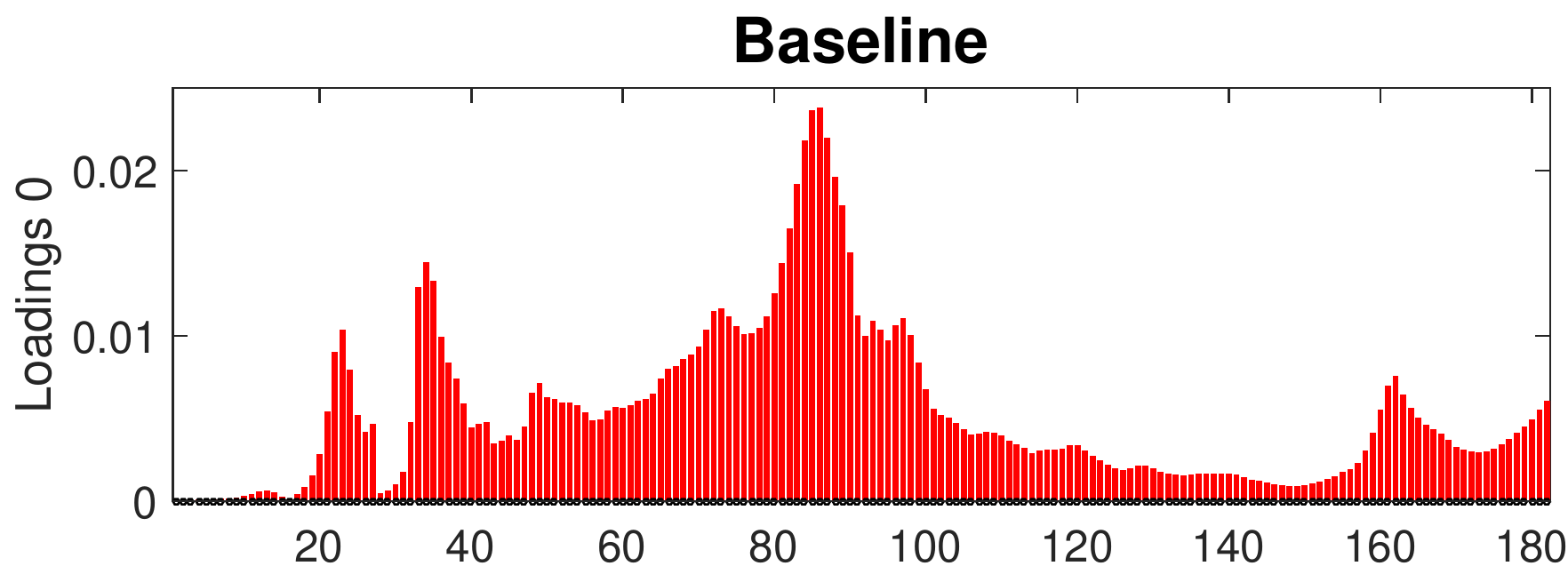}} \\
	{\includegraphics[width=0.45\textwidth]{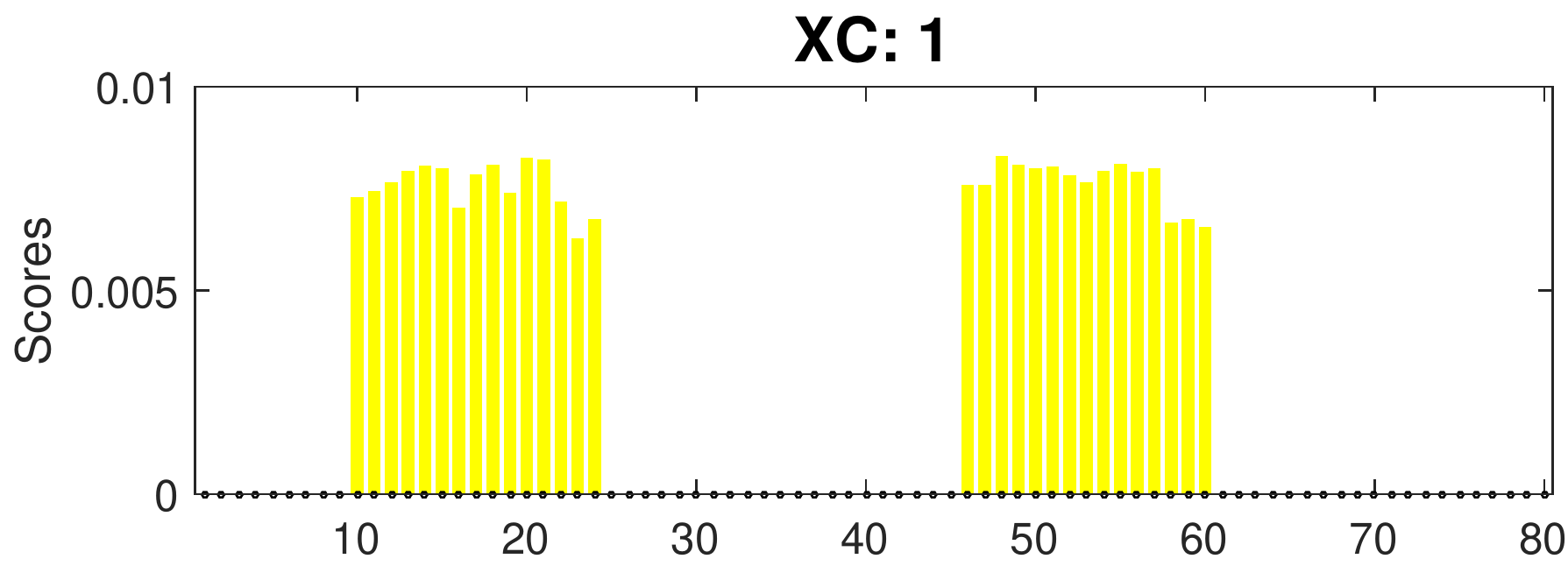}}  \hfill
	{\includegraphics[width=0.45\textwidth]{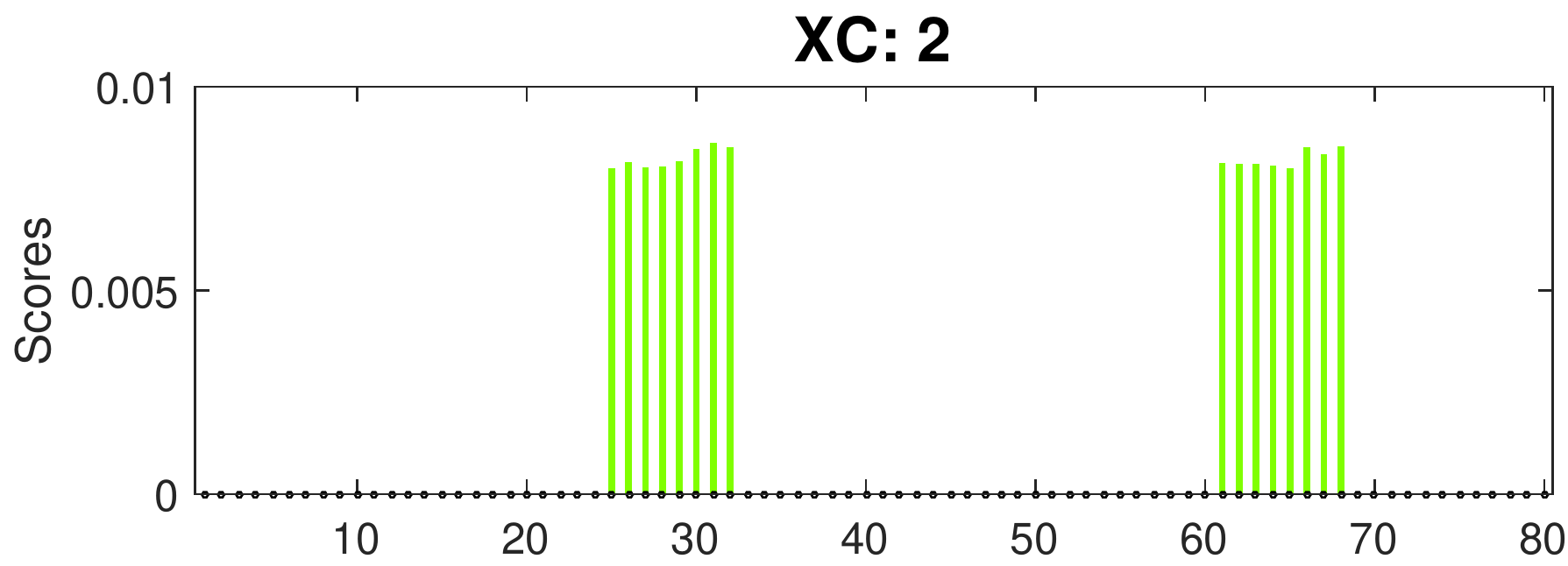}}  \\ 
	{\includegraphics[width=0.45\textwidth]{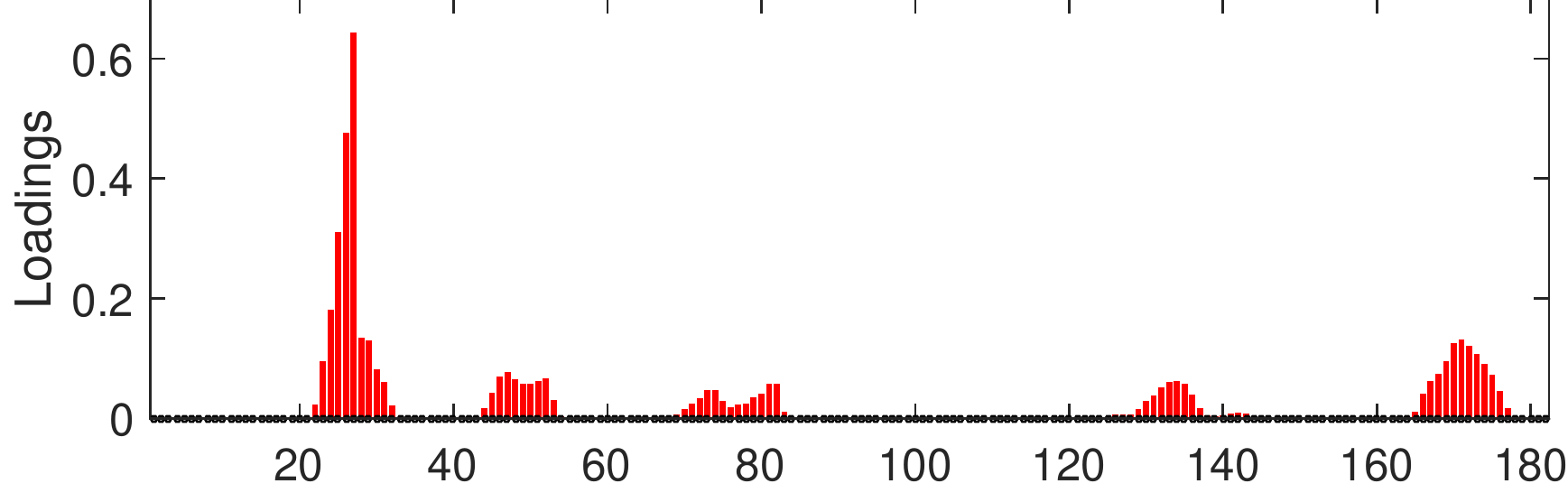}} \hfill
	{\includegraphics[width=0.45\textwidth]{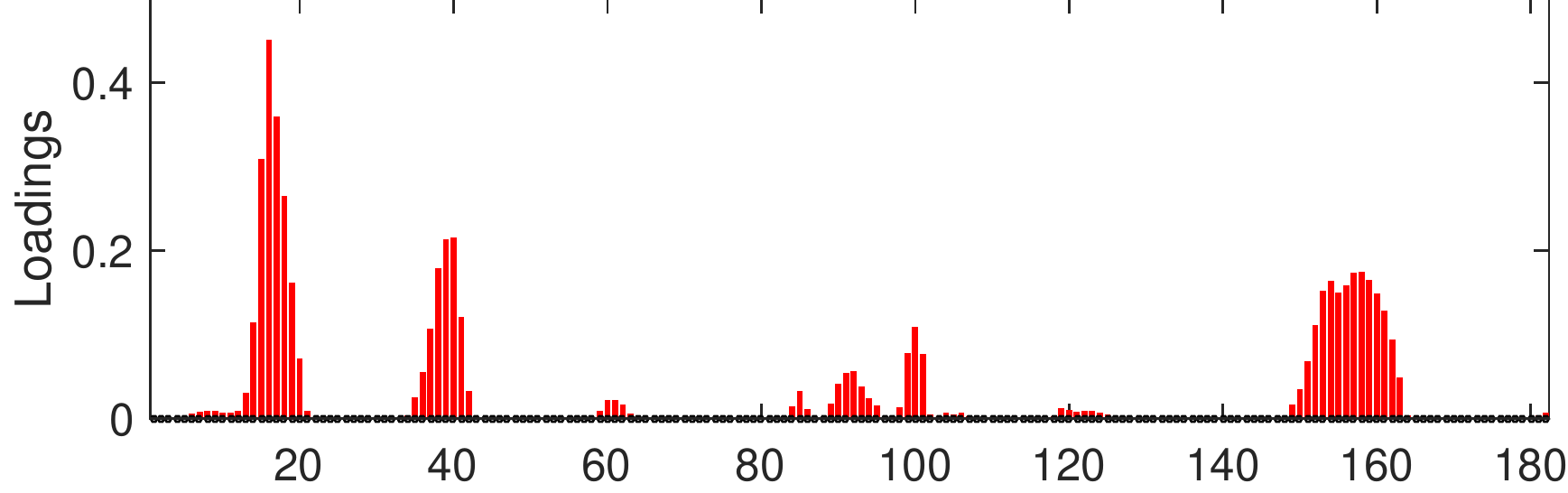}} \\
	{\includegraphics[width=0.45\textwidth]{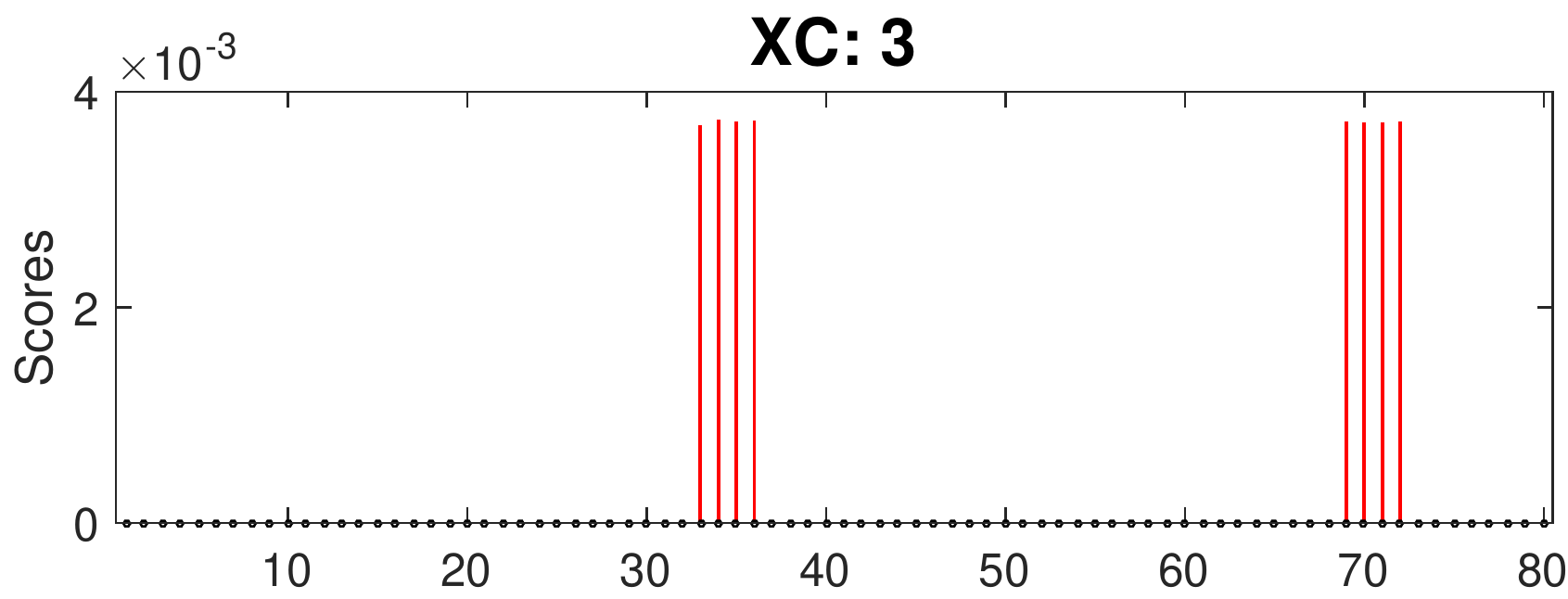}}  \hfill
	{\includegraphics[width=0.45\textwidth]{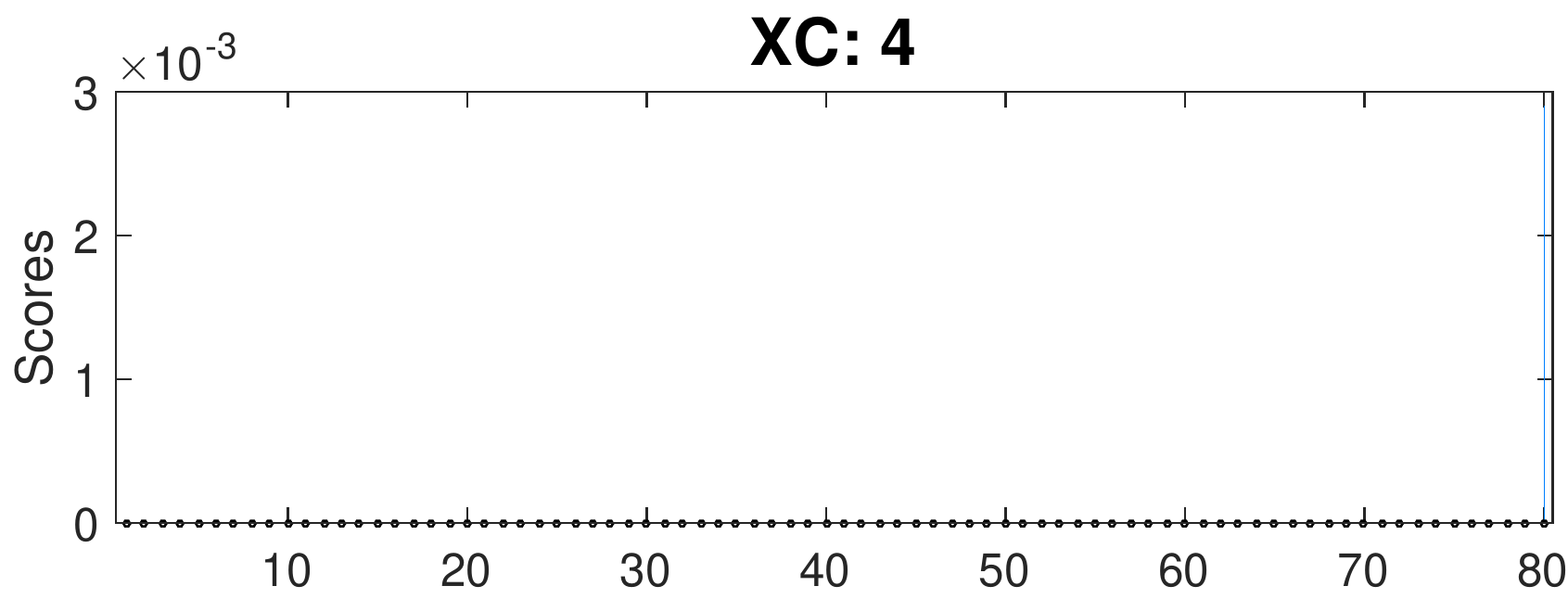}}  \\ 
	{\includegraphics[width=0.45\textwidth]{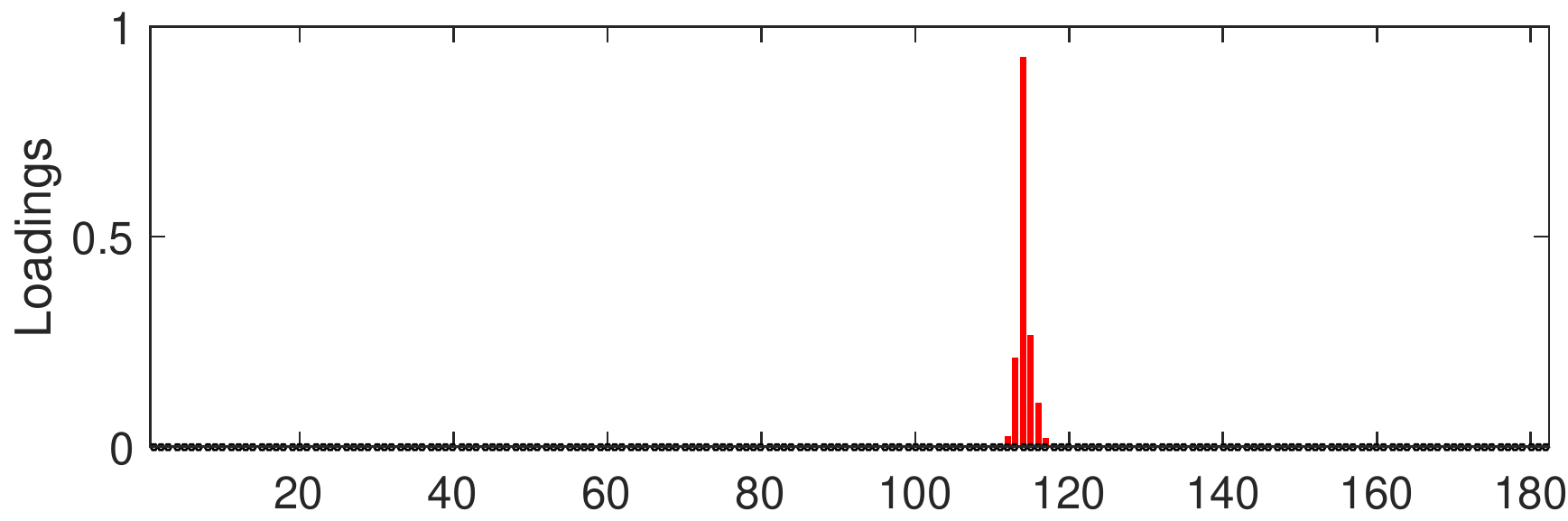}} \hfill
	{\includegraphics[width=0.45\textwidth]{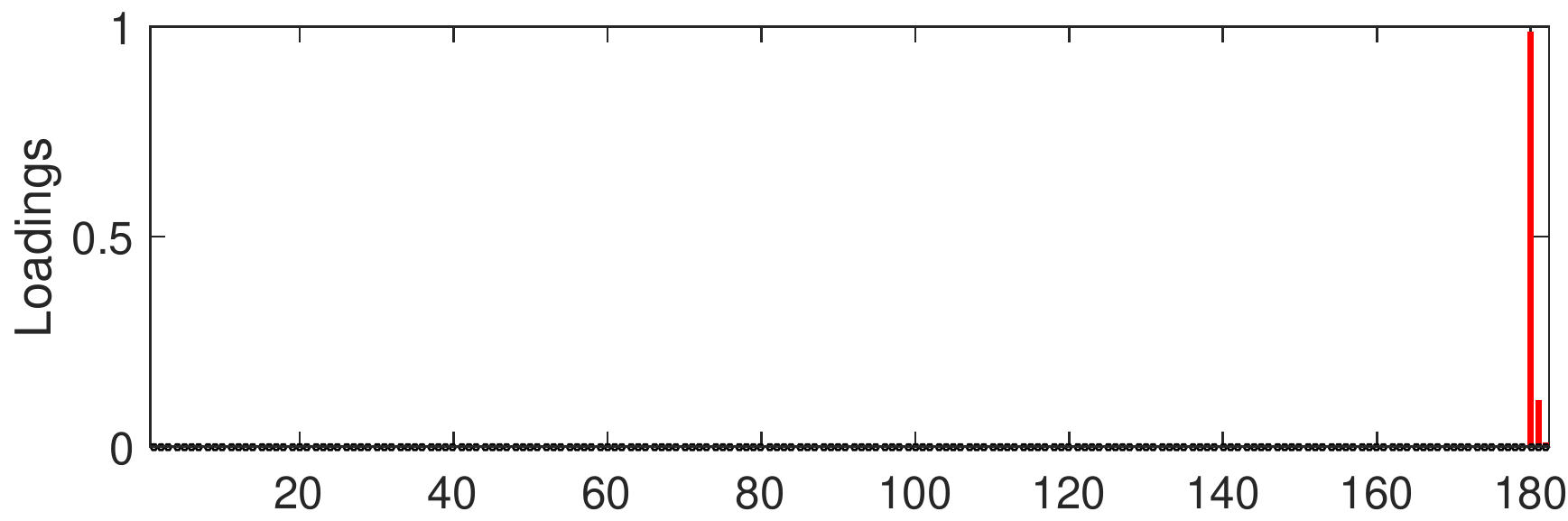}}  \caption{Olive data. XCAN model with baseline and non-negativity constraints on $\mathbf{\hat{U}}^X$, $\mathbf{\hat{S}}^X$ and $\mathbf{\hat{T}}^X$, penalized with the cross-product matrix in Figure \ref{fig:OlXX}(a) and (c).}
	\label{fig:XCAN_NN2}
\end{figure*}

The 4-component XCAN model constrained only with $\mathbf{XtX}$, along with the baseline, is shown in Figure \ref{fig:XCAN_NN}. The model is similar to the one maximizing variance but with sparse loadings. Figure \ref{fig:XCAN_NN2} shows the same analysis using the observation class map within XCAN. Now each component shows the particularities of a different oil brand. Notice this works as a description of the oil brands, not a discrimination. For instance, the first component shows the pattern of the yellow class, without an attempt to distinguish it from the other brands. It might not be a complete description, though. For instance, look at the fourth component. It is focused on the blue class of oil, which is different from the rest in the last interval of wavelengths. However, XCAN only shows one wavelength in this component, because this wavelength is uncorrelated to the others where the blue brand shows its peculiarities. This is interesting information that can improve interpretation but cannot be found with similar sparse methods.
 
\section{Conclusion}

In this paper, we introduce the Cross-product penalized component analysis (XCAN) method and illustrate its use with examples in different application domains. 
XCAN combines variance maximization and structural penalties, that are specified in the form of cross-product matrices and result in sparse matrix factorizations. This provides a flexible modeling framework to explore complex data {and enhance the structural information}.
We plan to extend the application of XCAN to a variety of problems, in particular to data fusion and the derivation of gray models with a-priori information.

\section*{Appendix: Partial Derivatives}

Let us call $F$ the loss function in eq. (\ref{F}). The partial derivatives are as follows:

\begin{equation} \label{derFB}
\mathbf{\frac{\partial F}{\partial U}} = -2(\textbf{X} - \textbf{U} \mathbf{S} \mathbf{P}^T ) \textbf{P} \mathbf{S} + \lambda_0 \mathbf{\frac{\partial F_0}{\partial U}} + \lambda_r \mathbf{\frac{\partial F_r}{\partial U}}
\end{equation}

\begin{equation} \label{derFD}
\mathbf{\frac{\partial F}{\partial S}} = -2 \textbf{U}^T (\textbf{X} - \textbf{U} \mathbf{S} \mathbf{P}^T ) \textbf{P} 
\end{equation}

\begin{equation} \label{derFA}
\mathbf{\frac{\partial F}{\partial P}} = -2(\textbf{X} - \textbf{U} \mathbf{S} \mathbf{P}^T )^T \textbf{U} \mathbf{S} + \lambda_0 \mathbf{\frac{\partial F_0}{\partial P}} + \lambda_c \mathbf{\frac{\partial F_c}{\partial P}}
\end{equation}

\noindent where all partial derivatives with respect to $\mathbf{U}$, $\mathbf{D}$ and $\mathbf{P}$ are of the same dimension of the corresponding matrices. The partial derivatives of the terms in the loss functions $F_0$, $F_r$ and $F_c$ can be calculated element-wise, {using the symmetric structure of $\mathbf{XtX}$ and $\mathbf{XXt}$}:

\begin{equation} \label{derF0B}
\frac{\partial F_0}{\partial u_{hi}} = 4 u_{hi} (\mathbf{u}_h^T\mathbf{u}_h-1)
\end{equation}

\begin{equation} \label{derFcB}
\frac{\partial F_r}{\partial u_{hi}} = 4 u_{hi} \mathbf{n}_{hi}^T\mathbf{n}_{hi}, 
\end{equation}

\begin{equation} \label{derF0A}
\frac{\partial F_0}{\partial p_{hj}} = 4 p_{hj} (\mathbf{p}_h^T\mathbf{p}_h-1)
\end{equation}

\begin{equation} \label{derFrA}
\frac{\partial F_c}{\partial p_{hj}} = 4 p_{hj} \mathbf{m}_{hj}^T\mathbf{m}_{hj}, 
\end{equation}

\noindent with  $\mathbf{n}_{hi} =  \mathbf{u}_h \oslash \mathbf{x}\mathbf{x}t_i$, being $\mathbf{x}\mathbf{x}t_i$ the $i$-th row of $ \mathbf{XXt}$, and $\mathbf{m}_{hj} =  \mathbf{p}_h \oslash \mathbf{x}t\mathbf{x}_j$, being $\mathbf{x}t\mathbf{x}_j$ the $j$-th row of $ \mathbf{XtX}$.
	
 \section*{Acknowledgement}
 \label{sec:Acknowledgments}
 This work is partly  supported by the Spanish Ministry of Economy and Competitiveness and FEDER funds through  project TIN2017-83494-R and the ``Plan Propio de la Universidad de Granada".

\bibliographystyle{ama}
\bibliography{Bibliography}

\end{document}